\documentclass[sigconf]{aamas} 

\usepackage{balance} 

\renewcommand*\backref[1]{\ifx#1\relax \else (Cited in Section #1) \fi}

\usepackage{amsfonts}  
\usepackage{amsmath, amssymb, amsthm}
\usepackage{bm}
\usepackage{wrapfig}
\usepackage{subcaption}
\usepackage{float}
\usepackage{array}
\usepackage{color}
\usepackage{xcolor}
\usepackage{xkcdcolors} 
\usepackage{enumitem}
\usepackage{algorithm}
\usepackage{algpseudocode}
\usepackage{setspace}

\definecolor{madiGreen}{HTML}{55A868}

\newtheorem{definition}{Definition}[section]
\newtheorem{corollary}[definition]{Corollary}

\usepackage{fancyvrb}
\usepackage[scaled=0.8]{DejaVuSansMono}
\usepackage{listings}
\definecolor{codegreen}{rgb}{0,0.5,0}
\definecolor{codered}{rgb}{0.7,0.1,0.1}
\definecolor{codegray}{rgb}{0.5,0.5,0.5}
\definecolor{codepurple}{rgb}{0.58,0,0.82}
\definecolor{backcolour}{rgb}{1,1,1}
\lstdefinestyle{python}{
    language=Python,
    backgroundcolor=\color{backcolour},   
    commentstyle=\color{codered}\textit,
    keywordstyle=\bfseries\color{codegreen},
    numberstyle=\tiny\color{codegray},
    stringstyle=\color{codepurple},
    basicstyle=\ttfamily\small,
    breakatwhitespace=false,         
    breaklines=true,                 
    captionpos=b,                    
    keepspaces=true,                 
    numbers=none,                    
    numbersep=5pt,                  
    showspaces=false,                
    showstringspaces=false,
    showtabs=false,                  
    tabsize=2,
    fancyvrb=true
}
\newenvironment{codesnippet}
  { \VerbatimEnvironment%
    \begin{Verbatim} }
  { \end{Verbatim}  }

\setcopyright{ifaamas}
\acmConference[AAMAS '24]{Proc.\@ of the 23rd International Conference
on Autonomous Agents and Multiagent Systems (AAMAS 2024)}{May 6 -- 10, 2024}
{Auckland, New Zealand}{N.~Alechina, V.~Dignum, M.~Dastani, J.S.~Sichman (eds.)}
\copyrightyear{2024}
\acmYear{2024}
\acmDOI{}
\acmPrice{}
\acmISBN{}

\acmSubmissionID{68}

\title[MaDi]{MaDi: Learning to Mask Distractions for\\ Generalization in Visual Deep Reinforcement Learning}

\author{
Bram Grooten$^1$, 
Tristan Tomilin$^1$, 
Gautham Vasan$^2$, 
Matthew E. Taylor$^{2,3}$,\\ 
A. Rupam Mahmood$^{2,3}$,
Meng Fang$^4$,
Mykola Pechenizkiy$^1$,
Decebal Constantin Mocanu$^{5,1}$
}
\affiliation{\institution{
$^1$Eindhoven University of Technology\ \ \ \  
$^2$University of Alberta\ \ \ \ 
$^3$Alberta Machine Intelligence Institute (Amii)\\
$^4$University of Liverpool\ \ \ \ 
$^5$University of Luxembourg
}
\city{
}
\country{}
}

\begin{abstract}
The visual world provides an abundance of information, but many input pixels received by agents often contain distracting stimuli. \mbox{Autonomous} agents need the ability to distinguish useful information from task-irrelevant perceptions, enabling them to generalize to unseen environments with new distractions. 
Existing works approach this problem using data augmentation or large auxiliary networks with additional loss functions.
We introduce \textit{MaDi}, a novel algorithm that learns to \textbf{ma}sk \textbf{di}stractions by the reward signal only. 
In MaDi, the conventional actor-critic structure of deep reinforcement learning agents is complemented by a small third sibling, the Masker.
This lightweight neural network generates a mask to determine what the actor and critic will receive, such that they can focus on learning the task. The masks are created dynamically, depending on the current input.
We run experiments on the DeepMind Control Generalization Benchmark, the Distracting Control Suite, and a real UR5 Robotic Arm. 
Our algorithm improves the agent's focus with useful masks, while its efficient Masker network only adds $0.2\%$ more parameters to the original structure, in contrast to previous work.
MaDi consistently achieves generalization results better than or competitive to state-of-the-art methods.%
\footnote{Code: \href{https://github.com/bramgrooten/mask-distractions}{github.com/bramgrooten/mask-distractions} and video: \href{https://youtu.be/2oImF0h1k48}{youtu.be/2oImF0h1k48}. Corresponding author: \texttt{b.j.grooten@tue.nl}}
\end{abstract}


\newcommand{\BibTeX}{\rm B\kern-.05em{\sc i\kern-.025em b}\kern-.08em\TeX}

\begin{document}

\pagestyle{fancy}
\fancyhead{}
\maketitle 

\fancypagestyle{plain}{%
\fancyhf{}
\fancyfoot[C]{\sffamily\fontsize{9pt}{9pt}\selectfont\thepage}
\renewcommand{\headrulewidth}{0pt}
\renewcommand{\footrulewidth}{0pt}}
\pagestyle{plain}

\section{Introduction}

Deep reinforcement learning (RL) has achieved remarkable success in a variety of complex tasks such as game playing \citep{schrittwieser2020mastering, mnih2015human}, robotics \citep{akkaya2019solving, haarnoja2023learning, open_x_embodiment_rt_x_2023}, nuclear fusion \citep{degrave2022magnetic}, and autonomous navigation \citep{mirowski2016learning, zhu2016target}. However, one of the major challenges faced by RL agents is their limited ability to generalize to unseen environments, particularly in the presence of distracting visual noise, such as a video playing in the background \citep{stone2021distracting, hansen2021soda}. 
These distractions can lead to \mbox{significant} degradation in the performance of deep RL agents, thereby hindering their applicability in the real world. To address this, we propose a novel algorithm, \textbf{Ma}sking \textbf{Di}stractions, which learns to filter out task-irrelevant visuals, enhancing generalization capabilities. 

\begin{figure*}
    \centering
    \includegraphics[width=0.9\textwidth]{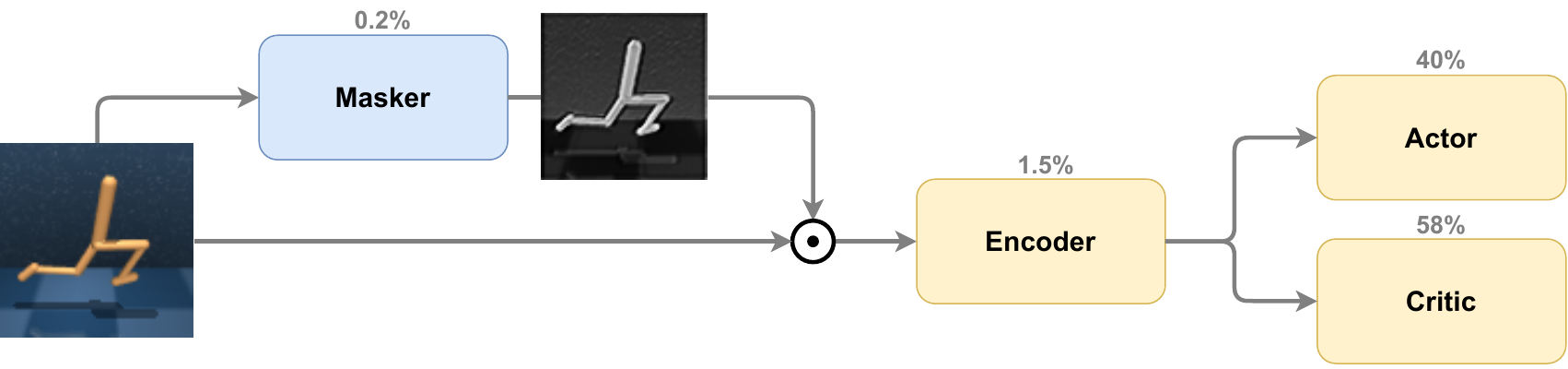}
    \vspace{-0.5em}
    \caption{In the MaDi architecture, the Masker produces a soft mask (values between 0 and 1) for each frame, which subsequently gets multiplied element-wise with the observation. The encoder is a shared ConvNet updated only by the critic loss, which is also the objective function for the Masker. We show rounded percentages for the number of parameters used in the \texttt{walker-walk} environment. The actor and critic contain most of it ($\sim 98\%$ together) as they consist of multiple fully-connected layers.}  
    \label{fig:overview}
    \Description{A schematic overview of the MaDi network architecture. Describing from left to right: an observation of the walker-walk environment starts on the left. Arrows from it point toward the "Masker" block and an Hadamard product sign (which is a circle with a dot in it). Out of the Masker comes a similar image of the walker, but now in light grey for the agent, and a dark background. From this mask there's an arrow to the Hadamard product sign as well. From that sign an arrow points to a "Encoder" block. Then, from that block two arrows point to "Actor" up top and "Critic" at the bottom. Each model block has a percentage above it. Masker: 0.2\%, Encoder: 1.5\%, Actor: 40\%, Critic: 58\%.}
\end{figure*}

The key idea behind MaDi is to supplement the conventional actor-critic architecture with a third lightweight component, the Masker (see Figure~\ref{fig:overview}). This small neural network generates a mask that dims the irrelevant pixels, allowing the actor and critic to focus on learning the task at hand without getting too distracted. 
Unlike previous approaches that have attempted to address this issue \citep{hansen2021soda, hansen2021svea, bertoin2022sgqn}, our method increases generalization performance while introducing minimal overhead in terms of model parameters, thus preserving the efficiency of the original architecture.

Furthermore, no additional loss function is necessary for the Masker to optimize its parameters.
To ensure that the Masker maintains visibility of the task-relevant pixels, it is trained on the critic's loss function. The Masker and critic networks are aligned in their objective, as pixels that are essential to determine the value of an observation should not be hidden. Figure~\ref{fig:overview} shows an example of a mask, visualizing the output produced by the Masker network corresponding to the current input frame. The Masker is able to learn such precise segmentations without any additional labels, bounding boxes, or other annotations. The reward alone is enough.

To evaluate the effectiveness of MaDi, we conduct experiments on multiple environments from three benchmarks: the DeepMind Control Generalization Benchmark \citep{hansen2021soda}, the Distracting Control Suite \citep{stone2021distracting}, and a real UR5 Robotic Arm for which we design a novel generalization experiment with visual distractions.
Our \mbox{results} demonstrate that MaDi significantly improves the agent's ability to focus on relevant visual information by generating helpful masks, leading to enhanced generalization performance. Furthermore, MaDi achieves state-of-the-art performance on many environments, surpassing well-known methods in vision-based reinforcement learning \citep{sac, laskin2020rad, drq, hansen2021soda, hansen2021svea, bertoin2022sgqn}.

Our main contributions are:
\begin{itemize}
    \item We introduce a novel algorithm, MaDi, which supplements the standard actor-critic architecture of deep RL agents with a lightweight Masker. This network learns to focus on the task-relevant pixels solely from the reward signal.
    \item We present a comprehensive set of experiments on the DeepMind Control Generalization Benchmark and the Distracting Control Suite. 
    MaDi consistently achieves state-of-the-art or competitive generalization performance.
    \item We test MaDi on a physical robot, demonstrating that our algorithm increases the performance of the UR5 Robotic Arm in a challenging VisualReacher task, even when \mbox{distracting} videos are playing in the background.
\end{itemize}


\noindent
The paper is structured as follows: Section~\ref{sec:related_work} reviews related work, Section~\ref{sec:preliminaries} formalizes our mathematical framework. Our algorithm MaDi is detailed in Section~\ref{sec:method}. We present simulation results in Section~\ref{sec:experiments} and robotic experiments in Section~\ref{sec:robotic-exps}. Section~\ref{sec:conclusion} concludes.


\section{Related Work}
\label{sec:related_work}

The problem of generalization in deep reinforcement learning has been an active area of research, with several approaches proposed to tackle the challenge of visual distractions. In this section, we review the most relevant literature, highlighting the differences between our proposed MaDi method and existing approaches.

\paragraph{Generalization in RL}

In reinforcement learning, generalization refers to an agent's ability to perform well on unseen environments or tasks \citep{kirk2023survey}. This can be challenging, as RL is prone to overfit to the training environment \citep{zhang2018study, farebrother2018generalization, cobbe2019quantifying, hansen2020pad}.
Several works have focused on improving generalization capabilities by employing techniques such as domain adaptation \citep{xing2021domain}, domain randomization \citep{tobin2017domain, akkaya2019solving}, meta-learning \citep{wang2016learning, duan2016rl}, contrastive learning \citep{laskin2020curl, agarwal2020contrastive}, imitation learning \citep{fan2021secant}, bisimulation metrics \citep{ferns2011, zhang2021learning}, and data augmentation \citep{hansen2021soda, drq, laskin2020rad, yuan2022tlda, raileanu2021automatic, wang2020improving}. Even using a ResNet \cite{he2016resnet} pretrained on ImageNet \citep{imagenet} as an encoder can improve generalization \citep{yuan2022pre}. 


\paragraph{Visual Learning in RL}

Learning tasks from visual input, i.e., image-based or vision-based deep RL, is typically more demanding than learning from direct features in a vector. DQN \citep{mnih2015human} was the first to learn Atari games at human-level performance directly from pixels. However, it has been shown that these algorithms can be quite brittle to changes in the environment, as altering a few pixel values can significantly decrease DQN's performance \citep{qu2020minimalistic, zhang2020robust}. 
Using data augmentation proved to be the key in visual RL. DrQ \citep{drq} and RAD \citep{laskin2020rad} use light augmentations such as random shifts or crops of the observation to increase the algorithm's robustness.


\paragraph{Distractions in RL}

Several approaches have been proposed to deal with the presence of task-irrelevant noise and distractions in reinforcement learning environments. Automatic Noise Filtering (ANF) \cite{grooten2023automatic} works on noisy environments that provide states as feature vectors, such as the MuJoCo Gym suite \citep{mujoco, gym}. 
We focus our work on RL agents that need to learn from image-based observations, like the pixel-wrapped DeepMind Control Suite \citep{tassa2018deepmind}. Two benchmarks that we use are extensions of this suite.  

Several works \citep{hansen2021soda, hansen2021svea, bertoin2022sgqn, yuan2022tlda, yuan2022pre} have tried to tackle the DeepMind Control Generalization Benchmark \citep{hansen2021soda} and the Distracting Control Suite \citep{stone2021distracting}.
Many of these methods use stronger\footnote{By \textit{stronger}, we mean augmentations that alter an image significantly more.}
data augmentations than the light shifting and cropping of DrQ and RAD. Usually, they apply one of two favored augmentation techniques: a randomly initialized convolution layer (\texttt{conv} augmentation) or overlaying the observation with random images from a large dataset, such as Places365 \citep{zhou2017places} (\texttt{overlay} augmentation). 

\paragraph{Masking in Visual RL}

There exist a few works that aim to improve the generalization ability of RL agents by masking parts of the input. Yu et al. \citep{yu2022mask} randomly mask parts of the inputs and use an auxiliary loss to reconstruct these pixels. SGQN \citep{bertoin2022sgqn} and the recent InfoGating \citep{tomar2023ignorance} apply more targeted masking similar to MaDi. InfoGating has only experimented on offline RL, and it uses a large U-Net \citep{ronneberger2015unet} to \mbox{determine} the appropriate masks, while MaDi uses a much smaller 3-layer convolutional neural network.

\paragraph{Baselines}

We select the following set of six baselines for our experiments, as these focus on online RL and do not use any pretrained models:

\begin{itemize}
    \item \textit{Soft Actor-Critic} \citep[\textbf{SAC}]{sac} is an off-policy actor-critic algorithm that optimizes the trade-off between exploration and exploitation by automatically tuning a temperature parameter for entropy regularization.
    \item \textit{Data-regularized Q-learning} \citep[\textbf{DrQ}]{drq} focuses on making Q-learning more stable and sample efficient by shifting the observations by a few pixels in a random direction.
    \item \textit{RL with Augmented Data} \citep[\textbf{RAD}]{laskin2020rad} improves the data efficiency in visual RL by randomly cropping the images.
    \item \textit{Soft Data Augmentation} \citep[\textbf{SODA}]{hansen2021soda} applies data augmentation in an auxiliary task that tries to minimize the distance of augmented and non-augmented images in its feature space.
    \item \textit{Stabilized Value Estimation under Augmentation} \citep[\textbf{SVEA}]{hansen2021svea} stabilizes learning by using augmentation solely in the critic. It combines clean and augmented data in every batch used for a critic update. The actor only sees clean data. 
    \item \textit{Saliency Guided Q-Networks} \citep[\textbf{SGQN}]{bertoin2022sgqn} is perhaps closest to our work, as it also uses masks to benefit learning. Its masks are not applied at the start of the architecture, but are learned by a third component after the encoder. This auxiliary model minimizes the difference between its masks and other masks generated by a saliency metric. By computing the gradient of the Q-function with respect to the input pixels, this saliency metric determines which pixels are important for the agent. A hyperparameter \texttt{sgqn\_quantile} (often set to $95\%-98\%$) determines how many pixels are masked.
\end{itemize}

\noindent
There are multiple significant differences between SGQN and MaDi. First of all, SGQN can be quite sensitive to the quantile hyperparameter. MaDi is free from this hyperparameter tuning, as it automatically finds the right fraction of pixels to mask. Furthermore, SGQN needs to compute gradients with respect to the inputs and weights, while MaDi only requires gradients of the weights. The additional components of SGQN are heavier, as they add about $1.6$M parameters (an extra $25\%$) to the base architecture, with MaDi just adding roughly $10$K ($0.2\%$), reducing the memory requirements.
MaDi also does not introduce any additional auxiliary loss function, as it is able to learn directly from the critic's objective. 
In essence, SGQN does not apply a mask to every input image, but uses them to learn better representations. MaDi tries to learn the most helpful masks such that the actor and critic receive only task-relevant information and are able to focus on the RL problem.




\section{Preliminaries}
\label{sec:preliminaries}



\paragraph{Problem formulation.}
We consider the problem of learning a policy for a Markov decision process (MDP) with the presence of visual distractions, similar to the formulation by Hansen et al. \citep{hansen2021svea}. Our approach, MaDi, aims to learn a policy that generalizes well across MDPs with varying state spaces. 

We formulate the interaction between the environment and policy as an MDP $\mathcal{M}=\langle\mathcal{S}, \mathcal{A}, \mathcal{P}, r, \gamma\rangle$, where $\mathcal{S}$ is the state space, $\mathcal{A}$ is the action space, $\mathcal{P}\colon\mathcal{S}\times\mathcal{A} \to \mathcal{S}$ is the state transition function, $r\colon \mathcal{S}\times\mathcal{A} \to \mathbb{R}$ is the reward function, and $\gamma$ is the discount factor. 
To address the challenges of partial observability \citep{kaelbling1998planning}, we define a state $\mathbf{s}_{t}$ as a sequence of $k$ consecutive frames $(\mathbf{o}_{t}, \mathbf{o}_{t-1}, \dots, \mathbf{o}_{t-(k-1)}),~\mathbf{o}_i \in \mathcal{O}$, where $\mathcal{O}$ is the high-dimensional image space. In the particular benchmarks we employ for evaluation, $\mathcal{O} = \mathbb{R}^{84\times84\times3}$ for the simulation environments and $\mathcal{O} = \mathbb{R}^{160\times90\times3}$ for the robotic environment, as we receive RGB colored images as input with $84\times84$ and $160\times90$ pixels respectively.

Our goal is to learn a stochastic policy $\pi: \mathcal{S} \rightarrow \Delta(\mathcal{A})$, where $\Delta(\mathcal{A})$ denotes the space of probability distributions over the action space $\mathcal{A}$.
This policy aims to maximize the discounted return 
$R_{t} =\mathbb{E}_{\Gamma\sim\pi,\mathcal{P}} \big{[} \sum_{t=0}^{T} \gamma^{t} r(\mathbf{s}_{t}, \mathbf{a}_{t}) \big{]}$ 
along a trajectory $\Gamma = (\mathbf{s}_{0}, \mathbf{s}_{1},\dots,\mathbf{s}_{T})$. The policy $\pi$ is parameterized by a collection of learnable parameters $\theta$. We aim to learn parameters $\theta$ such that $\pi_{\theta}$ generalizes well across MDPs with perturbed observation spaces, denoted as $\overline{\mathcal{M}} = \langle \overline{\mathcal{S}}, \mathcal{A}, \mathcal{P}, r, \gamma\rangle$, where states $\overline{\mathbf{s}}_{t} \in \overline{\mathcal{S}}$ are constructed from observations $\overline{\mathbf{o}}_{t} \in \overline{\mathcal{O}}$. 
The original observation space $\mathcal{O}$ is a subset of the perturbed observation space $\overline{\mathcal{O}}$, which may contain distractions. 

\paragraph{Distractions.}

We define a distraction to be any input feature that is irrelevant to the task of the MDP, meaning that an optimal policy $\pi^{*}$ and value function $Q^{*}$ remain invariant under alterations of the feature value.
In our case, input features are pixels $p_i \in \mathbb{R}^3$, where a state $\bm{s}$ consists of $n$ pixels: $\bm{s} = (p_1, p_2, \ldots, p_n)$. 

Suppose we encounter a state $\hat{\bm{s}}$, and we wish to determine whether pixel $p_i$ is a distraction in that particular state. Let $\mathcal{S}_i(\hat{\bm{s}})$ be the set of states where only pixel $p_i$ is changed in comparison to $\hat{\bm{s}}$.
Then pixel $p_i$ is considered a distraction in state $\hat{\bm{s}}$ if $\pi^{*}$ and $Q^{*}$ remain invariant across the entire set $\mathcal{S}_i(\hat{\bm{s}})$.
More formally:

\begin{definition}\label{def:distraction}
    A pixel $p_i$ is a \textbf{distraction} in state $\hat{\bm{s}}$ if, for an optimal policy $\pi^*$ and value function $Q^*$ it holds that, for an arbitrary but fixed action $\bm{a}$, we have $\ \forall \ \bm{s} \in \mathcal{S}_i(\hat{\bm{s}}) :$
    \begin{align*}
        \pi^{*}(\bm{a} | \bm{s}) & = \rho^*  & \text{ where probability } \rho^* \in \mathbb{R} \text{ is constant,} \\
        \ Q^{*}(\bm{s}, \bm{a}) & = q^*  & \text{ where value } q^* \in \mathbb{R} \text{ is constant.}
    \end{align*}
\end{definition}

\noindent 
In other words: pixel $p_i$ can take on any value, but the optimal policy will not change. In that particular state $\hat{\bm{s}}$, the pixel $p_i$ is irrelevant to the task and thus a distraction.
From this definition we can derive that the partial derivative of $\pi^{*}$ and $Q^{*}$ with respect to the input feature $p_i$ is zero.

\begin{corollary}
    If $p_i$ is a distraction in state $\hat{\bm{s}}$, then for an arbitrary action $\bm{a}$ we have that
    \begin{equation*}
        \frac{\partial}{ \partial p_i } \pi^{*}(\bm{a} | \hat{\bm{s}})  = 0  \text{ \ \ \ \ \   and \ \ \ \ \  }  \frac{\partial}{ \partial p_i } Q^{*}(\hat{\bm{s}}, \bm{a})  = 0.
    \end{equation*}
\end{corollary}

\noindent
This follows from Definition~\ref{def:distraction} since $\pi^{*}$ and $Q^{*}$ remain constant for varying $p_i$. Optimal policies perfectly ignore distractions, while suboptimal policies (i.e., neural networks during training) may be hindered by distractions.
As distractions have no effect on the optimal policy, they can be safely masked when using $\pi^{*}$. This suggests that when striving to approximate $\pi^{*}$, it may be advantageous to mask distractions as well, a concept at the core of MaDi.

\paragraph{Soft Actor-Critic.}
In this work, we build upon the model-free off-policy reinforcement learning algorithm Soft Actor-Critic (SAC; \citep{sac}). SAC aims to estimate the optimal state-action value function $Q^*$ with its parameterized critic $Q_{\theta_{Q}}$. The actor is represented by a stochastic policy $\pi_{\theta_{\pi}}$, which aims to maximize the value outputted by the critic while simultaneously maintaining high entropy.
The optional shared encoder $f_{\theta_f}$ is often used for SAC in image-based environments. The critic and shared encoder have target networks that start with the same parameters $\theta^{\textnormal{tgt}} = \theta$. These are gradually updated throughout training by an exponential moving average: 
$\theta^{\textnormal{tgt}} \longleftarrow (1 - \tau) \theta^{\textnormal{tgt}} + \tau \theta$.
We will often omit the implied parameter $\theta_N$ in our notation of any network $N$.



\section{MaDi}
\label{sec:method}

MaDi aims to mask distractions that hinder the agent from learning and performing well. We supplement the conventional actor-critic architecture of deep reinforcement learning agents by integrating a third, lightweight component, the Masker network $M$. The Masker adjusts the input by dimming irrelevant pixels, allowing the actor and critic networks to focus on learning the task at hand. The Masker and encoder compute internal representations using the Hadamard product and one forward call each: $f(\bm{s}_t \odot M(\bm{s}_t))$.
Algorithm~\ref{alg:madi_short} indicates the few adjustments necessary to standard SAC (or SVEA, when using the augmentation on line 7). Note that MaDi does not need a target network for the Masker, reducing the additional number of parameters required.

\begin{spacing}{0.5}
\begin{algorithm}[t]   
\caption{~~{\color{xkcdBrightBlue}MaDi} based on SAC~~}
\label{alg:madi_short}
\begin{algorithmic}[1]
\algnotext{EndFor}
\Statex randomly initialize all networks: $\pi$, $Q$, $f$, {\color{xkcdBrightBlue}$M$}
\Statex copy parameters to target networks: $Q^{\textnormal{tgt}}$, $f^{\textnormal{tgt}}$
\For{timestep $t=1...T$}
\Statex \textbf{act:}
\State $\mathbf{a}_{t} \sim \pi\left(\cdot | f(\mathbf{s}_{t} {\color{xkcdBrightBlue}\odot M(\bm{s}_t)})\right)$ \hfill Sample action
\State $\mathbf{s}'_{t}, r_t \sim \mathcal{P}(\cdot | \mathbf{s}_{t}, \mathbf{a}_{t})$ \hfill Perform action in env
\State $\mathcal{B} \leftarrow \mathcal{B} \cup (\mathbf{s}_{t}, \mathbf{a}_{t}, r_t, \mathbf{s}'_{t})$ \hfill Add to replay buffer
\Statex \textbf{update:}
\State $\{\mathbf{s}_{b}, \mathbf{a}_{b}, r_{b}, \mathbf{s}'_{b} \} \sim \mathcal{B}$ \hfill Sample batch $b \subset \mathcal{B}$
\State $\theta_{\pi} \leftarrow \theta_{\pi} - \eta \nabla_{\theta_{\pi}}\! \mathcal{L}_{\pi}\! \left(\mathbf{s}_{b} \right)$  \hfill Update $\pi$  
\State {\color{xkcdDarkGreyBlue}$\mathbf{s}_{b} \leftarrow \textnormal{concat}(\mathbf{s}_{b}, \delta(\mathbf{s}_{b}))$ \hfill Apply augmentation}
\For{network $N$ in [$Q$, $f$, {\color{xkcdBrightBlue}$M$}]}
\State $\theta_N \leftarrow \theta_N - \eta \nabla_{\theta_N}\!  \mathcal{L}_{Q}\! \left( \mathbf{s}_{b}, \mathbf{a}_{b}, r_{b}, \mathbf{s}'_{b} \right)$ \hfill Update $Q$, $f$, {\color{xkcdBrightBlue}$M$} 
\EndFor
\For{network $N$ in [$Q$, $f$]}
\State $\theta^{\textnormal{tgt}}_N \leftarrow (1 - \tau) \theta^{\textnormal{tgt}}_N + \tau \theta_N$ \hfill Update $Q^{\textnormal{tgt}}$, $f^{\textnormal{tgt}}$
\EndFor
\EndFor
\end{algorithmic}
\end{algorithm}    
\end{spacing}


\subsection{The MaDi architecture}
\label{sec:masker-network}

\begin{figure*}
    \centering
    \begin{subfigure}[b]{0.3\textwidth}
        \includegraphics[width=\textwidth]{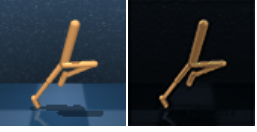}
        \caption{\texttt{training\_env (walker)}}
        \label{fig:subfig1-mask-walker}
    \end{subfigure}
    \hfill
    \begin{subfigure}[b]{0.3\textwidth}
        \includegraphics[width=\textwidth]{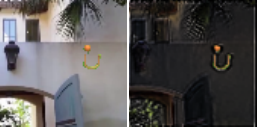}
        \caption{\texttt{video\_hard (ball\_in\_cup)}}
        \label{fig:subfig2-mask-ballcup}
    \end{subfigure}
    \hfill
    \begin{subfigure}[b]{0.3\textwidth}
        \includegraphics[width=\textwidth]{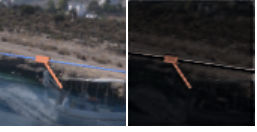}
        \caption{\texttt{distracting\_cs (cartpole)}}
        \label{fig:subfig3-mask-cartp}
    \end{subfigure}
    \vspace{-0.5em}
    \caption{Examples of original observations (left) and their masked versions (right) generated by MaDi in training (a) and testing (b, c) environments. Masks from other benchmark and domain combinations are shown in Appendix~\ref{app:mask-examples}.}
    \label{fig:mask-examples}
    \Description{This shows three sets of images. The first set, labeled as "(a) training env (walker)", displays two images side-by-side, both showing a simplistic 2D figure (walker-walk environment) resembling a walking creature, rendered in golden-yellow against a black background. The image on the right has dark shading around the agent.
The second set, denoted "(b) video_hard (ball_in_cup)", features an image of an outdoor setting. A cup is attached to a string with a ball at its end, all set against walls, plants, and a door in the background. Again, the mask on the right shows the agent (the cup and the ball) more clearly, because the background has been dimmed.
The third set, "(c) distracting_cs (cartpole)", shows two images of an outdoor environment with the blue ground from the DeepMind Control environment below. A pole, depicted in the same golden-yellow shade, is suspended at an angle, in the same position in both images. The right image shows the cartpole more clearly, again masking the background.}
\end{figure*}

As shown in Figure~\ref{fig:overview}, the Masker network is placed at the front of the agent's architecture. 
It produces a scalar multiplier for each pixel in the input image to determine the degree to which the pixel ought to be darkened. We refer to the output of this network as a \textit{soft} mask.\footnote{We also tried \textit{hard} (i.e., \textit{binary}) masks, but they proved more challenging to train.}
These soft masks are applied element-wise to the observation frames, effectively reducing distractions (see Figure~\ref{fig:mask-examples}). 

The Masker network is composed of three convolutional layers with ReLU non-linearities in between. 
The last layer outputs one channel with a Sigmoid activation function to squeeze values into the interval $[0, 1]$. The Masker receives three input channels, representing the RGB values of one frame $\bm{o}_t$. In Section~\ref{sec:preliminaries}, we defined a full state $\bm{s}_t$ to be a stack of $k$ frames, which is indeed what the actor and critic receive as input. The Masker is the only network that processes each frame separately.
However, we still require only one forward pass through the Masker network for each input $\bm{s}_t$ of $k$ frames, as we efficiently reshape the channels into the batch dimension.
See Appendix~\ref{app:implementation_details} for further implementation details.


\subsection{How does the Masker learn?}
\label{sec:madi-learn}

One may expect that learning to output useful masks requires us to define a separate loss function, but this is not the case. The Masker can simply be updated via the critic's objective function\footnote{Future work could study whether the Masker network can also learn from the actor loss. In many SAC-based implementations with a shared ConvNet, the encoder is only updated by the critic loss, and it made sense to use these gradients for the Masker.}
to update its parameters, as shown on line 9 of Algorithm~\ref{alg:madi_short}. This means the masks are trained without any additional segmentation labels or saliency metrics.
Our hypothesis on the surprising ability of MaDi to determine the task-relevant pixels solely from a scalar reward signal, pertains to the following:
\begin{itemize}
    \item For \textit{relevant} pixels, if the Masker network masks away essential pixels needed to determine an accurate Q-value, then the critic loss will presumably be high, and the Masker will thus be encouraged to leave these pixels visible.
    \item For \textit{irrelevant} pixels, we believe (and empirically show in Section~\ref{sec:madi-sac}) that strong and varying data augmentation helps. It gives an irrelevant pixel in a particular state $\bm{s}$ a varying pixel-value each time state $\bm{s}$ is sampled from the replay buffer, while the pixel's contribution to the $Q^*$-value remains the same (none, because it is irrelevant). The Masker is thus incentivized to mask this pixel, such that the actor and critic networks always see the same pixel-value for state $\bm{s}$, no matter which augmentation is used.
\end{itemize}
The Masker is updated together with the critic, which happens once for every environment step in case of synchronous runs. The robotic experiments use an asynchronous version of each algorithm. In that case, the Masker still gets as many updates as the critic, but it is no longer equal to the number of environment steps.




\section{Simulation Experiments}
\label{sec:experiments}

We present the experiments done on generalization benchmarks based on the DeepMind Control Suite \citep{tassa2018deepmind} in this section, while our robotic experiments are shown in Section~\ref{sec:robotic-exps}.
We describe our experimental setup and provide the results obtained from our method, MaDi, compared with state-of-the-art approaches. Our experiments are designed to demonstrate the effectiveness of MaDi in masking distractions and improving generalization in vision-based RL.

\subsection{Experimental Setup}

\paragraph{Benchmarks.} We evaluate the performance of MaDi on the DeepMind Control Generalization Benchmark \cite[DMControl-GB]{hansen2021soda} and the Distracting Control Suite \cite[DistractingCS]{stone2021distracting}. These benchmarks consist of a range of environments with varying levels of complexity and noise, providing a comprehensive assessment of an agent's ability to generalize to unseen, distracting environments.

\begin{itemize}
    \item \textbf{DMControl-GB} has two setups with task-irrelevant pixels in the background: \texttt{video\_easy} and \texttt{video\_hard}. For the easy setup, there are just 10 videos to randomly sample from, and the surface from the training environment is still shown. In the hard counterpart, the surface is no longer visible, and one of 100 videos is selected. Note that all the frames in these videos are unseen --- they do not overlap with the images in the augmentation dataset we use.

    \item \textbf{DistractingCS} goes a step further, also adjusting the camera's orientation and the agent's color during an episode, while displaying a randomly selected video in the background. The surface remains visible, such that the agent can orient itself during a changing camera angle. The intensity of the DistractingCS determines the difficulty. With higher intensity, the environment (1) samples from a larger set of videos, (2) changes the agent's color faster and to more extreme limits, and (3) adjusts the camera's orientation faster and to more severe angles. We use the default intensity level of $0.1$. See the Supplementary Material for example videos. 
\end{itemize}


\begin{figure}
    \vspace{-0.5em}
    \centering
    \includegraphics[width=0.9\columnwidth]{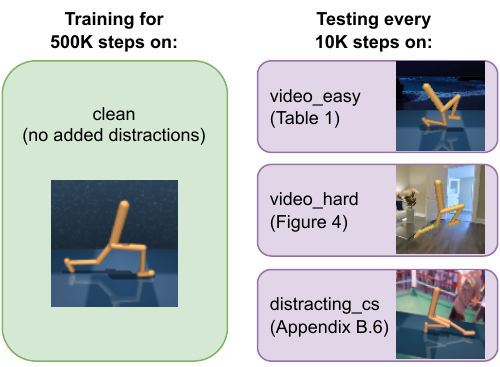}
    \vspace{-0.3em}
    \caption{The training and testing setup.}
    \label{fig:train-test}
    \vspace{-1.0em}
    \Description{The figure illustrates the training and testing setup for our simulation experiments. On the left side, under the heading "Training for 500K steps on:", there's a green rounded rectangular box labeled "clean (no added distractions)". Inside this box is an image of a simple walker figure against the standard blue background of the DeepMind Control Suite.
On the right side, under the heading "Testing every 10K steps on:", three purple rectangular boxes stack vertically, each containing a distinct image:
1. The top box, labeled "video_easy (Table 1)", shows an image of a walker figure on a more intricate dark background. The ground surface of the walker environment is still shown.
2. The middle box, titled "video_hard (Figure 4)", presents the walker figure against a complex indoor setting, a room with furniture and a door opening. No surface of the DMControl environment is shown anymore. Just the agent and the video background.
3. The bottom box, named "distracting_cs (Appendix B.5)", displays a walker figure on an even more colorful and detailed background, suggesting a lot of visual noise. The ground surface is shown again, but it is on an angle, since the camera angle changes in this setting. The background video shows a frame of a boxing match.
This setup visualizes the contrast between the simple training environment and the progressively more challenging testing scenarios. It also shows where the results of each testing environment can be found.}
\end{figure}

\paragraph{Environments.} Within these two benchmarks, we run all algorithms on six distinct environments, listed in Table~\ref{tab:results-video_easy}. From the \texttt{cartpole} and \texttt{walker} domains we select two tasks, which differ by their starting positions and reward functions.
See Appendix~\ref{app:task-descriptions} for a detailed description of each task. 

\paragraph{Models \& Training.} For a fair comparison, we use the same base actor-critic architecture for all the methods considered in this study, including MaDi.
All algorithms are trained for 500K timesteps on the clean training environment without distractions.
We use the default hyperparameters for all baselines, as specified by the DMControl-GB \citep{hansen2021soda}. See Appendix~\ref{app:implementation_details} for an overview of the hyperparameters.

\paragraph{Augmentation.}

As discussed in Section~\ref{sec:related_work}, all of our baselines (except SAC) use some form of data augmentation. 
MaDi is built on top of SVEA \citep{hansen2021svea}, which performs best with the \texttt{overlay} augmentation for distracting video backgrounds. Therefore, we choose to apply \texttt{overlay} for MaDi as well.
This strong augmentation combines an observation frame from the training environment, $\mathbf{o}_{t}$, with a random image $\bm{x}$ from a large dataset as follows:
\begin{equation*}
    \delta_{\bm{x}}(\mathbf{o}_{t}) = \alpha \cdot \mathbf{o}_{t} + (1-\alpha) \cdot \bm{x}
\end{equation*}
where $\delta$ denotes the augmentation function. In the experiments we use the default overlay factor of $\alpha = 0.5$ and sample images from the same dataset as used in SVEA, namely Places365 \citep{zhou2017places}.

\paragraph{Evaluation.}
To assess generalization, we evaluate the trained agents zero-shot on a set of unseen environments with different levels of distractions. 
Specifically, we test on \texttt{video\_easy} and \texttt{video\_hard} from DMControl-GB, and on the Distracting Control Suite.
Every 10K steps we evaluate the current policy for $20$ episodes on the test environments; see Figure~\ref{fig:train-test}.
We report the average undiscounted return over five random seeds during the last $10\%$ of training, a metric often used to reduce variance \citep{grooten2023automatic, graesser2022state}.
We run statistical tests to verify significance, shown in Appendix~\ref{app:additional_results}.



\begin{table}
    \caption{Generalization performance of MaDi and various baseline algorithms on six different environments trained for $500K$ steps. We show undiscounted return on \texttt{video\_easy} with mean and standard error over five seeds. MaDi outperforms or comes close to the state-of-the-art in all environments.}
    \label{tab:results-video_easy}
    \vspace{-0.3em}
    \centering
    \resizebox{\columnwidth}{!}{%
    \begin{tabular}{lccccccc}
    \toprule
    \texttt{video\_easy}  & SAC & DrQ & RAD & SODA & SVEA & SGQN & MaDi \\
\midrule 
\texttt{ball\_in\_cup} & $602$ & $714$ & $561$ & $750$ & $757$ & $761$ & $\bm{807}$  \vspace{-0.75ex} \\
\texttt{catch} & $\scriptstyle{\pm91}$ & $\scriptstyle{\pm131}$ & $\scriptstyle{\pm147}$ & $\scriptstyle{\pm98}$ & $\scriptstyle{\pm138}$ & $\scriptstyle{\pm171}$ &$\bm{\scriptstyle{\pm144}}$ \vspace{0.75ex} \\
\texttt{cartpole} & $924$ & $932$ & $801$ & $961$ & $967$ & $965$ & $\bm{982}$  \vspace{-0.75ex} \\
\texttt{balance} & $\scriptstyle{\pm19}$ & $\scriptstyle{\pm33}$ & $\scriptstyle{\pm95}$ & $\scriptstyle{\pm10}$ & $\scriptstyle{\pm2}$ & $\scriptstyle{\pm5}$ &$\bm{\scriptstyle{\pm4}}$ \vspace{0.75ex} \\
\texttt{cartpole} & $782$ & $613$ & $658$ & $215$ & $786$ & $798$ & $\bm{848}$  \vspace{-0.75ex} \\
\texttt{swingup} & $\scriptstyle{\pm21}$ & $\scriptstyle{\pm74}$ & $\scriptstyle{\pm17}$ & $\scriptstyle{\pm125}$ & $\scriptstyle{\pm15}$ & $\scriptstyle{\pm13}$ &$\bm{\scriptstyle{\pm6}}$ \vspace{0.75ex} \\
\texttt{finger} & $227$ & $543$ & $479$ & $429$ & $645$ & $592$ & $\bm{679}$  \vspace{-0.75ex} \\
\texttt{spin} & $\scriptstyle{\pm26}$ & $\scriptstyle{\pm50}$ & $\scriptstyle{\pm65}$ & $\scriptstyle{\pm100}$ & $\scriptstyle{\pm39}$ & $\scriptstyle{\pm11}$ &$\bm{\scriptstyle{\pm17}}$ \vspace{0.75ex} \\
\texttt{walker} & $507$ & $954$ & $961$ & $147$ & $\bm{977}$ & $672$ & $967$  \vspace{-0.75ex} \\
\texttt{stand} & $\scriptstyle{\pm113}$ & $\scriptstyle{\pm10}$ & $\scriptstyle{\pm1}$ & $\scriptstyle{\pm17}$ &$\bm{\scriptstyle{\pm3}}$& $\scriptstyle{\pm153}$ & $\scriptstyle{\pm3}$  \vspace{0.75ex} \\
\texttt{walker} & $334$ & $821$ & $726$ & $479$ & $\bm{936}$ & $882$ & $895$  \vspace{-0.75ex} \\
\texttt{walk} & $\scriptstyle{\pm37}$ & $\scriptstyle{\pm38}$ & $\scriptstyle{\pm42}$ & $\scriptstyle{\pm168}$ &$\bm{\scriptstyle{\pm14}}$& $\scriptstyle{\pm26}$ & $\scriptstyle{\pm24}$  \\
\midrule
\texttt{avg} & $563$ & $763$ & $698$ & $497$ & $845$ & $778$ & $\bm{863}$  \\
    \bottomrule
    \end{tabular}
    }
    \vspace{-1em}
    \end{table}

\begin{figure*}
    \centering
    \includegraphics[width=0.95\textwidth]{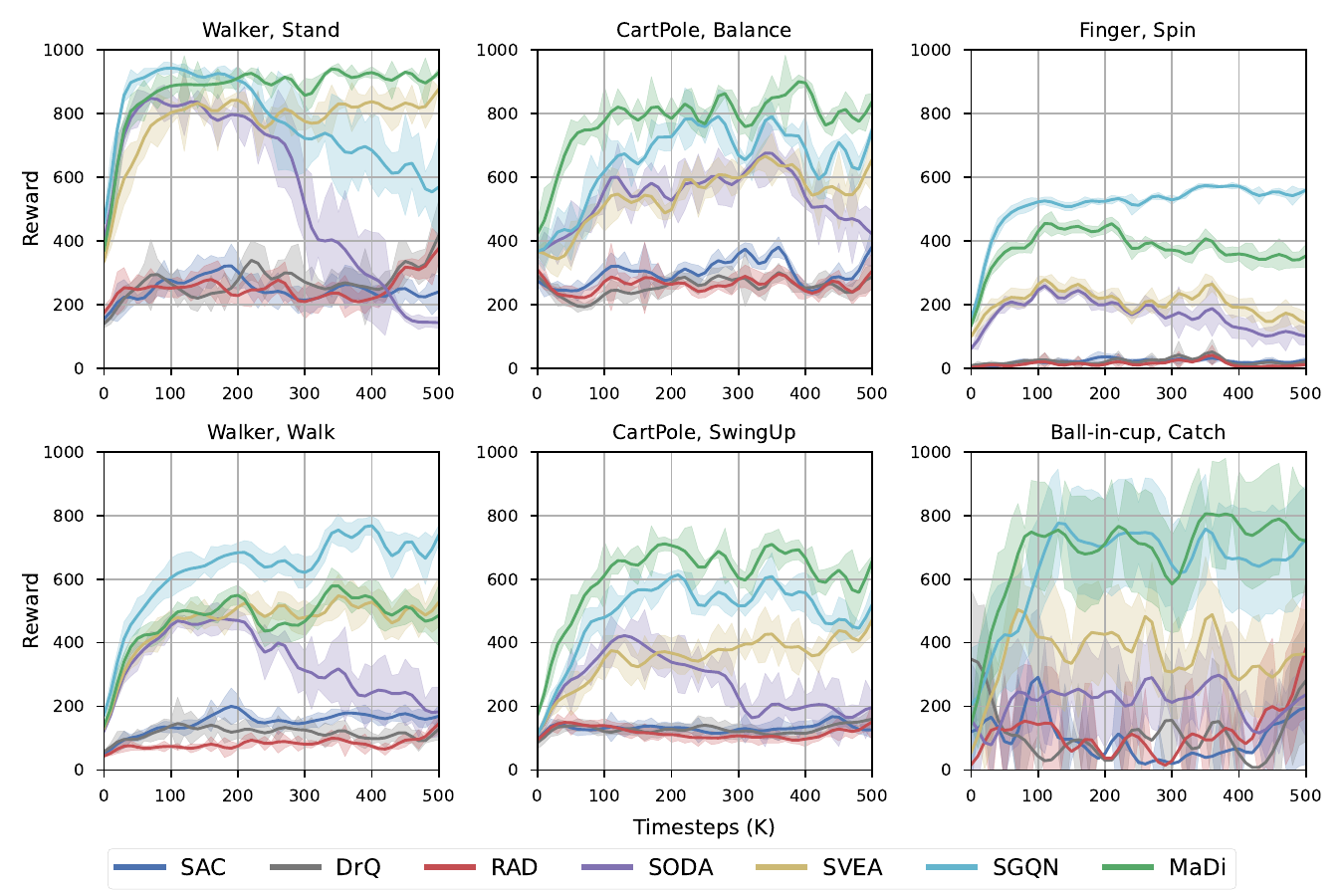}
    \vspace{-0.5em}
    \caption{Learning curves of MaDi and six baselines on \texttt{video\_hard}. Agents are trained on clean data for $500K$ steps and tested on \texttt{video\_hard} every $10K$ steps. MaDi often reaches the top of the class, while some baselines can overfit to the training environment and decrease in generalizability. The curves show the mean over five seeds with standard error shaded alongside.}
    \label{fig:video_hard}
    \Description{This figure displays six graph plots showing the learning curves of MaDi and six other baselines. Each graph plot tracks the total reward on the vertical y-axis, ranging from 0 to 1000 on all plots, against 'Timesteps (K)' on the horizontal x-axis, spanning from 0 to 500 on all. The learning curves are represented using colored lines, with each color corresponding to a different model.
From top-left to bottom-right, the titles of the graph plots are:
1. "Walker, Stand"
2. "CartPole, Balance"
3. "Finger, Spin"
4. "Walker, Walk"
5. "CartPole, SwingUp"
6. "Ball-in-cup, Catch"
Each plot features a variety of curves for the models: SAC (blue), DrQ (gray), RAD (red), SODA (purple), SVEA (yellow), SGQN (light blue), and MaDi (green). The plots depict how each model performs across different timesteps, with some curves reaching higher rewards than others. Notably, the MaDi curve often peaks or is among the highest in each graph.
Only in walker-walk and finger-spin it is not the highest. SGQN performs best there. 
SAC, DrQ, RAD are all near the bottom in each graph, while the others are usually higher.}
\end{figure*}

\subsection{Generalization Results}
\label{sec:gen-results-vid-hard}

In Table~\ref{tab:results-video_easy} we show the results of MaDi and its baselines when generalizing to the \texttt{video\_easy} setup of the DMControl-GB benchmark. MaDi is able to achieve the best or competitive performance in all environments. 
The learning curves of Figure~\ref{fig:video_hard} show that MaDi also generalizes well to the more challenging \texttt{video\_hard} environments. Furthermore, the curves show that MaDi has a high sample efficiency, often reaching adequate performance in just $100$K environment steps, by its ability to focus on the task-relevant pixels.

We present tables with results on the DistractingCS benchmark and the original training environments in Appendix~\ref{app:additional_results}. MaDi also shows competitive performance in these additional settings.
Note that the environments \texttt{ball\_in\_cup-catch} and \texttt{finger-spin} have \textit{sparse} rewards, but even in this setting MaDi is able to perform well and generate useful masks. In Appendix~\ref{app:mask-examples} we provide examples of different masks produced for each domain.



\begin{table}
    \caption{Ablation study showing the effect of the \texttt{overlay} augmentation. SVEA and MaDi both use it, while SAC and MaDi-SAC do not. All algorithms are trained for $500K$ steps. We show mean undiscounted return and standard error over five seeds evaluated on \texttt{video\_hard}. The results reveal that MaDi benefits from data augmentation.}
    \label{tab:results-madi-sac}
    \vspace{-0.25em}
    \centering
    \begin{tabular}{lccccccc}
    \toprule
    \texttt{video\_hard}  & SAC & MaDi-SAC & SVEA & MaDi \\
\midrule 
\texttt{ball\_in\_cup} & $176$ & $190$ & $327$ & $758$  \vspace{-0.75ex} \\
\texttt{catch} & $\scriptstyle{\pm38}$ & $\scriptstyle{\pm52}$ & $\scriptstyle{\pm59}$ & $\scriptstyle{\pm135}$  \vspace{0.75ex} \\
\texttt{cartpole} & $314$ & $237$ & $579$ & $827$  \vspace{-0.75ex} \\
\texttt{balance} & $\scriptstyle{\pm12}$ & $\scriptstyle{\pm6}$ & $\scriptstyle{\pm26}$ & $\scriptstyle{\pm25}$  \vspace{0.75ex} \\
\texttt{cartpole} & $140$ & $132$ & $453$ & $619$  \vspace{-0.75ex} \\
\texttt{swingup} & $\scriptstyle{\pm10}$ & $\scriptstyle{\pm9}$ & $\scriptstyle{\pm26}$ & $\scriptstyle{\pm24}$  \vspace{0.75ex} \\
\texttt{finger} & $21$ & $121$ & $154$ & $358$  \vspace{-0.75ex} \\
\texttt{spin} & $\scriptstyle{\pm4}$ & $\scriptstyle{\pm36}$ & $\scriptstyle{\pm31}$ & $\scriptstyle{\pm25}$  \vspace{0.75ex} \\
\texttt{walker} & $233$ & $320$ & $847$ & $920$  \vspace{-0.75ex} \\
\texttt{stand} & $\scriptstyle{\pm28}$ & $\scriptstyle{\pm88}$ & $\scriptstyle{\pm18}$ & $\scriptstyle{\pm14}$  \vspace{0.75ex} \\
\texttt{walker} & $168$ & $95$ & $526$ & $504$  \vspace{-0.75ex} \\
\texttt{walk} & $\scriptstyle{\pm19}$ & $\scriptstyle{\pm24}$ & $\scriptstyle{\pm55}$ & $\scriptstyle{\pm33}$  \\
\midrule
\texttt{avg} & $175$ & $183$ & $481$ & $664$  \\
    \bottomrule
    \end{tabular}
    \vspace{-1em}
\end{table}

\subsection{Ablation on Augmentation}
\label{sec:madi-sac}


In Section~\ref{sec:madi-learn} we described the expectation that MaDi would perform better with augmentations, as that can help it to recognize which pixels are irrelevant. To verify whether this intuition holds, we run five seeds of MaDi without the \texttt{overlay} augmentation on all six environments. We call this variant MaDi-SAC, as it now builds on top of SAC \citep{sac} instead of SVEA.

The results are shown in Table~\ref{tab:results-madi-sac}. The generalization performance of the algorithms that use augmentation is much better than those without. MaDi indeed benefits from data augmentation to become more robust against previously unseen distractions.
Furthermore, in Appendix~\ref{app:augmentation-details} we present results on a few other types of augmentations that could be combined with our method.

\subsection{Does MaDi work with Vision Transformers?}
\label{sec:vitmadi}

In image-based RL, the use of Vision Transformers \citep[\textbf{ViT}]{dosovitskiy2020image} has recently gained in popularity \citep{hansen2021svea, tao2022evaluating}.
We set up a small experiment to verify whether MaDi still works in this setting. The shared encoder (see Figure~\ref{fig:overview}), which is originally an 11-layer ConvNet, is now replaced by a ViT of 4 blocks with 8 attention heads each. We maintain the same architecture for the Masker network. More implementation details on the ViT encoder are in Appendix~\ref{app:vit-implementation}.

\begin{figure}
    \centering
    \includegraphics[width=\columnwidth]{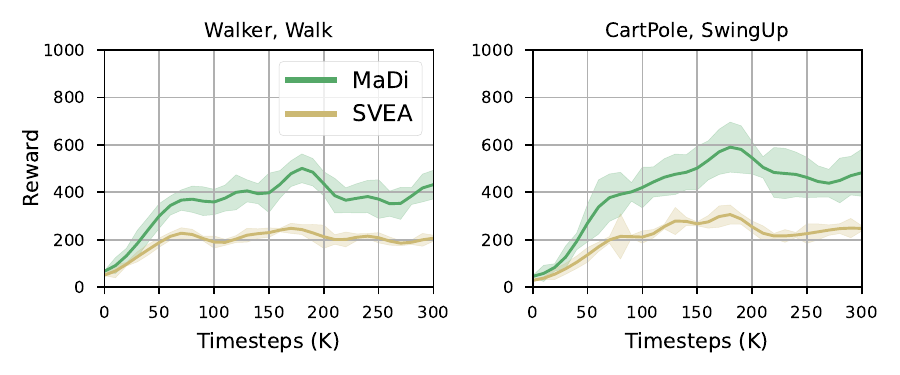}
    \vspace{-2.5em}
    \caption{Generalization performance of MaDi and SVEA on \texttt{video\_hard} when trained with a \textbf{ViT} encoder for $300K$ steps. We show the mean and standard error over five seeds.}
    \label{fig:vit-video_hard}
    \Description{The figure presents two line graphs side-by-side, illustrating the generalization performance of MaDi and SVEA on the video_hard environment when trained with a ViT encoder (instead of a ConvNet encoder) over 300K steps.
On the left graph, titled "Walker. Walk":
The vertical axis represents the 'Reward', ranging from 0 to 1000.
The horizontal axis depicts 'Timesteps (K)' spanning from 0 to 300.
The MaDi learning curve, shown in green, starts around 100 and steadily rises, reaching a plateau around 400 after about 75K steps.
The SVEA curve, in yellow, begins close to the MaDi curve, grows a little bit, but stays relatively flat after 75K as well, fluctuating mildly around 200.
On the right graph, titled "CartPole. SwingUp":
The y-axis and x-axis scales are the same as the left graph.
The MaDi curve, again in green, starts at about 50 and gradually rises, peaking around 600 before stabilizing around the 500 range.
The SVEA curve, in yellow, begins near 50 as well and grows up to about 250 and then plateaus. Both SVEA and MaDi reach their final plateau value after around 125K steps here.
In both graphs, shaded regions around the curves indicate the standard error over five seeds. The overall portrayal shows that MaDi outperforms SVEA in these environments.}
\end{figure}

We train the ViT-based versions of SVEA and MaDi for 300K timesteps on the clean environments of \texttt{walker-walk} and \texttt{cartpole-swingup}, and test on \texttt{video\_hard} at every 10K steps. The results presented in Figure~\ref{fig:vit-video_hard} show that not only does MaDi work well with a ViT encoder, but it even boosts the generalization performance.

\section{Robotic Experiments}
\label{sec:robotic-exps}

In this section, we describe our experiments on a robotic arm visual reacher task, showing that MaDi can learn to generalize when facing distracting backgrounds, even in real-world environments.

\subsection{Experimental Setup}

\paragraph{Robotic Arm}
The UR5 industrial robot arm consists of six joints that can rotate to move the tip to a desired position. It uses the conventional TCP/IP protocol to transmit the state of the arm to the host computer and receive actuation packets in return at an interval of 8 milliseconds. 
A state packet includes the angles, velocities, target accelerations, and currents of all six joints. 
We configured the UR5 to use the velocity control mode. 
Since the UR5 does not come with a camera, we attached a Logitech RGB camera to the tip of the arm to facilitate vision-based tasks. 
We use \emph{SenseAct}, a computational framework for robotic learning experiments to communicate with the robot~\citep{mahmood2018benchmarking}. 
This setup enables robust and reproducible learning across different locations and conditions.

\begin{figure}
    \centering
    \begin{subfigure}[b]{0.48\columnwidth}
        \includegraphics[width=\textwidth]{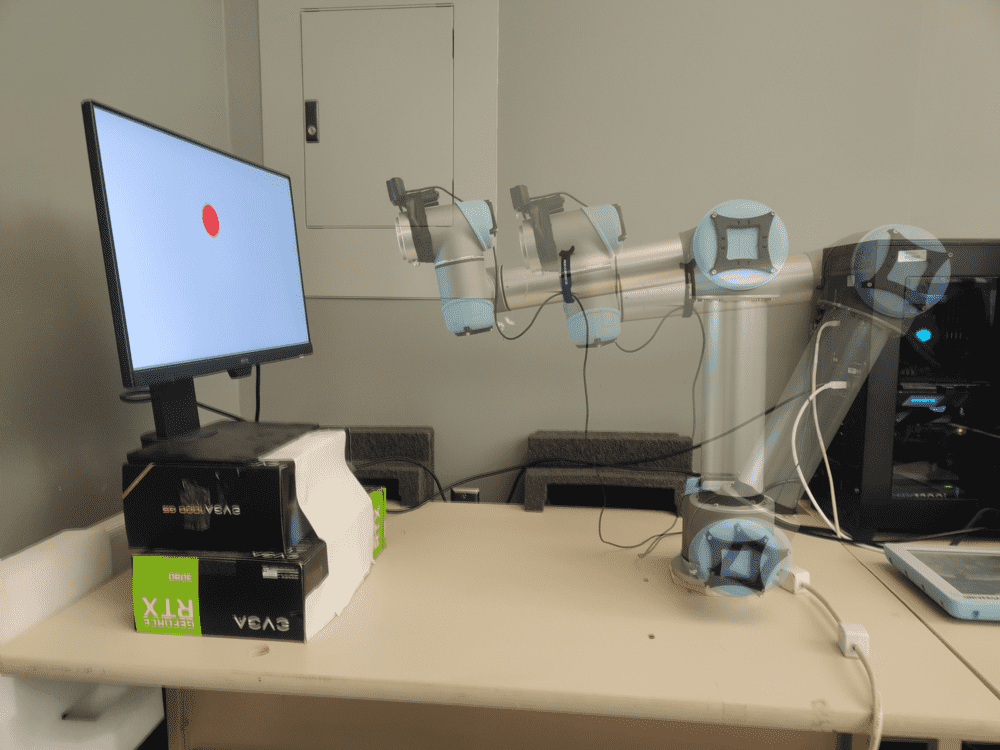}
        \caption{Training environment}
        \label{fig:subfig1-ur5-train-env}
    \end{subfigure}
    \hfill
    \begin{subfigure}[b]{0.48\columnwidth}
        \includegraphics[width=\textwidth]{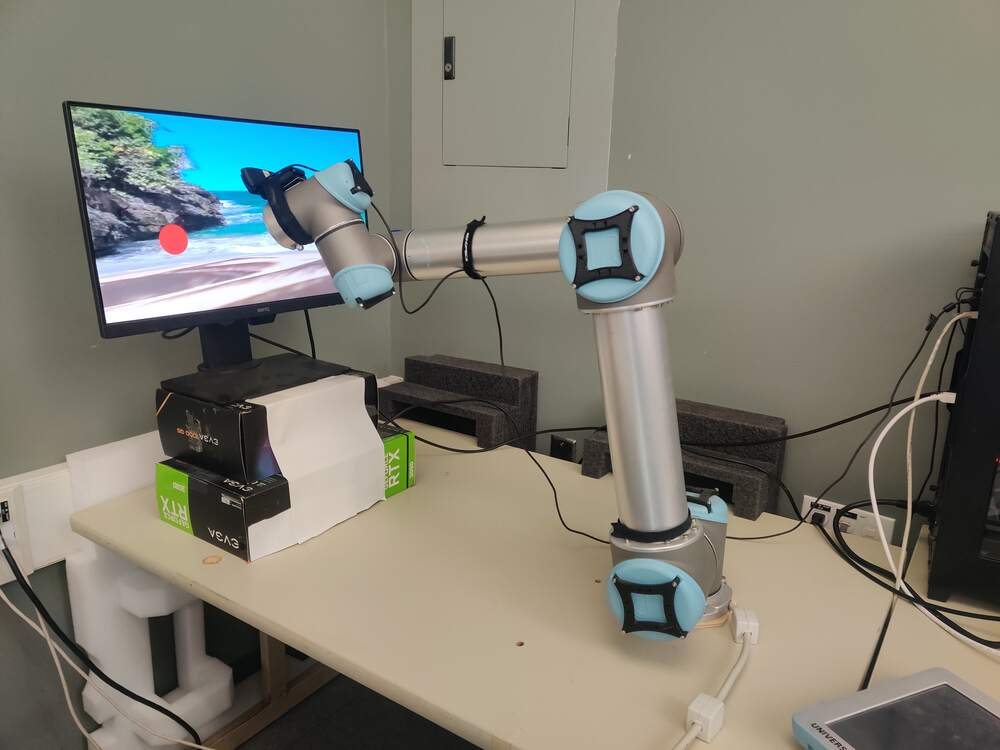}
        \caption{Testing environment}
        \label{fig:subfig2-ur5-test-env}
    \end{subfigure}
    \vspace{-0.2em}
    \caption{The UR5-VisualReacher(-VideoBackgrounds) benchmark used in our experiments. The agent is rewarded for getting its camera as close as possible to the red circle randomly located on the screen. For visual demonstrations of the robot's performance with each algorithm, please refer to the videos in the Supplementary Material.} 
    \label{fig:ur5}
    \vspace{-1em}
\Description{This figure showcases two side-by-side photographs, both depicting a side view of the lab setup for the UR5-VisualReacher(-VideoBackgrounds) benchmark.
On the left, figure (a) Training environment:
This image captures a workspace with a computer monitor showing a white background and a prominently displayed red dot about 3 cm wide. In front of the monitor, the UR5 robotic arm, a six-axis industrial robot, is prominently displayed in two overlaid positions, indicating its movement range during training.
On the right, figure (b) Testing environment:
In this image, the same UR5 robotic arm is in a single position. The computer monitor, instead of the simple blue background from the training setup, now displays a real-world scene of a beach, still with the red dot as well. The UR5 robotic arm is reaching towards this red circle with its webcam attached to the top of the tip of the arm.}
\end{figure}

\paragraph{Training environment}
We train on the UR5-VisualReacher task \citep{yuan2022asynchronous, wang2023real} for real-world robot experiments. Figure~\ref{fig:ur5} depicts the experimental setup. 
This task involves using a camera to guide a UR5 robotic arm to reach a red target on a monitor. 
We ensure that the arm stays within a safe bounding box to avoid collisions.
The available actions are represented by the desired angular velocities for five\footnote{We do not move the sixth joint because its sole purpose is to control a gripper.} 
joints, ranging from $-0.7$ to $0.7$ radians per second.
The full observation that the learning agent receives includes three consecutive RGB images from the Logitech camera of dimensions $160 \times 90$ pixels, joint angles, joint angular velocities, and the previous action taken.
The reward function is defined as follows \citep{yuan2022asynchronous}:
\begin{equation*}
    r_t= \frac{c}{hw} M_t \odot W   
\end{equation*}
where $c$ is a scaling coefficient, $h$ and $w$ are the height and width of the image in pixels, $M_t$ is a binary mask\footnote{Note that this is a preprogrammed mask based on RGB thresholds, not made by MaDi.} of shape $h \times w$ that detects for each pixel whether it is currently red, and $W$ is a weighting matrix of shape $h \times w$ with values decreasing from $1$ at its center to $0$ near the edges.
These are multiplied element-wise by the Hadamard product $\odot$.
The reward incentivizes the robot to move its camera closer to the target and keep the target at the center of the frame. We set the coefficient $c = 800$ for all experiments and clip the rewards to be between $0$ and $4$.
An episode lasts $150$ timesteps of $40$ ms each, for $6$ seconds in total. The agent sends an action at every timestep, which is repeated by the SenceAct system five times at every $8$ ms.

\paragraph{Real-time asynchronous learning} We use the ReLoD system \citep{wang2023real} to facilitate effective, real-time learning. 
In simulation, environments can be internally paused while agents carry out learning updates. 
However, in the real world, the environment does not wait for the agent to complete its sequential computations and learning updates. 
We use an asynchronous implementation of SAC \citep{yuan2022asynchronous, wang2023real} to gather data in real-time and implement learning updates through a separate process. We reimplemented MaDi and all baselines used in these experiments to their asynchronous versions.
Each learning update can take anywhere from 70 to 500 ms, depending on the chosen algorithm and hyperparameter configuration, meaning that there will be fewer updates than steps in the environment.

\paragraph{Testing environment}
We test the generalization performance on a similar task, but now with videos playing in the background on the screen. We define this new generalization benchmark as UR5-VisualReacher-VideoBackgrounds (see Figure~\ref{fig:subfig2-ur5-test-env}), selecting five videos from the DMControl-GB \citep{hansen2021soda} to use as backgrounds. We test on this benchmark after every 4500 timesteps in the training environment. This evaluation is always done for $10$ episodes, twice on each video. See Appendix~\ref{app:task-descriptions} for examples of frames.

\paragraph{Baselines}
We compare the performance of SAC \citep{sac}, RAD \citep{laskin2020rad}, SVEA \citep{hansen2021svea}, and MaDi. 
The base architecture is the same for all algorithms, but it differs somewhat from the simulation experiments of Section~\ref{sec:experiments}. See Appendix~\ref{app:implementation_details} for more details.

\subsection{Results}

\begin{figure}
    \centering
    \includegraphics[width=\columnwidth]{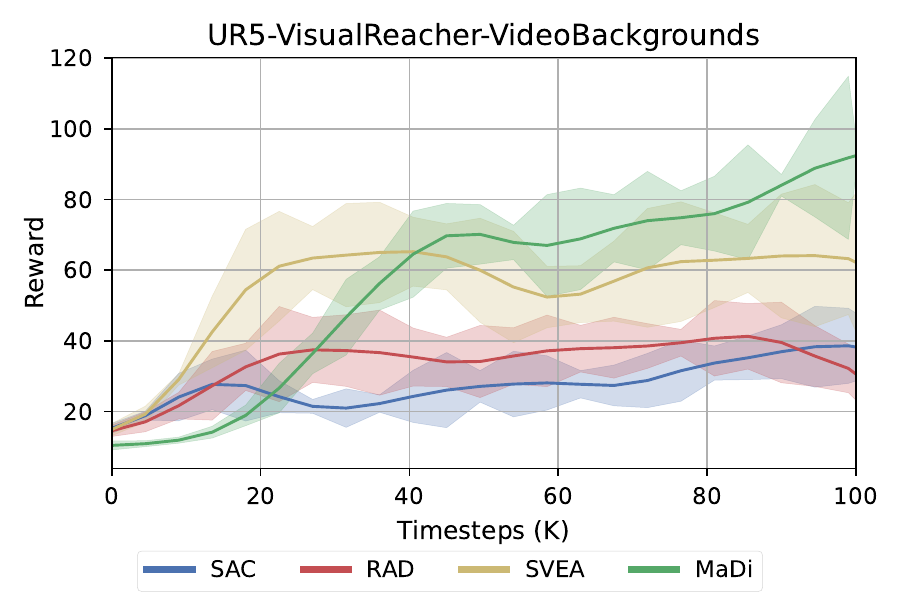}
    \caption{Performance of MaDi and three baselines on the UR5-VisualReacher task with random videos playing in the background. Agents are trained on the clean environment for $100K$ steps. We display the mean and standard error over five seeds. MaDi already generalizes best after $50K$ timesteps.} 
    \label{fig:ur5-eval}
    \Description{This figure presents a graph showcasing the performance of MaDi and three other baselines on the UR5-VisualReacher-VideoBackgrounds task. The vertical y-axis, labeled 'Reward', ranges from 0 to 120, while the horizontal x-axis, labeled 'Timesteps (K)', spans from 0 to 100. Four distinct learning curves are displayed, each represented by a colored line: SAC (blue), RAD (red), SVEA (yellow), and MaDi (green). The curves represent the performance of each model over time, with areas of uncertainty around each curve shaded in their respective colors. The MaDi curve is consistently higher in terms of reward compared to the other three after 40K timesteps, indicating superior performance. It reaches about 90 in the end, while others finish at: SVEA 60, SAC 40, RAD 35. SVEA's curve does go up sooner than MaDi's, but plateaus around 60 after 20K steps.}
\end{figure}

In Figure~\ref{fig:ur5-eval}, we present the results on the testing environment with video backgrounds. Without being trained on these distracting visuals, MaDi is able to generalize well in this challenging task.

The algorithms are trained asynchronously in this robotic environment, which means that MaDi will make fewer updates as it uses the additional Masker network. However, as Figure~\ref{fig:ur5-eval} shows, this only incurs a small delay in learning while gaining superior generalization capability through the increased focus on task-relevant pixels. A similar pattern is present on the original training environment, as shown in Figure~\ref{fig:ur5-train} of Appendix~\ref{app:additional_results:train_ur5}.

In these experiments, both MaDi and SVEA use the \texttt{overlay} augmentation. See Appendix~\ref{app:ur5_conv} for results showing the effect of the \texttt{conv} augmentation in this environment. MaDi also outperforms the baselines with the \texttt{conv} augmentation but scores slightly lower, despite the arguably clearer masks it generates in that setting.

In Appendix~\ref{app:additional_results:ur5_sparse}, we present an additional experiment with a \textit{sparse reward} function. In this UR5-VisualReacher-SparseRewards task, the agent is only rewarded with a $+1$ whenever its camera is close enough to the red target (according to a predefined threshold), and receives zero reward otherwise. Even in this challenging setting, MaDi is able to surpass the baselines in generalization performance.

\subsection{Analysis}

\begin{figure}
    \centering
    \vspace{0.5em}
    \includegraphics[width=0.95\linewidth]{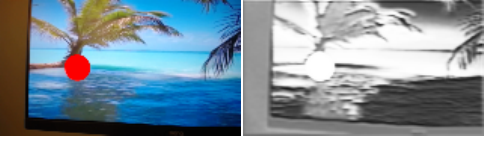}
    \vspace{-0.5em}
    \caption{Observation and the corresponding mask generated by MaDi in the UR5-VisualReacher-VideoBackgrounds task. The mask is subtle, but clearly leaves the red dot's pixels intact while dimming other areas of the frame. See Appendix~\ref{app:mask-examples} for more examples of masks.}
    \label{fig:ur5-masks}
    \vspace{-0.4em}
    \Description{This figure presents side-by-side images illustrating an observation and the corresponding mask generated by MaDi for the UR5-VisualReacher-VideoBackgrounds task.
The first image displays a colorful image of an ocean with palm trees and a red dot located over the scene near the horizon. The adjacent image on the right shows a grayscale mask that highlights the area where the red dot is, making it a bright spot, with other parts of the image, including the beach and sky, appearing more subdued.}    
\end{figure}

In Figure~\ref{fig:ur5-masks} we show that MaDi can learn to recognize the task-relevant features even in this real-world robotic task. It is able to generalize to the unseen testing environment and produce helpful masks. There seem to be fewer pixels dimmed in this environment compared to the other benchmarks, which may be because it can be useful for the agent to know where the entire screen is positioned. 

\begin{table}
\centering
\caption{The average reward per timestep on the UR5-VisualReacher task during the last 10\% of steps in an episode. MaDi receives higher rewards in the final position of an episode in both training and testing environments, showing that it finds the red target with higher accuracy.}
\vspace{-0.3em}
\label{tab:last-step-rew}
\resizebox{\columnwidth}{!}{%
\begin{tabular}{lllll}
\toprule
Reward per step & SAC  & RAD  & SVEA  & MaDi \\
\midrule
Training env. & $1.38\ \scriptstyle{\pm0.09}$ & $1.32\ \scriptstyle{\pm0.11}$ & $1.18\ \scriptstyle{\pm0.37}$ & $1.95\ \scriptstyle{\pm0.09}$ \\
Testing env. & $0.32\ \scriptstyle{\pm0.04}$ & $0.24\ \scriptstyle{\pm0.07}$ & $0.47\ \scriptstyle{\pm0.14}$ & $0.74\ \scriptstyle{\pm0.07}$ \\
\bottomrule
\end{tabular}
}
\end{table}

\paragraph{Distinction between training and testing}
MaDi performs well in both the training and test environments, but there is quite a large discrepancy between the total rewards received. For all algorithms, the reward in the testing environment is substantially lower than in the training environment. Taking a qualitative look at the behavior of the robotic arm, it seems this is mostly due to the fact that the robotic arm moves slower toward the target in the testing environment than in the training environment.
The agents encounter unseen observations that significantly deviate from the training environment, likely driving them to select different actions that do not match well in sequence, causing the arm to slow down.
The SAC and RAD agents rarely complete the task at all when there are videos playing in the background.

Even though the movement towards the goal is slower in the testing environment for all algorithms, MaDi does often reach a (near) optimal state at the end of its trajectory, similar to training. 
See Table~\ref{tab:last-step-rew} for an overview of the rewards in the last 10\% of steps for each baseline. MaDi shows a higher accuracy in finding the red target near the end of an episode.\footnote{Moreover, MaDi generally moves to the target faster, see \href{https://youtu.be/TQMazg6dntE}{youtu.be/TQMazg6dntE}.}

\section{Conclusion}
\label{sec:conclusion}

In the domain of vision-based deep reinforcement learning, we formalize the problem setting of distracting task-irrelevant features. We propose a novel method, MaDi, which learns directly from the reward signal to \textit{ma}sk \textit{di}stractions with a lightweight Masker network, without requiring any additional segmentation labels or loss functions.
Our experiments show that MaDi is competitive with state-of-the-art algorithms on the DeepMind Control Generalization Benchmark and the Distracting Control Suite, while only using $0.2\%$ additional parameters. 
The masks generated by MaDi enhance the agent's focus by dimming visual distractions. Even in the sparse reward setting, the Masker network is able to learn where the task-relevant pixels are in each state.
Furthermore, we test MaDi on a real UR5 Robotic Arm, showing that it can outperform the baselines not only in simulation environments but also on our newly defined UR5-VisualReacher-VideoBackgrounds generalization benchmark.

\paragraph{Limitations \& Future Work.}

The algorithms in this work build on the model-free off-policy deep RL algorithm SAC, while other options remain open for investigation. In future work, we seek to apply MaDi to other reinforcement learning algorithms such as PPO \citep{schulman2017ppo} or DQN \citep{mnih2015human} and improve the clarity of its masks.
We have experimented with MaDi on one robotic arm, it would be interesting to see whether the Masker network can also produce useful masks on a diverse set of robots.
Lastly, in future research, we aim to explore the possibility of using MaDi for transfer learning.





\section*{Ethics statement}

Our research aims to contribute to the development of reinforcement learning algorithms to facilitate their application in practical scenarios that positively impact society. 
We believe our work with MaDi has the potential to contribute to, for instance, enhancing a household robot's ability to focus on relevant visual information amidst a clutter of distractions.
Furthermore, we aim to minimize the computational footprint of our algorithms by adding a negligible amount of parameters to the original structure, supporting the development of sustainable and energy-efficient AI.




\begin{acks}
This work is part of the AMADeuS project of the Open Technology Programme (project number 18489), which is partly financed by the Dutch Research Council (NWO).
This research used the Dutch national e-infrastructure with the support of the SURF Cooperative, using grant number EINF-7063.
Part of this work has taken place in the Intelligent Robot Learning (IRL) Lab at the University of Alberta, which is supported in part by research grants from the Alberta Machine Intelligence Institute (Amii); a Canada CIFAR AI Chair, Amii; Compute Canada; Huawei; Mitacs; and NSERC. 
Furthermore, this research was partly funded by the RLAI lab and the Ocado Group's donation of the UR5 arm.
We are grateful to Fahim Shahriar and Homayoon Farrahi for their advice and assistance in the RLAI Robotics lab, as well as Christopher Gadzinski and \mbox{Calarina} Muslimani for their detailed feedback on the writing. 
\mbox{Finally}, we thank all anonymous reviewers who helped to enhance the quality of our paper.
\end{acks}



\balance

\bibliographystyle{ACM-Reference-Format} 
\bibliography{citations}

\newpage
\onecolumn
\appendix
\section*{Appendix}

\vspace{1em}
\section{Implementation Details}
\label{app:implementation_details}

We provide further details on our implementation of MaDi. Table~\ref{tab:hyper} presents an overview of the hyperparameters used in our study.
For the simulation experiments we used the default hyperparameters as specified by the DMControl Generalization Benchmark \citep{hansen2021soda} and the particular baselines. For the robotic task, we obtained hyperparameters from \citep{wang2023real} while aiming to keep the majority the same.

\begin{table}[H]
\centering
\caption{Overview of the hyperparameters used in our experiments.}
\label{tab:hyper}
\vspace{-0.7em}
\begin{tabular}{lll}\toprule
Hyperparameter                                        & Simulation experiments        & Robotic experiments (if different)  \\ \midrule
\textit{Shared by all algorithms}                     &                               &         \\ 
\quad optimizer                                       & Adam \cite{adam}              &         \\ 
\quad learning rate (actor and critic)                & $10^{-3}$                     & $3\cdot 10^{-4}$   \\
\quad Adam $\beta_1$, $\beta_2$ (all networks)        & $0.9$, $0.999$                &         \\
\quad discount ($\gamma$)                             & $0.99$                        &         \\ 
\quad frame stack                                     & $3$                           &         \\
\quad action repeat                                   & $4$                           & $5$     \\
\quad batch size                                      & $128$                         &         \\
\quad initial collect steps                           & $1000$                        &         \\
\quad replay buffer size                              & $500$K                        & $100$K  \\
\quad environment steps                               & $500$K                        & $100$K  \\
\quad train every $k_c$ env steps (critic), $k_c=$    & $1$                           & asynchronous \\ 
\quad train every $k_a$ env steps (actor), $k_a=$     & $2$                           & once every critic update \\
\quad gradient steps per training step                & $1$                           &          \\
\quad target update interval ($k_{tar}$)              & $2$                           & once every critic update \\  
\quad target smoothing coefficient critic ($\tau$)    & $0.01$                        & $0.005$       \\
\quad target smoothing coefficient encoder ($\tau_{\text{enc}}$)  & $0.05$            & $0.005$       \\
\quad initial temperature ($\alpha$ in SAC)           & $0.1$                         &    \\
\quad learning rate (for temperature $\alpha$)        & $10^{-4}$                     & $3\cdot 10^{-4}$    \\
\quad Adam $\beta_1$, $\beta_2$ (for temperature $\alpha$)   & $0.5$, $0.999$         &      \\
\midrule
\textit{DrQ} \citep{drq}                              &                               &      \\
\quad K, M ($Q$-target, $Q$-function averaging)       & $1$, $1$                      &      \\
\quad random shift                                    & Up to $\pm 4$ pixels          &      \\
\midrule
\textit{RAD} \citep{laskin2020rad}                    &                               &      \\
\quad random crop                                     & $100\times100 \longrightarrow 84\times84$ &  $160\times90 \longrightarrow 156\times88$   \\
\midrule
\textit{SODA} \citep{hansen2021soda}                  &                               &      \\
\quad learning rate auxiliary network                 & $10^{-3}$                     &      \\
\quad batch size auxiliary network                    & $256$                         &      \\
\quad target smoothing coefficient aux net ($\tau_{\text{aux}}$)  & $0.005$           &      \\ 
\quad train every $k_{\text{aux}}$ env steps (aux), $k_{\text{aux}}=$  & $2$          &      \\
\midrule
\textit{SVEA} \citep{hansen2021svea}                  &                               &       \\
\quad svea\_alpha, svea\_beta                         & $0.5$, $0.5$                  &       \\
\midrule
\textit{SGQN} \citep{bertoin2022sgqn}                 &                               &       \\
\quad threshold \texttt{sgqn\_quantile}\protect\footnotemark    & $0.95$ or $0.98$    &       \\
\quad weight decay critic                             & $10^{-5}$                     &        \\
\quad learning rate SGQN head                         & $3 \cdot 10^{-4}$             &         \\
\quad train every $k_{\text{sgqn}}$ env steps (SGQN head), $k_{\text{sgqn}}=$  & $2$  &          \\
\midrule
\textit{MaDi}                                         &                               &       \\
\quad learning rate masker network                    & $10^{-3}$                     & $3\cdot 10^{-4}$  \\
\quad train every $k_{\text{m}}$ env steps (masker), $k_{\text{m}}=$  & $1$           &       \\
\bottomrule         
\end{tabular}
\end{table}

\footnotetext{The \texttt{sgqn\_quantile} is set to $0.95$ for domains with large agents (e.g., \texttt{walker} and \texttt{finger}), and $0.98$ for domains with smaller agents (e.g., \texttt{cartpole} and \texttt{ball\_in\_cup}).}



\paragraph{Network Architecture.}
All the algorithms in our experiments use the same base architecture, acquired from the DMControl Generalization Benchmark \citep{hansen2021soda}, to ensure a fair comparison. 
This base architecture (yellow modules in Figure~\ref{fig:overview}) consists of:
\begin{itemize}
    \item \textbf{Encoder:} a shared $11$-layer ConvNet that takes a stack of $3$ RGB frames (shape $[9,84,84]$) and outputs spatial features of size $[32,21,21]$. Each layer has $32$ channels with a kernel size of $3\times3$, a stride of $1$ and no padding. Only the first layer has a stride of $2$, to reduce the channel size quicker. We use ReLU activations between all convolutional layers.
    \item \textbf{Actor:} a multi-layer perceptron (MLP) that outputs actions. It first projects the $32\cdot21\cdot21 = 14,112$ flattened features to a vector of just $100$ through a large MLP layer. We then apply LayerNorm \citep{ba2016layer} and a Tanh activation function. The bulk of the actor network follows, with three MLP layers of hidden dimension $1024$ with ReLU activations in between. The output layer has twice the size of the action dimension, to output means $\mu$ and standard deviations $\sigma$.
    \item \textbf{Critic:} two independent MLPs that output state-action values. Both have the same architecture as the actor. The large projection (up to and including the Tanh activation) is shared between the two critics. The critics receive the current action $\bm{a}$ as input as well, this is concatenated to the output of the large projection. Their last layers only have one output neuron, determining the $Q$-value estimate.
\end{itemize}

\noindent
For the robotic experiments the architectures are a bit different:
\begin{itemize}
    \item \textbf{Encoder:} a shared $4$-layer ConvNet that takes a stack of $3$ RGB frames (shape $[9,90,160]$), followed by a spatial softmax layer \citep{finn2016deep}. We use $32$ channels in each layer, kernel size is $3\times3$, no padding, and stride $2$ on all layers except the last one, which uses $1$. ReLU activation functions between the layers. 
    The encoder outputs a vector of length $64$.  
    \item \textbf{Actor:} an MLP that outputs actions. It first concatenates the encoder features with a proprioception vector of length $15$, which consists of the angles, angular velocities, and previous actions of each of the five joints that we control. (The sixth joint is for a gripper, which we do not use.) After that, two MLP layers of hidden dimension $512$ follow. 
    \item \textbf{Critic:} two independent MLPs that output state-action values. Both have the same architecture as the actor, except again the concatenation of the action $\bm{a}$ in the input and an output layer of dimension $1$.
\end{itemize}

\begin{wrapfigure}[5]{r}{0.4\textwidth}
    \centering
    \includegraphics[width=0.38\textwidth]{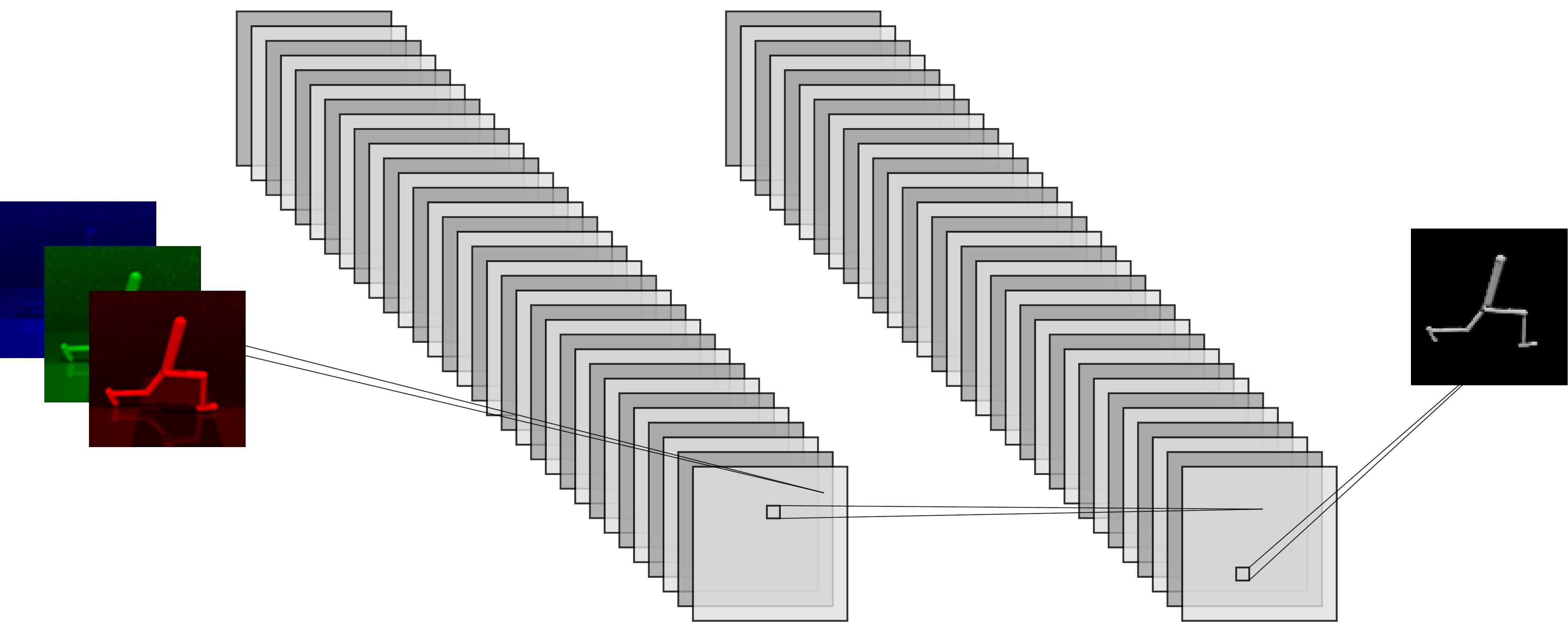}
    \vspace{-0.5em}
    \caption{The Masker network architecture contains two hidden layers with 32 channels and ReLU activations, while the output layer produces just 1 channel with a Sigmoid activation.} 
    \label{fig:maskernet}
    \Description{The figure visually represents the Masker network architecture. On the left, there's a small square divided into three colored sections: red, green, and blue, representing the RGB channels of the walker-walk observation image. Two lines emerge from this square, leading to a series of vertically stacked rectangles, illustrating the hidden convolutional layers. These layers are depicted with repetitive patterns, signifying multiple channels - specifically 32.
    Following these layers, another set of lines coming from a small square (that represents the kernel) guides us to a single square (channel) on the far right, symbolizing the output layer that produces just 1 channel. This output square contains a white silhouette of a walker-walk figure. The overall diagram conveys the process of the input RGB channels passing through the hidden layers and being processed to a single output.}
\end{wrapfigure}

\noindent
And of course, the critic network has an exact replica, its \textit{target network}, following the SAC \citep{sac} design.
The algorithms \textbf{SAC}, \textbf{DrQ}, \textbf{RAD}, and \textbf{SVEA} all use this exact architecture. The other methods include an additional component:
\begin{itemize}
    \item \textbf{SODA} has an auxiliary network that tries to minimize the distance of augmented and non-augmented images in their feature space.
    \item \textbf{SGQN} comes with another head attached to the critic's large projection. This auxiliary component tries to minimize the difference between the masks it outputs and the masks generated by a saliency metric.
    \item \textbf{MaDi} introduces the Masker network at the start of the architecture. A small ConvNet comprising 3 layers, each with a kernel size of $3$x$3$, stride $1$, and padding $1$ (with value $0$). Each layer is followed by a ReLU activation, except for the output layer, which applies a Sigmoid activation function to ensure the generated mask values are between 0 and 1. See Figure~\ref{fig:maskernet}.
\end{itemize}

\paragraph{Mask Application.}
In MaDi, the mask generated by the Masker is applied to the visual input by element-wise multiplication before it is passed to the actor and critic networks. 
Masks are generated and applied per frame, but we only need one forward pass through the Masker to get masks for all 3 frames of a stacked input observation. This is shown in the following code snippet:

\vspace{1em}
\noindent\hspace*{0.32cm}%
\begin{minipage}{\dimexpr\textwidth-1cm}
\begin{codesnippet}
def apply_mask(self, obs):  # obs shape: (B,9,H,W)
    """Apply masks generated by one forward pass of the Masker""" 

    # split the stacked frames
    frames = obs.chunk(3, dim=1)  # [(B,3,H,W), (B,3,H,W), (B,3,H,W)]

    # concat in batch dim
    frames_cat = torch.cat(frames, dim=0)  # (3B,3,H,W)

    # one forward pass through the Masker network
    masks_cat = self.masker(frames_cat)  # (3B,1,H,W)

    # split the batch dim back
    masks = masks_cat.chunk(3, dim=0)  # [(B,1,H,W), (B,1,H,W), (B,1,H,W)]

    # element-wise multiplication 
    # uses broadcasting over the 3 RGB channels within 1 frame
    masked_frames = [m*f for m,f in zip(masks,frames)]  # [(B,3,H,W), (B,3,H,W), (B,3,H,W)]

    # concat in channel dim
    return torch.cat(masked_frames, dim=1)  # (B,9,H,W)
    # 
\end{codesnippet}
\end{minipage}


\subsection{Vision Transformer implementation}
\label{app:vit-implementation}

The design of our Vision Transformer \citep[ViT]{dosovitskiy2020image} experiments closely follows the setup of SVEA \citep{hansen2021svea}, where the ConvNet encoder of Figure~\ref{fig:overview} is replaced by a ViT encoder, while introducing minimal changes to the rest of the architecture.
For this ViT encoder, we use RGB frames with dimensions $96\times96\times3$ as input instead of the $84\times84\times3$ used for the ConvNet. From this, we extract a total of $144$ non-overlapping $8\times8\times3k$ image patches (with $k$ being the frame count in a stacked observation as described in Section~\ref{sec:preliminaries}). Each patch is mapped to a $128$-dimensional embedding, which becomes a token to be processed by the ViT encoder. Consistent with the original ViT design, our model incorporates learned positional encodings and a modifiable \texttt{class} token. The architecture comprises four Transformer \citep{vaswani2017attention} blocks. Each block applies Multi-Head Attention encompassing eight heads, after which an MLP layer with hidden dimension $128$ and a GELU \citep{hendrycks2016gelu} activation follows.
We optimize the ViT encoder with the critic loss, just like the original ConvNet. We maintain the same optimizer (Adam) and learning rates as in Table~\ref{tab:hyper}, but we double the batch size to $256$. 
Our ViT encoder starts without any pre-trained parameters, and we do not use weight decay. The entire training process spans around three days on a single NVIDIA RTX A4000 GPU. 


\vphantom{X}



\section{Additional Results}
\label{app:additional_results}

This section provides additional results from our experiments, demonstrating the performance of MaDi in the original training environments and various other environments with differing levels of distractions. The table with results on \texttt{video\_easy} from the DMControl Generalization Benchmark \citep{hansen2021soda} are already presented in Table~\ref{tab:results-video_easy} of the main text, see Section~\ref{sec:gen-results-vid-hard}.
Here we show the results on the original training environments in Sections~\ref{app:additional_results:train_ur5} and \ref{app:additional_results:train_dmcontrol}, as well as evaluations on \texttt{video\_hard} in Section~\ref{app:additional_results:hard}, and the Distracting Control Suite in Section~\ref{app:additional_results:dcs}. 

\paragraph{Statistical tests}
We verify the statistical significance of the results in our tables by running the unequal variances $t$-test (also called Welch's $t$-test) between MaDi and the best baseline for each environment. This test does not assume similar variances, unlike the Student's $t$-test. All our tests are two-sided, and we consider p-values less than $0.05$ as statistically significant.


\paragraph{Training directly on distractions}
Interestingly enough, training directly on the challenging environments with video backgrounds does not work well. In our preliminary experiments, this holds for all algorithms in both the simulation and robotic environments. Some clean data seems necessary to learn an adequate policy, before agents can learn to deal with distractions. 
Finding the ideal curriculum could be an interesting direction for future work.

\subsection{Results on training environment UR5 Robotic Arm}
\label{app:additional_results:train_ur5}

\begin{figure}[H]
    \centering
    \captionsetup{width=0.75\textwidth}
    \includegraphics[width=0.6\textwidth]{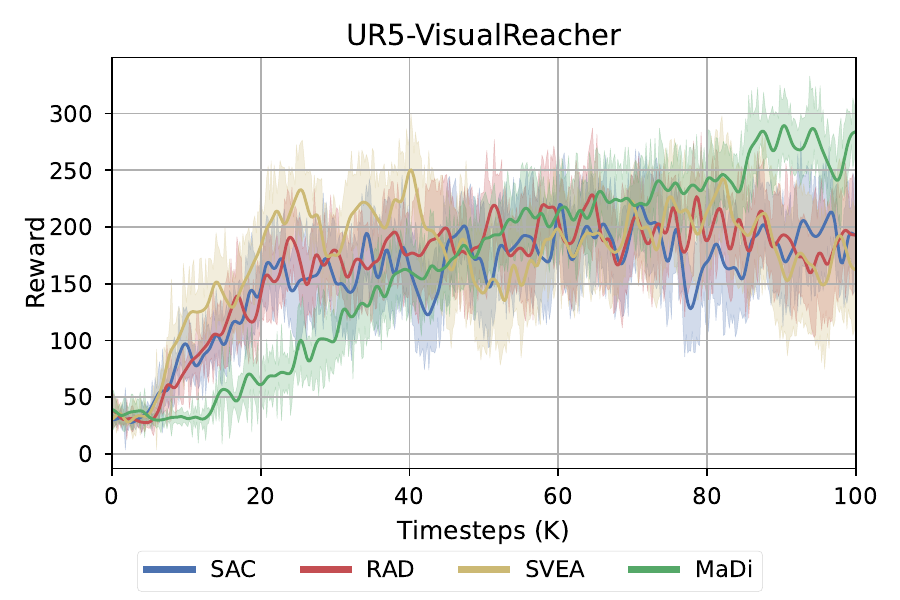}
    \vspace{-0.5em}
    \caption{Performance of MaDi and three baselines on the UR5-VisualReacher training environment. Agents are trained on this clean setup (white background with the red dot) for $100K$ steps. Curves show the mean over five seeds with standard error shaded alongside. Results on the generalization task are shown in Figure~\ref{fig:ur5-eval}.}
    \label{fig:ur5-train}
\Description{This figure showcases a graph detailing the performance of MaDi and three other baselines on the UR5-VisualReacher task. The vertical y-axis, marked 'Reward', ranges from 0 to a bit above 300, while the horizontal x-axis, labeled 'Timesteps (K)', spans from 0 to 100. There are four distinct learning curves, each represented by a unique colored line: SAC (blue), RAD (red), SVEA (yellow), and MaDi (green). The performance of each model is tracked over time, with areas of uncertainty surrounding each curve highlighted with a lighter shade of the respective color. Over the span of 100K timesteps, MaDi's performance generally trends upward, achieving a reward of around 280. Meanwhile, the other models a learn a bit quicker in the beginning, but plateau around a value of 190. MaDi surpasses the other three around halfway, at 50K steps.}
\end{figure}

In the UR5 Robotic Arm experiments, the algorithms are trained asynchronously. This means that MaDi makes fewer updates than the others, as it takes some extra time to use the Masker network. However, even with fewer updates it surpasses the performance of other baselines on the training task; see Figure~\ref{fig:ur5-train}. The corresponding results on the generalization task are presented in Section~\ref{sec:robotic-exps}, Figure~\ref{fig:ur5-eval}.

\vspace{-1em}

\newpage
\subsection{Results on UR5 Robotic Arm with \textit{sparse} rewards}
\label{app:additional_results:ur5_sparse}

\begin{figure}[H]
    \centering
    \begin{subfigure}[b]{0.48\textwidth}
        \includegraphics[width=\textwidth]{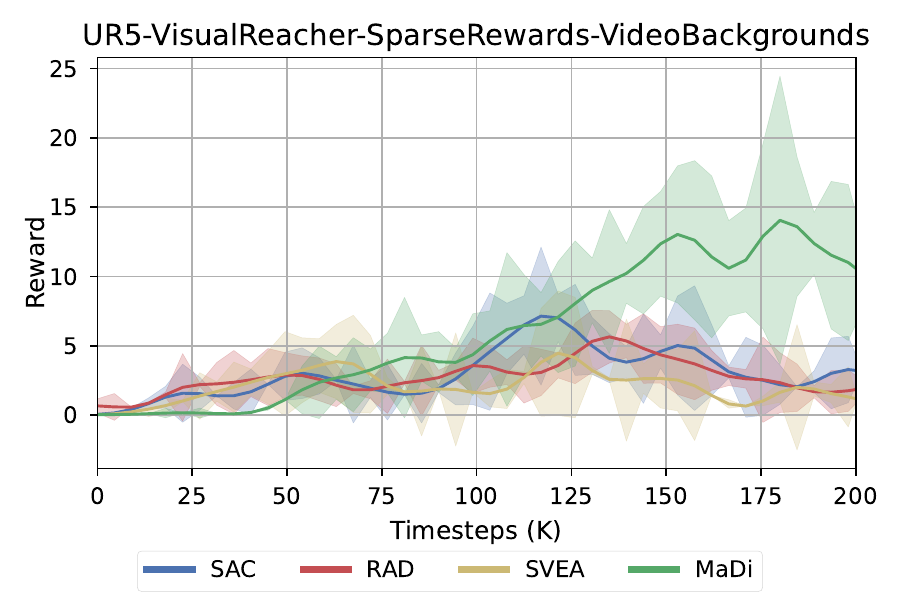}
        \caption{testing environment}
        \vspace{-0.5em}
        \label{fig:subfig1-ur5sparse}
    \end{subfigure}
    \hfill
    \begin{subfigure}[b]{0.48\textwidth}
        \includegraphics[width=\textwidth]{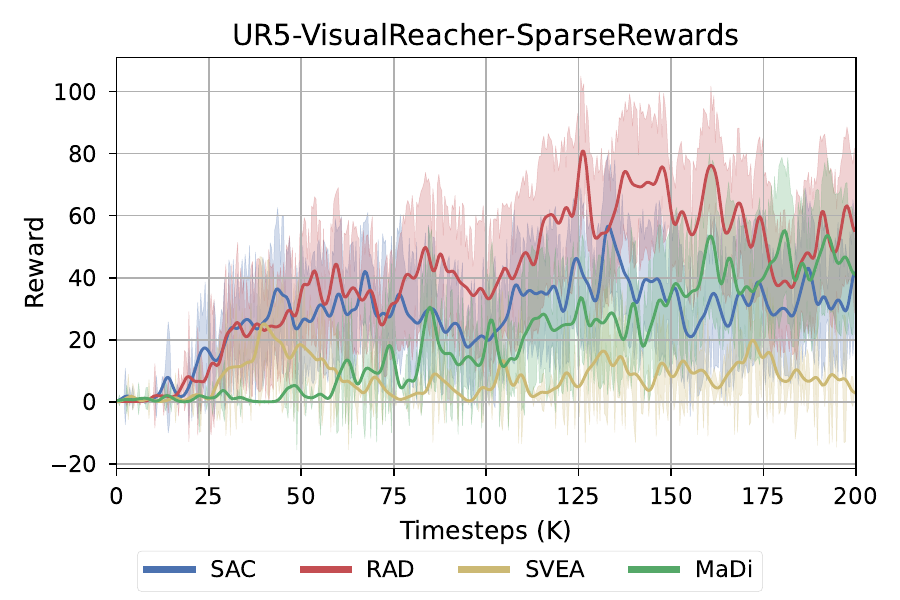}
        \caption{training environment}
        \vspace{-0.5em}
        \label{fig:subfig2-ur5sparse}
    \end{subfigure}
    \captionsetup{width=0.97\textwidth}
    \caption{Performance of MaDi and three baselines on the \textit{sparse reward} version of the UR5-VisualReacher environment. Agents are trained on the clean setup (white background with the red dot) for $200K$ steps. Curves show the mean over five seeds with standard error shaded alongside.}
    \label{fig:ur5_sparse_graphs}
\Description{The figure comprises two side-by-side line graphs, each illustrating the performance of four algorithms: SAC (in blue), RAD (in red), SVEA (in yellow), and MaDi (in green) over time. The x-axis for both graphs represents 'Timesteps (K)' from 0 to 200, and the y-axis denotes 'Reward'.
On the left, figure (a) Testing environment, titled "UR5-VisualReacher-SparseRewards-VideoBackgrounds":
The graph displays the performance of the four algorithms when the agent is in the testing environment, which includes video backgrounds. The reward range is between 0 and 25. The curves for all four algorithms begin near zero. SAC, RAD and SVEA barely increase, never going above 6. MaDi grows to a score of about 12 after 150K timesteps.
On the right, figure (b) Training environment, titled "UR5-VisualReacher-SparseRewards":
This graph demonstrates the performance of the four algorithms in the original training environment. The reward range here is between 0 and 100. While all curves start at 0 reward value, RAD and SAC both show consistent and steady improvement up to 60. SVEA, however, appears to face challenges and remains relatively flat around 10 throughout, with some fluctuations. MaDi, on the other hand, grows slower but displays a consistent upward trend, with its performance curve moving to around 50 after 150K steps.
In both graphs, the shaded regions around each curve indicate the standard error based on five seeds. The overall representation conveys the relative performance of each algorithm in two distinct environments: one with video backgrounds (testing) and one without (training).}
\end{figure}

In the \textit{sparse reward} setting of the UR5-VisualReacher task, the agent is only rewarded with a $+1$ whenever its camera is close enough to the red target (according to a predefined threshold). For all other situations, it receives zero reward. Even in this more challenging setting, MaDi is able to outperform the baselines in terms of generalization, as shown in Figure~\ref{fig:subfig1-ur5sparse}. On the training environment, presented in Figure~\ref{fig:subfig2-ur5sparse}, MaDi learns a bit slower but eventually reaches similar performance to SAC and RAD. SVEA seems to struggle a lot with sparse rewards. The MaDi and SVEA agents both use the default \texttt{overlay} augmentation here.

\subsection{Results on training environment DMControl}
\label{app:additional_results:train_dmcontrol}

The results of MaDi and multiple baselines on the original training environments of the DeepMind Control Suite are shown in Table~\ref{tab:results-clean}. The best performance is achieved by DrQ, an algorithm that only uses light augmentations (random shifts). The stronger augmentations that SODA, SVEA, SGQN, and MaDi use do not improve the results on the training environment, as they are specifically designed to enhance generalization to other domains. MaDi does still come close to the state-of-the-art in the training environments as well.

    \begin{table}[H]
    \captionsetup{width=0.6\textwidth}
    \caption{Performance on the original training environments of MaDi and various baseline algorithms after training for $500K$ steps. Showing mean undiscounted return and standard error over five seeds. MaDi is competitive (no significant difference) with the best baseline in 5/6 cases, denoted by $\star$.}
    \label{tab:results-clean}
    \vspace{-0.5em}
    \centering
    \begin{tabular}{lcccccccl}
\toprule
\texttt{clean}  & SAC & DrQ & RAD & SODA & SVEA & SGQN & MaDi & \texttt{p-value} \\
\midrule 
\texttt{ball\_in\_cup} & $727$ & $\bm{973}$ & $651$ & $965$ & $811$ & $778$ & $808$  & $0.313^{\star}$  \vspace{-0.75ex} \\
\texttt{catch} & $\scriptstyle{\pm117}$ &$\bm{\scriptstyle{\pm4}}$& $\scriptstyle{\pm178}$ & $\scriptstyle{\pm4}$ & $\scriptstyle{\pm144}$ & $\scriptstyle{\pm174}$ & $\scriptstyle{\pm144}$  &  \vspace{0.75ex} \\
\texttt{cartpole} & $\bm{996}$ & $993$ & $985$ & $981$ & $978$ & $966$ & $986$  & $0.024$  \vspace{-0.75ex} \\
\texttt{balance} &$\bm{\scriptstyle{\pm1}}$& $\scriptstyle{\pm3}$ & $\scriptstyle{\pm4}$ & $\scriptstyle{\pm4}$ & $\scriptstyle{\pm2}$ & $\scriptstyle{\pm7}$ & $\scriptstyle{\pm3}$  &  \vspace{0.75ex} \\
\texttt{cartpole} & $861$ & $\bm{865}$ & $861$ & $230$ & $842$ & $806$ & $851$  & $0.077^{\star}$  \vspace{-0.75ex} \\
\texttt{swingup} & $\scriptstyle{\pm8}$ &$\bm{\scriptstyle{\pm4}}$& $\scriptstyle{\pm8}$ & $\scriptstyle{\pm142}$ & $\scriptstyle{\pm8}$ & $\scriptstyle{\pm14}$ & $\scriptstyle{\pm5}$  &  \vspace{0.75ex} \\
\texttt{finger} & $324$ & $643$ & $593$ & $493$ & $\bm{711}$ & $597$ & $681$  & $0.626^{\star}$  \vspace{-0.75ex} \\
\texttt{spin} & $\scriptstyle{\pm9}$ & $\scriptstyle{\pm36}$ & $\scriptstyle{\pm13}$ & $\scriptstyle{\pm117}$ &$\bm{\scriptstyle{\pm55}}$& $\scriptstyle{\pm13}$ & $\scriptstyle{\pm17}$  &  \vspace{0.75ex} \\
\texttt{walker} & $580$ & $\bm{979}$ & $965$ & $147$ & $973$ & $694$ & $972$  & $0.147^{\star}$  \vspace{-0.75ex} \\
\texttt{stand} & $\scriptstyle{\pm117}$ &$\bm{\scriptstyle{\pm4}}$& $\scriptstyle{\pm3}$ & $\scriptstyle{\pm17}$ & $\scriptstyle{\pm2}$ & $\scriptstyle{\pm144}$ & $\scriptstyle{\pm2}$  &  \vspace{0.75ex} \\
\texttt{walker} & $366$ & $892$ & $837$ & $496$ & $\bm{937}$ & $892$ & $893$  & $0.193^{\star}$  \vspace{-0.75ex} \\
\texttt{walk} & $\scriptstyle{\pm42}$ & $\scriptstyle{\pm33}$ & $\scriptstyle{\pm21}$ & $\scriptstyle{\pm173}$ &$\bm{\scriptstyle{\pm13}}$& $\scriptstyle{\pm27}$ & $\scriptstyle{\pm27}$  &  \\
\midrule
\texttt{avg} & $642$ & $891$ & $816$ & $552$ & $875$ & $789$ & $865$ &  \\
    \bottomrule
    \end{tabular}
    \end{table}

\subsection{Results on \texttt{video\_easy}}
\label{app:additional_results:video_easy}

The performance of MaDi and various other baseline algorithms in the \texttt{video\_easy} environment are presented in Table~\ref{tab:results-video_easy_app}. This is the same as Table~\ref{tab:results-video_easy} in the main body, but here we have enough space to include p-values of our statistical tests between MaDi and the best baseline. The \texttt{video\_easy} environment represents a scenario with a low level of visual distractions, making it a suitable benchmark to test the generalization performance of reinforcement learning algorithms in a still relatively clean environment.


\begin{table}[H]
    \captionsetup{width=0.6\textwidth}
    \caption{Generalization performance of MaDi and various baseline algorithms on six different environments after training for $500K$ steps. Evaluated on \texttt{video\_easy} from the DMControl-GB benchmark, showing mean and standard error over five seeds. MaDi significantly outperforms ($\star\star$) or is competitive to ($\star$) the state-of-the-art in 5/6 environments.}
    \label{tab:results-video_easy_app}
    \centering
    \begin{tabular}{lcccccccl}
\toprule
\texttt{video\_easy}  & SAC & DrQ & RAD & SODA & SVEA & SGQN & MaDi & \texttt{p-value} \\
\midrule 
\texttt{ball\_in\_cup} & $602$ & $714$ & $561$ & $750$ & $757$ & $761$ & $\bm{807}$  & $0.841^{\star}$  \vspace{-0.75ex} \\
\texttt{catch} & $\scriptstyle{\pm91}$ & $\scriptstyle{\pm131}$ & $\scriptstyle{\pm147}$ & $\scriptstyle{\pm98}$ & $\scriptstyle{\pm138}$ & $\scriptstyle{\pm171}$ &$\bm{\scriptstyle{\pm144}}$ &  \vspace{0.75ex} \\
\texttt{cartpole} & $924$ & $932$ & $801$ & $961$ & $967$ & $965$ & $\bm{982}$  & $0.011^{\star\star}$  \vspace{-0.75ex} \\
\texttt{balance} & $\scriptstyle{\pm19}$ & $\scriptstyle{\pm33}$ & $\scriptstyle{\pm95}$ & $\scriptstyle{\pm10}$ & $\scriptstyle{\pm2}$ & $\scriptstyle{\pm5}$ &$\bm{\scriptstyle{\pm4}}$ &  \vspace{0.75ex} \\
\texttt{cartpole} & $782$ & $613$ & $658$ & $215$ & $786$ & $798$ & $\bm{848}$  & $0.015^{\star\star}$  \vspace{-0.75ex} \\
\texttt{swingup} & $\scriptstyle{\pm21}$ & $\scriptstyle{\pm74}$ & $\scriptstyle{\pm17}$ & $\scriptstyle{\pm125}$ & $\scriptstyle{\pm15}$ & $\scriptstyle{\pm13}$ &$\bm{\scriptstyle{\pm6}}$ &  \vspace{0.75ex} \\
\texttt{finger} & $227$ & $543$ & $479$ & $429$ & $645$ & $592$ & $\bm{679}$  & $0.461^{\star}$  \vspace{-0.75ex} \\
\texttt{spin} & $\scriptstyle{\pm26}$ & $\scriptstyle{\pm50}$ & $\scriptstyle{\pm65}$ & $\scriptstyle{\pm100}$ & $\scriptstyle{\pm39}$ & $\scriptstyle{\pm11}$ &$\bm{\scriptstyle{\pm17}}$ &  \vspace{0.75ex} \\
\texttt{walker} & $507$ & $954$ & $961$ & $147$ & $\bm{977}$ & $672$ & $967$  & $0.041$  \vspace{-0.75ex} \\
\texttt{stand} & $\scriptstyle{\pm113}$ & $\scriptstyle{\pm10}$ & $\scriptstyle{\pm1}$ & $\scriptstyle{\pm17}$ &$\bm{\scriptstyle{\pm3}}$& $\scriptstyle{\pm153}$ & $\scriptstyle{\pm3}$  &  \vspace{0.75ex} \\
\texttt{walker} & $334$ & $821$ & $726$ & $479$ & $\bm{936}$ & $882$ & $895$  & $0.183^{\star}$  \vspace{-0.75ex} \\
\texttt{walk} & $\scriptstyle{\pm37}$ & $\scriptstyle{\pm38}$ & $\scriptstyle{\pm42}$ & $\scriptstyle{\pm168}$ &$\bm{\scriptstyle{\pm14}}$& $\scriptstyle{\pm26}$ & $\scriptstyle{\pm24}$  &  \\
\midrule
\texttt{avg} & $563$ & $763$ & $698$ & $497$ & $845$ & $778$ & $863$ &  \\
    \bottomrule
    \end{tabular}
    \end{table}

\subsection{Results on \texttt{video\_hard}}
\label{app:additional_results:hard}

The performance of MaDi and various other baseline algorithms in the \texttt{video\_hard} environment from the DMControl Generalization Benchmark \citep{hansen2021soda} are presented in Table~\ref{tab:results-video_hard}. The \texttt{video\_hard} environment represents a scenario with a higher level of visual distractions, making it a suitable benchmark to test the generalization performance of reinforcement learning algorithms in an environment full of distractions.
Similar to the results of Table~\ref{tab:results-video_easy} on \texttt{video\_easy}, MaDi also shows the best generalization capability overall on \texttt{video\_hard}.

\begin{table}[H]
    \captionsetup{width=0.6\textwidth}
    \caption{Generalization performance of MaDi and various baseline algorithms on six different environments trained for $500K$ steps. Evaluated on \texttt{video\_hard} from the DMControl-GB benchmark, showing mean undiscounted return and standard error over five seeds. MaDi significantly outperforms ($\star\star$) or is competitive to ($\star$) the state-of-the-art in 4/6 environments.}
    \label{tab:results-video_hard}
    \centering
    \begin{tabular}{lcccccccl}
\toprule
\texttt{video\_hard}  & SAC & DrQ & RAD & SODA & SVEA & SGQN & MaDi & \texttt{p-value} \\
\midrule 
\texttt{ball\_in\_cup} & $176$ & $180$ & $261$ & $194$ & $327$ & $687$ & $\bm{758}$  & $0.737^{\star}$  \vspace{-0.75ex} \\
\texttt{catch} & $\scriptstyle{\pm38}$ & $\scriptstyle{\pm51}$ & $\scriptstyle{\pm58}$ & $\scriptstyle{\pm39}$ & $\scriptstyle{\pm59}$ & $\scriptstyle{\pm155}$ &$\bm{\scriptstyle{\pm135}}$ &  \vspace{0.75ex} \\
\texttt{cartpole} & $314$ & $250$ & $263$ & $449$ & $579$ & $703$ & $\bm{827}$  & $0.005^{\star\star}$  \vspace{-0.75ex} \\
\texttt{balance} & $\scriptstyle{\pm12}$ & $\scriptstyle{\pm6}$ & $\scriptstyle{\pm18}$ & $\scriptstyle{\pm46}$ & $\scriptstyle{\pm26}$ & $\scriptstyle{\pm17}$ &$\bm{\scriptstyle{\pm25}}$ &  \vspace{0.75ex} \\
\texttt{cartpole} & $140$ & $155$ & $133$ & $172$ & $453$ & $488$ & $\bm{619}$  & $0.003^{\star\star}$  \vspace{-0.75ex} \\
\texttt{swingup} & $\scriptstyle{\pm10}$ & $\scriptstyle{\pm10}$ & $\scriptstyle{\pm19}$ & $\scriptstyle{\pm46}$ & $\scriptstyle{\pm26}$ & $\scriptstyle{\pm18}$ &$\bm{\scriptstyle{\pm24}}$ &  \vspace{0.75ex} \\
\texttt{finger} & $21$ & $16$ & $8$ & $107$ & $154$ & $\bm{554}$ & $358$  & $0.001$  \vspace{-0.75ex} \\
\texttt{spin} & $\scriptstyle{\pm4}$ & $\scriptstyle{\pm3}$ & $\scriptstyle{\pm2}$ & $\scriptstyle{\pm30}$ & $\scriptstyle{\pm31}$ &$\bm{\scriptstyle{\pm8}}$& $\scriptstyle{\pm25}$  &  \vspace{0.75ex} \\
\texttt{walker} & $233$ & $361$ & $343$ & $143$ & $847$ & $606$ & $\bm{920}$  & $0.012^{\star\star}$  \vspace{-0.75ex} \\
\texttt{stand} & $\scriptstyle{\pm28}$ & $\scriptstyle{\pm44}$ & $\scriptstyle{\pm31}$ & $\scriptstyle{\pm12}$ & $\scriptstyle{\pm18}$ & $\scriptstyle{\pm92}$ &$\bm{\scriptstyle{\pm14}}$ &  \vspace{0.75ex} \\
\texttt{walker} & $168$ & $111$ & $116$ & $214$ & $526$ & $\bm{718}$ & $504$  & $0.003$  \vspace{-0.75ex} \\
\texttt{walk} & $\scriptstyle{\pm19}$ & $\scriptstyle{\pm16}$ & $\scriptstyle{\pm18}$ & $\scriptstyle{\pm68}$ & $\scriptstyle{\pm55}$ &$\bm{\scriptstyle{\pm37}}$& $\scriptstyle{\pm33}$  &  \\
\midrule
\texttt{avg} & $175$ & $179$ & $187$ & $213$ & $481$ & $626$ & $664$ &  \\
    \bottomrule
    \end{tabular}
    \end{table}

\subsection{Results on DistractingCS}
\label{app:additional_results:dcs}

We demonstrate the performance of MaDi and our baseline algorithms in the Distracting Control Suite \citep{stone2021distracting} with intensity $0.1$ in Table~\ref{tab:results-distracting_cs}. DistractingCS is an extreme benchmark that features different distractions, such as varying camera angles, changing agent colors, and random videos in the background. 
The results indicate that MaDi performs well on this very challenging benchmark.

\begin{table}[H]
    \captionsetup{width=0.65\textwidth}
    \caption{Generalization performance of MaDi and various baseline algorithms on six different environments after training for $500K$ steps. We evaluate on the Distracting Control Suite with intensity $0.1$, showing mean undiscounted return and standard error over five seeds. MaDi is competitive (no significant difference) with the best baseline in 5/6 environments, denoted by $\star$.}
    \label{tab:results-distracting_cs}
    \centering
    \begin{tabular}{lcccccccl}
\toprule
\texttt{distracting\_cs}  & SAC & DrQ & RAD & SODA & SVEA & SGQN & MaDi & \texttt{p-value} \\
\midrule 
\texttt{ball\_in\_cup} & $112$ & $177$ & $33$ & $107$ & $105$ & $\bm{529}$ & $477$  & $0.738^{\star}$  \vspace{-0.75ex} \\
\texttt{catch} & $\scriptstyle{\pm69}$ & $\scriptstyle{\pm67}$ & $\scriptstyle{\pm30}$ & $\scriptstyle{\pm30}$ & $\scriptstyle{\pm44}$ &$\bm{\scriptstyle{\pm111}}$& $\scriptstyle{\pm99}$  &  \vspace{0.75ex} \\
\texttt{cartpole} & $310$ & $237$ & $233$ & $210$ & $278$ & $358$ & $\bm{377}$  & $0.373^{\star}$  \vspace{-0.75ex} \\
\texttt{balance} & $\scriptstyle{\pm22}$ & $\scriptstyle{\pm19}$ & $\scriptstyle{\pm19}$ & $\scriptstyle{\pm6}$ & $\scriptstyle{\pm22}$ & $\scriptstyle{\pm17}$ &$\bm{\scriptstyle{\pm12}}$ &  \vspace{0.75ex} \\
\texttt{cartpole} & $146$ & $158$ & $162$ & $138$ & $219$ & $\bm{283}$ & $252$  & $0.412^{\star}$  \vspace{-0.75ex} \\
\texttt{swingup} & $\scriptstyle{\pm8}$ & $\scriptstyle{\pm13}$ & $\scriptstyle{\pm6}$ & $\scriptstyle{\pm8}$ & $\scriptstyle{\pm18}$ &$\bm{\scriptstyle{\pm20}}$& $\scriptstyle{\pm30}$  &  \vspace{0.75ex} \\
\texttt{finger} & $28$ & $55$ & $38$ & $52$ & $137$ & $\bm{386}$ & $211$  & $0.002$  \vspace{-0.75ex} \\
\texttt{spin} & $\scriptstyle{\pm5}$ & $\scriptstyle{\pm13}$ & $\scriptstyle{\pm15}$ & $\scriptstyle{\pm18}$ & $\scriptstyle{\pm21}$ &$\bm{\scriptstyle{\pm13}}$& $\scriptstyle{\pm30}$  &  \vspace{0.75ex} \\
\texttt{walker} & $353$ & $727$ & $539$ & $157$ & $\bm{871}$ & $641$ & $866$  & $0.875^{\star}$  \vspace{-0.75ex} \\
\texttt{stand} & $\scriptstyle{\pm66}$ & $\scriptstyle{\pm58}$ & $\scriptstyle{\pm43}$ & $\scriptstyle{\pm20}$ &$\bm{\scriptstyle{\pm22}}$& $\scriptstyle{\pm110}$ & $\scriptstyle{\pm16}$  &  \vspace{0.75ex} \\
\texttt{walker} & $220$ & $429$ & $291$ & $235$ & $568$ & $\bm{642}$ & $570$  & $0.303^{\star}$  \vspace{-0.75ex} \\
\texttt{walk} & $\scriptstyle{\pm12}$ & $\scriptstyle{\pm19}$ & $\scriptstyle{\pm30}$ & $\scriptstyle{\pm91}$ & $\scriptstyle{\pm28}$ &$\bm{\scriptstyle{\pm44}}$& $\scriptstyle{\pm49}$  &  \\
\midrule
\texttt{avg} & $195$ & $297$ & $216$ & $150$ & $363$ & $473$ & $459$ &  \\
    \bottomrule
    \end{tabular}
    \end{table}

\newpage
\section{Mask Examples}
\label{app:mask-examples}

\begin{figure}[H]
    \centering
    \begin{subfigure}[b]{0.3\textwidth}
        \includegraphics[width=\textwidth]{figures/explanation/mask-examples/overlay-aug/mask-examples-walker-clean.pdf}
        \vspace{-1.6em}
        \caption{\texttt{training, walker}}
        \vspace{0.8em}
    \end{subfigure}
     \hspace{0.02\textwidth}
    \begin{subfigure}[b]{0.3\textwidth}
        \includegraphics[width=\textwidth]{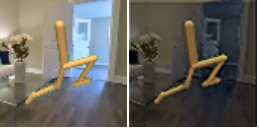}
        \vspace{-1.6em}
        \caption{\texttt{video\_hard, walker}}
        \vspace{0.8em}
    \end{subfigure}
     \hspace{0.02\textwidth}
    \begin{subfigure}[b]{0.3\textwidth}
        \includegraphics[width=\textwidth]{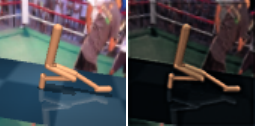}
        \vspace{-1.6em}
        \caption{\texttt{distracting\_cs, walker}}
        \vspace{0.8em}
    \end{subfigure}
    \hfill
        \begin{subfigure}[b]{0.3\textwidth}
        \includegraphics[width=\textwidth]{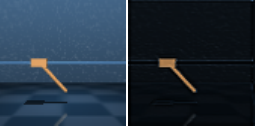}
        \vspace{-1.6em}
        \caption{\texttt{training, cartpole}}
        \vspace{0.8em}
    \end{subfigure}
    \hspace{0.02\textwidth}
    \begin{subfigure}[b]{0.3\textwidth}
        \includegraphics[width=\textwidth]{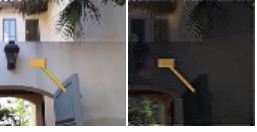}
        \vspace{-1.6em}
        \caption{\texttt{video\_hard, cartpole}}
        \vspace{0.8em}
    \end{subfigure}
    \hspace{0.02\textwidth}
    \begin{subfigure}[b]{0.3\textwidth}
        \includegraphics[width=\textwidth]{figures/explanation/mask-examples/overlay-aug/mask-examples-cartpole-dcs.pdf}
        \vspace{-1.6em}
        \caption{\texttt{distracting\_cs, cartpole}}
        \vspace{0.8em}
    \end{subfigure}
    \hfill
        \begin{subfigure}[b]{0.3\textwidth}
        \includegraphics[width=\textwidth]{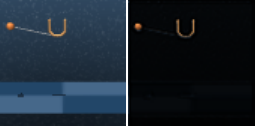}
        \vspace{-1.6em}
        \caption{\texttt{training, ball\_in\_cup}}
        \vspace{0.8em}
    \end{subfigure}
    \hspace{0.02\textwidth}
    \begin{subfigure}[b]{0.3\textwidth}
        \includegraphics[width=\textwidth]{figures/explanation/mask-examples/overlay-aug/mask-examples-ball-hard.pdf}
        \vspace{-1.6em}
        \caption{\texttt{video\_hard, ball\_in\_cup}}
        \vspace{0.8em}
    \end{subfigure}
    \hspace{0.02\textwidth}
    \begin{subfigure}[b]{0.3\textwidth}
        \includegraphics[width=\textwidth]{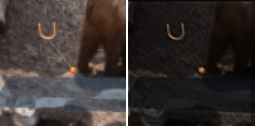}
        \vspace{-1.6em}
        \caption{\texttt{distracting\_cs, ball\_in\_cup}}
        \vspace{0.8em}
    \end{subfigure}
    \hfill
        \begin{subfigure}[b]{0.3\textwidth}
        \includegraphics[width=\textwidth]{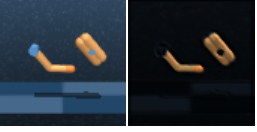}
        \vspace{-1.6em}
        \caption{\texttt{training, finger}}
        \vspace{0.8em}
    \end{subfigure}
    \hspace{0.02\textwidth}
    \begin{subfigure}[b]{0.3\textwidth}
        \includegraphics[width=\textwidth]{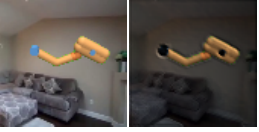}
        \vspace{-1.6em}
        \caption{\texttt{video\_hard, finger}}
        \vspace{0.8em}
    \end{subfigure}
    \hspace{0.02\textwidth}
    \begin{subfigure}[b]{0.3\textwidth}
        \includegraphics[width=\textwidth]{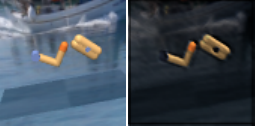}
        \vspace{-1.6em}
        \caption{\texttt{distracting\_cs, finger}}
        \vspace{0.8em}
    \end{subfigure}
    \vspace{-0.5em}
    \captionsetup{width=0.97\textwidth}
    \caption{Examples of masked observations by MaDi under the \texttt{overlay} augmentation in training and testing environments (\texttt{video\_hard} and \texttt{distracting\_cs}) for all domains used. The Masker network produces useful masks, displaying task-relevant information and dimming the distractions.}
    \label{fig:mask-examples-all-domains}
        \Description{The figure displays a grid of 12 sets of two images, grouped into four rows, each with three pairs of side-by-side pictures. Each row corresponds to a different environment (domain), and each column represents a different evaluation mode.
1. Walker Environment:
- Training (a): A walker figure in a simple, blue background on the left, then a sharply masked version of that same image on the right. Only the agent is visible, the rest is black.
- Video Hard (b): The same walker figure, but against a complex background of an indoor setting. The mask is showing the agent very brightly, and the room is very much darkened and dimmed.
- Distraction Control Suite (c): The walker figure on an angle, with a picture of a boxing ring in the background. The mask dims the agent's surroundings well.
2. CartPole Environment:
- Training (d): A vertical pole pointing downwards and to the right attached to a rectangular cart. Clear mask.
- Video Hard (e): The cartpole, with a background of an outdoor scene with walls, plants, and a door. Good mask again.
- Distracting Control Suite (f): Cartpole at an angle with an outdoor background. The agent is visible in the mask, while surroundings are dimmed a lot.
3. Ball in Cup Environment:
- Training (g): A ball attached to a string inside a cup, displayed against a blue background. The mask is clear.
- Video Hard (h): The cup and ball again, against a living room background. Masks now mostly shows the ball and the cup, the rest is darkened but slightly visible, as in the other video_hard masks.
- Distracting Control Suite (i): An image of sand and a bear in the back, cup and ball in front. The masks shows the cup and the ball nicely. Surroundings are darkened mostly.
4. Finger Environment:
- Training (j): A mechanical finger against a blue background. Mask is clear, but it does also mask some small blue parts of the agent.
- Video Hard (k): The same finger set against a background of a living room. Mask is a bit worse, but still quite clear.
- Distracting Control Suite (l): The finger now at an angle, an image of a boat on water in the background. The mask is quite good.
The images effectively demonstrate the different environments and evaluation modes, illustrating the Masker network's ability to produce useful masks that dim the background.}
\end{figure}

In Figure~\ref{fig:mask-examples-all-domains}, we show examples of masked observations by the Masker network of MaDi. These masked observations are seen by the actor and critic networks, enabling them to perform well even in the presence of visual distractions.

\newpage
\subsection{Masks over time}

\begin{figure}[H]
  \centering
  \resizebox{0.9\linewidth}{!}{
  \Large
  \begin{tabular}{m{0.1\textwidth} m{0.3\textwidth} m{0.3\textwidth} m{0.3\textwidth}}
    & \textbf{Mask} & \textbf{Masked Observation} & \textbf{Observation} \\
    \textbf{Step 0} & 
      \includegraphics[width=\linewidth]{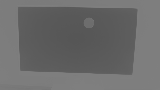} &
      \includegraphics[width=\linewidth]{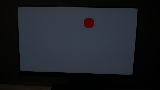} &
      \includegraphics[width=\linewidth]{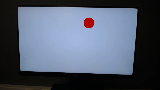} \\
    \textbf{Step 20K} &
      \includegraphics[width=\linewidth]{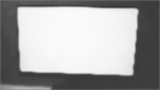} &
      \includegraphics[width=\linewidth]{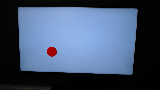} &
      \includegraphics[width=\linewidth]{figures/robot/masks/overlay/seed1400/train/step21000_frame0_maskedobs.png} \\
    \textbf{Step 100K} &
      \includegraphics[width=\linewidth]{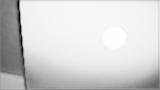} &
      \includegraphics[width=\linewidth]{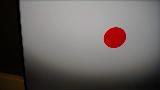} &
      \includegraphics[width=\linewidth]{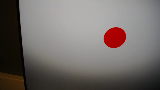} \\
  \end{tabular}
  }
  \captionsetup{width=0.9\textwidth}
  \vspace{-0.2cm}
  \caption{Masks produced by MaDi (with \texttt{overlay} augmentation) on the UR-VisualReacher training environment. After just $20,000$ timesteps MaDi has learned to output a subtle, but useful mask. The initial Masker outputs values close to $0.5$, shown by the gray image in the top-left corner (vaguely revealing the shape of the target circle).}
  \label{fig:masks-ur5-overlay-train}
  \Description{The figure displays a series of images highlighting the development and effect of masks produced by MaDi in the UR-VisualReacher training environment over time. The image set is structured in three rows and three columns, rows representing three different time steps (Step 0, Step 20K, and Step 100K) and columns representing three types of visuals (Mask, Masked Observation, and Observation).
Step 0:
Mask: A uniformly dark gray image, but with a light gray dot showing.
Masked Observation: An entirely grayed-out image with a bit of red, representing the application of the mask to the observation.
Observation: A simple white background with a clearly visible red dot.
Step 20K:
Mask: Predominantly gray with a white rectangle located centrally, where the monitor is.
Masked Observation: The original observation visible through the white rectangle of the mask, showing the white background and the red dot.
Observation: Similar to the observation at Step 0.
Step 100K:
Mask: Mostly gray but has a white region vaguely outlining the red dot.
Masked Observation: The red dot in the observation is more visible because the camera is much closer now.
Observation: Much like the previous observations, it has a white background and a red dot.
The sequence showcases the algorithm's progression in producing increasingly refined masks over time to emphasize the red dot in the observation. By Step 100K, the mask somewhat captures the shape and location of the target red dot, indicating its learning and adaptability.}
\end{figure}

\begin{figure}[H]
  \centering
  \resizebox{0.9\linewidth}{!}{
  \Large
  \begin{tabular}{m{0.1\textwidth} m{0.3\textwidth} m{0.3\textwidth} m{0.3\textwidth}}
    & \textbf{Mask} & \textbf{Masked Observation} & \textbf{Observation} \\
    \textbf{Step 0} & 
      \includegraphics[width=\linewidth]{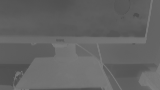} &
      \includegraphics[width=\linewidth]{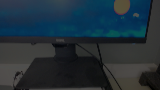} &
      \includegraphics[width=\linewidth]{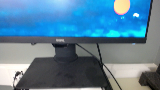} \\
    \textbf{Step 20K} &
      \includegraphics[width=\linewidth]{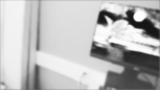} &
      \includegraphics[width=\linewidth]{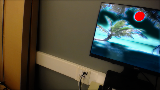} &
      \includegraphics[width=\linewidth]{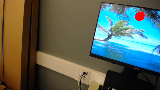} \\
    \textbf{Step 100K} &
      \includegraphics[width=\linewidth]{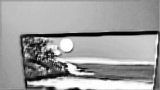} &
      \includegraphics[width=\linewidth]{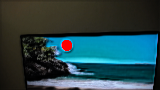} &
      \includegraphics[width=\linewidth]{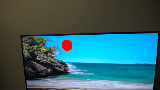} \\
  \end{tabular}
  }
  \captionsetup{width=0.9\textwidth}
  \vspace{-0.2cm}
  \caption{Masks produced by MaDi (with \texttt{overlay} augmentation) on the UR-VisualReacher-VideoBackgrounds testing environment. After $20,000$ timesteps MaDi has learned to output a subtle, but helpful mask.}
  \label{fig:masks-ur5-overlay-eval}
  \Description{The figure presents a series of images that display the evolution of masks produced by MaDi on the UR-VisualReacher-VideoBackgrounds testing environment over three distinct time intervals: Step 0, Step 20K, and Step 100K. Each row represents one of these time intervals and is divided into three columns, corresponding to the Mask, Masked Observation, and the raw Observation.
Step 0:
Mask: An almost uniformly dark gray image with faint outlines that slightly hint at the shape of the monitor stand and bottom of the screen.
Masked Observation: A largely dimmed image showing a faint silhouette of the monitor with a video playing in the background, barely discernible.
Observation: The computer monitor displaying an ocean scene from above with azure waters. The red dot appears a bit yellowish here, the camera was maybe not calibrated well automatically in this timestep.
Step 20K:
Mask: A more light gray mask, but this visibly shows that the screen is on the right side, and there's white dot clearly visible.
Masked Observation: The red dot comes through well, while other parts are dimmed slightly. The beach video is still visible. 
Observation: The beach video showing on the screen on the right, with the red dot.
Step 100K:
Mask: The mask now sharply delineates the red dot, and slightly dims the rest. Shapes of the video are still coming through.
Masked Observation: The red dot is pronounced, video somewhat dimmed.
Observation: The video frame of the beach showing, with the red dot pretty big, as the camera is closer.
The progression signifies MaDi's capability to produce increasingly refined masks over time, allowing for more focused observation of the robot against varying video backgrounds.}
\end{figure}

In this section we show how the masks evolve over time. The Masker network is initialized in such a way that the first masks consist of values very close to $0.5$, resulting in a fully gray image. Then, after just a small number of updates, the Masker can already generate useful masks that focus on the relevant features. See Figures~\ref{fig:masks-ur5-overlay-train} and \ref{fig:masks-ur5-overlay-eval}.

\section{Other Augmentations}
\label{app:augmentation-details}

\begin{figure}[H]
  \centering
  \resizebox{0.62\linewidth}{!}{
  \LARGE
  \begin{tabular}{m{0.12\textwidth} m{0.2\textwidth} m{0.2\textwidth} m{0.2\textwidth} m{0.2\textwidth}}
    & {\centering\texttt{original}\par} & {\centering\texttt{conv}\par} & {\centering\texttt{overlay}\par} & {\centering\texttt{splice}\par} \\
    \texttt{ball\_in\_cup} & 
      \includegraphics[width=\linewidth]{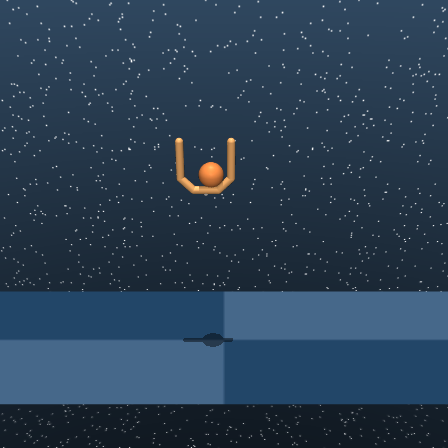} &
      \includegraphics[width=\linewidth]{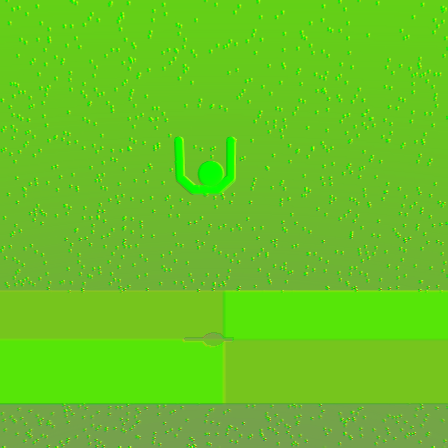} &
      \includegraphics[width=\linewidth]{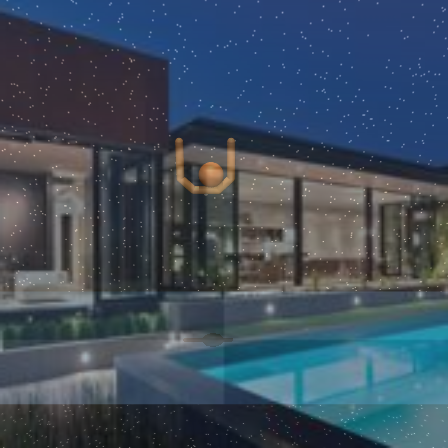} &
      \includegraphics[width=\linewidth]{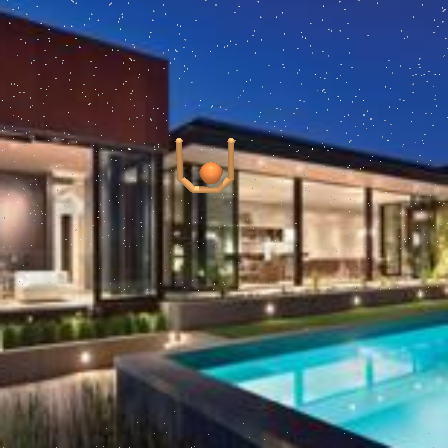} \\
    \texttt{cartpole} &
      \includegraphics[width=\linewidth]{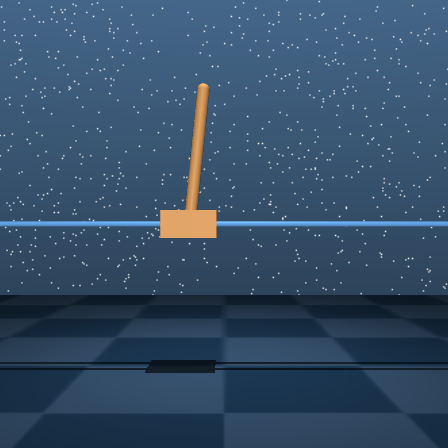} &
      \includegraphics[width=\linewidth]{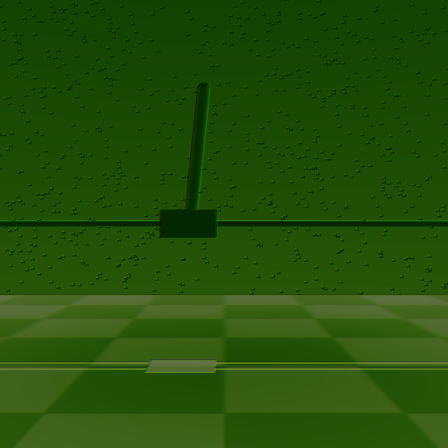} &
      \includegraphics[width=\linewidth]{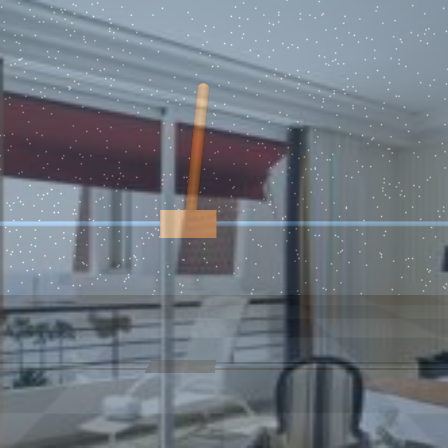} &
      \includegraphics[width=\linewidth]{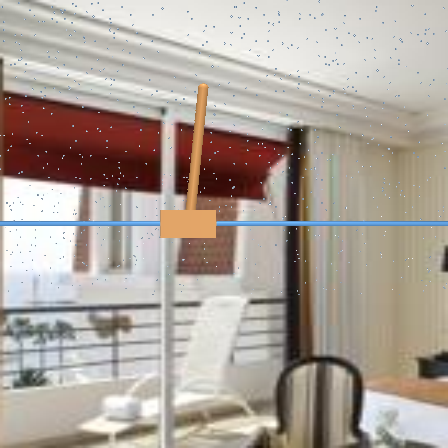} \\
    \texttt{finger} &
      \includegraphics[width=\linewidth]{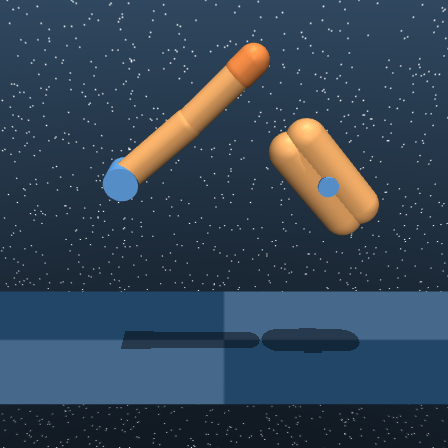} &
      \includegraphics[width=\linewidth]{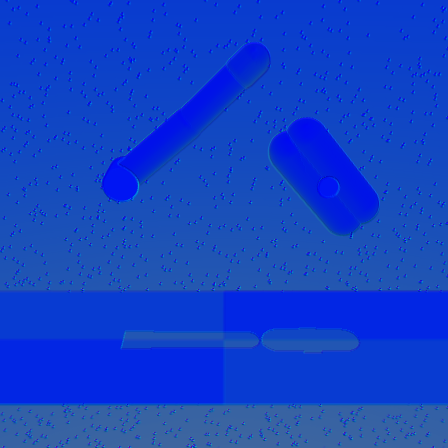} &
      \includegraphics[width=\linewidth]{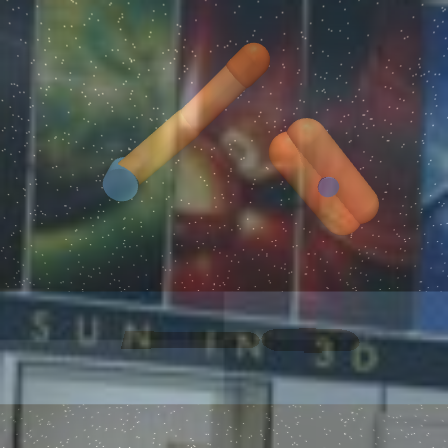} &
      \includegraphics[width=\linewidth]{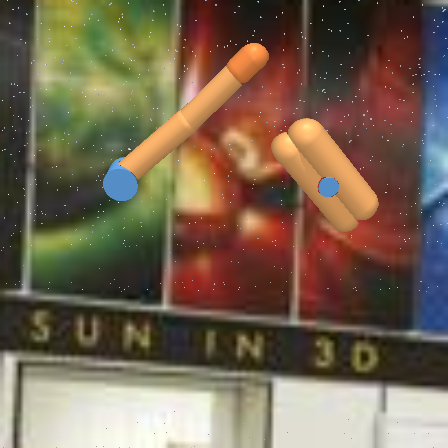} \\
    \texttt{walker} &
      \includegraphics[width=\linewidth]{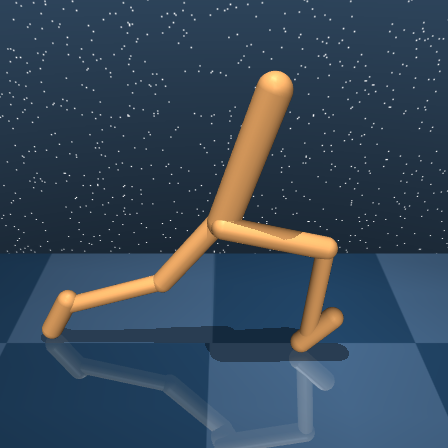} &
      \includegraphics[width=\linewidth]{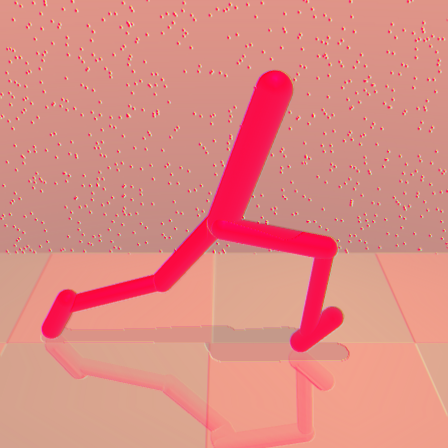} &
      \includegraphics[width=\linewidth]{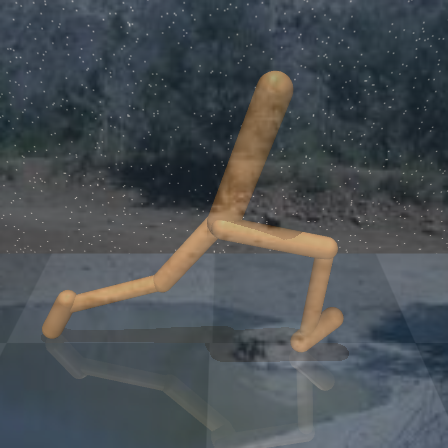} &
      \includegraphics[width=\linewidth]{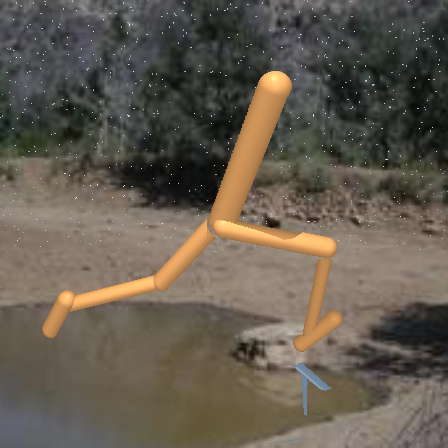} \\
  \end{tabular}
  }
  \captionsetup{width=0.9\textwidth}
  \vspace{-0.2cm}
  \caption{Various augmentations on the four domains of the DeepMind Control Suite used in our study. Applying \texttt{overlay} results in dark-blue and blurry visuals, whereas \texttt{splice} produces crisper images. The \texttt{splice} augmentation requires some prior knowledge about the environment to set the HSV thresholds.}
  \label{fig:augmentation-overview}
  \Description{
The figure displays a grid of images, arranged in 4 rows and 4 columns, each column representing an augmentation applied to the four domains of the DeepMind Control Suite used in our study.
1. Ball in Cup:
- Original: (so actually no augmentation in this column) A ball laying in a cup against a dark blue background.
- Conv: A bright green gradient filter over the same original image.
- Overlay: The same ball in cup scene with an overlaid image, making the background a blurred version of a poolside setting.
- Splice: The ball in cup merged into an image of a swimming pool setting, seamlessly integrating both scenes, without the blurriness. The ball and cup are "cut out" from the original image and "pasted" on top of the random augmentation image, which is the poolside in this case.
2. CartPole:
- Original: A vertical pole affixed to a cart on a blue platform.
- Conv: A dark green gradient filter over the original image.
- Overlay: The cartpole with a blurry, overlaid image of an indoor room setting.
- Splice: The same setting but again less blurry.
3. Finger:
- Original: the finger and spin object against a blue background.
- Conv: same image but now with a different lighter blue gradient filter.
- Overlay: The finger against a merged blurry backdrop of an movie poster setting.
- Splice: same, less blurry.
4. Walker:
- Original: A walker figure posed in a walking stance on a blue background.
- Conv: The walker now with a red gradient filter.
- Overlay: The walker against a merged blurry backdrop of a outdoor scene. Some plants and a dirt path.
- Splice: same scene, less blurry.
These augmentations demonstrate the effects of applying a random convolution, overlay, and splice on original images from the DeepMind Control Suite, highlighting the changes in clarity, coloration, and context.}
\vspace{-1em}
\end{figure}

In this section, we present the results of MaDi under different augmentation techniques instead of the default \texttt{overlay}. We experimented with a random convolutional layer as augmentation, called \texttt{conv} in \citep{hansen2021soda}, and an augmentation based on hue-saturation-value (HSV) thresholds named \texttt{splice} \citep{yung2022splice_augmentation}. In Figure~\ref{fig:augmentation-overview}, we present an overview of the augmentations used.

\begin{figure}[H]
    \centering
    \begin{subfigure}[b]{0.3\textwidth}
        \includegraphics[width=\textwidth]{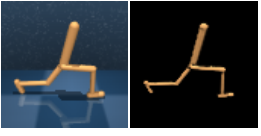}
        \vspace{-1.6em}
        \caption{\texttt{training, walker}}
        \vspace{0.8em}
    \end{subfigure}
    \hspace{0.02\textwidth}
    \begin{subfigure}[b]{0.3\textwidth}
        \includegraphics[width=\textwidth]{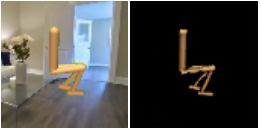}
        \vspace{-1.6em}
        \caption{\texttt{video\_hard, walker}}
        \vspace{0.8em}
    \end{subfigure}
    \hspace{0.02\textwidth}
    \begin{subfigure}[b]{0.3\textwidth}
        \includegraphics[width=\textwidth]{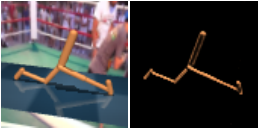}
        \vspace{-1.6em}
        \caption{\texttt{distracting\_cs, walker}}
        \vspace{0.8em}
    \end{subfigure}
    \hfill
        \begin{subfigure}[b]{0.3\textwidth}
        \includegraphics[width=\textwidth]{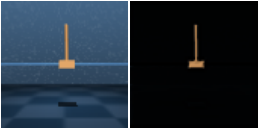}
        \vspace{-1.6em}
        \caption{\texttt{training, cartpole}}
        \vspace{0.8em}
    \end{subfigure}
    \hspace{0.02\textwidth}
    \begin{subfigure}[b]{0.3\textwidth}
        \includegraphics[width=\textwidth]{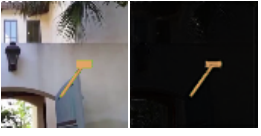}
        \vspace{-1.6em}
        \caption{\texttt{video\_hard, cartpole}}
        \vspace{0.8em}
    \end{subfigure}
    \hspace{0.02\textwidth}
    \begin{subfigure}[b]{0.3\textwidth}
        \includegraphics[width=\textwidth]{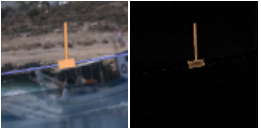}
        \vspace{-1.6em}
        \caption{\texttt{distracting\_cs, cartpole}}
        \vspace{0.8em}
    \end{subfigure}
    \hfill
        \begin{subfigure}[b]{0.3\textwidth}
        \includegraphics[width=\textwidth]{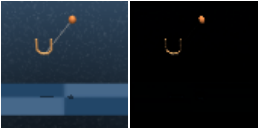}
        \vspace{-1.6em}
        \caption{\texttt{training, ball\_in\_cup}}
        \vspace{0.8em}
    \end{subfigure}
    \hspace{0.02\textwidth}
    \begin{subfigure}[b]{0.3\textwidth}
        \includegraphics[width=\textwidth]{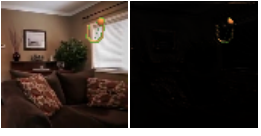}
        \vspace{-1.6em}
        \caption{\texttt{video\_hard, ball\_in\_cup}}
        \vspace{0.8em}
    \end{subfigure}
    \hspace{0.02\textwidth}
    \begin{subfigure}[b]{0.3\textwidth}
        \includegraphics[width=\textwidth]{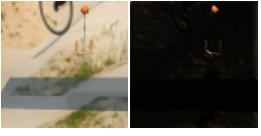}
        \vspace{-1.6em}
        \caption{\texttt{distracting\_cs, ball\_in\_cup}}
        \vspace{0.8em}
    \end{subfigure}
    \hfill
        \begin{subfigure}[b]{0.3\textwidth}
        \includegraphics[width=\textwidth]{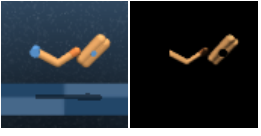}
        \vspace{-1.6em}
        \caption{\texttt{training, finger}}
        \vspace{0.8em}
    \end{subfigure}
    \hspace{0.02\textwidth}
    \begin{subfigure}[b]{0.3\textwidth}
        \includegraphics[width=\textwidth]{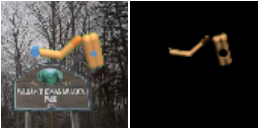}
        \vspace{-1.6em}
        \caption{\texttt{video\_hard, finger}}
        \vspace{0.8em}
    \end{subfigure}
    \hspace{0.02\textwidth}
    \begin{subfigure}[b]{0.3\textwidth}
        \includegraphics[width=\textwidth]{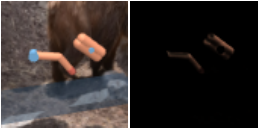}
        \vspace{-1.6em}
        \caption{\texttt{distracting\_cs, finger}}
        \vspace{0.8em}
    \end{subfigure}
    \vspace{-0.7em}
    \captionsetup{width=0.97\textwidth}
    \caption{Examples of masked observations by MaDi under the \texttt{splice} augmentation in training and testing environments (\texttt{video\_hard} and \texttt{distracting\_cs}) for all domains used. The Masker network generally produces clear and precise masks, but it can sometimes mask too much (i.e., parts of the agent) when using \texttt{splice}, as seen in the bottom right.}
    \label{fig:mask-examples-all-domains-splice}
    \Description{The figure displays a grid of 12 sets of two images, grouped into four rows, each with three pairs of side-by-side pictures. Each row corresponds to a different environment (domain), and each column represents a different evaluation mode.
1. Walker Environment:
- Training (a): A walker figure in a simple, blue background on the left, then a sharply masked version of that same image on the right. Only the agent is visible, the rest is black.
- Video Hard (b): The same walker figure, but against a complex background of an indoor setting. The mask is very clear again.
- Distraction Control Suite (c): The walker figure on an angle, with a picture of a boxing ring in the background. The mask is dimly visible, with its surroundings darkened well, but the agent is also darkened a bit here.
2. CartPole Environment:
- Training (d): A vertical pole pointing upwards on a cart. Clear mask.
- Video Hard (e): The cartpole, with a background of an outdoor scene with walls, plants, and a door. Clear mask again.
- Distracting Control Suite (f): Cartpole at an angle with an outdoor background. The agent is visible in the mask, while surroundings are dimmed a lot.
3. Ball in Cup Environment:
- Training (g): A ball attached to a string inside a cup, displayed against a blue background. The mask is clear, but the cup is masked a bit too much.
- Video Hard (h): The cup and ball again, against a living room background. Mask now mostly shows the ball, and a bit of the cup, all else black.
- Distracting Control Suite (i): An image of sand and a bicycle wheel in the back, cup and ball in front. The masks barely shows the cup now. Ball is still visible. Surroundings are darkened fully.
4. Finger Environment:
- Training (j): A mechanical finger against a blue background. Mask is clear, but it does mask also some small blue parts of the agent.
- Video Hard (k): The same finger set against a background of a forest. Mask is a bit worse, but still very clear.
- Distracting Control Suite (l): The finger now at an angle, an image of a real animal (bear) in the background. The mask is not good, large parts of the agent are also masked here.
The images effectively demonstrate the different environments and evaluation modes, illustrating the Masker network's ability to produce clear and precise masks in most cases, by highlighting only task-relevant information and dimming distractions.}
\end{figure}

MaDi can perform quite well with the \texttt{splice} augmentation. The masks often look even clearer than with \texttt{overlay}, as shown in Figure~\ref{fig:mask-examples-all-domains-splice}. However, we have also occasionally seen masks that turn fully-black with the \texttt{splice} augmentation, which hides all relevant information needed by the actor to take an optimal action. This is likely why the performance of MaDi with \texttt{splice} is not the best in all environments. Further research is necessary to determine why MaDi sometimes produces black masks under this augmentation technique.

\vspace{-0.8em}

\begin{figure}[H]
\centering
\captionsetup{width=0.95\textwidth}
\captionof{table}{Comparing the effect of different data augmentations. We show the generalization performance of MaDi on multiple benchmarks, using three different augmentations. The agents are all trained for $500K$ steps. We present the mean undiscounted return and standard error over five seeds.}
\vspace{-1em}
\subfloat[video\_easy]{
    \resizebox{0.3\textwidth}{!}{%
    \begin{tabular}{lcccccc}
    \toprule
 augmentation & overlay & conv & splice \\
\midrule
\texttt{ball\_in\_cup} & $\bm{807}$ & $603$ & $617$  \vspace{-0.75ex} \\
\texttt{catch} &$\bm{\scriptstyle{\pm144}}$& $\scriptstyle{\pm115}$ & $\scriptstyle{\pm196}$  \vspace{0.75ex} \\
\texttt{cartpole} & $\bm{982}$ & $903$ & $881$  \vspace{-0.75ex} \\
\texttt{balance} &$\bm{\scriptstyle{\pm4}}$& $\scriptstyle{\pm37}$ & $\scriptstyle{\pm91}$  \vspace{0.75ex} \\
\texttt{cartpole} & $\bm{848}$ & $693$ & $785$  \vspace{-0.75ex} \\
\texttt{swingup} &$\bm{\scriptstyle{\pm6}}$& $\scriptstyle{\pm34}$ & $\scriptstyle{\pm22}$  \vspace{0.75ex} \\
\texttt{finger} & $\bm{679}$ & $304$ & $568$  \vspace{-0.75ex} \\
\texttt{spin} &$\bm{\scriptstyle{\pm17}}$& $\scriptstyle{\pm85}$ & $\scriptstyle{\pm131}$  \vspace{0.75ex} \\
\texttt{walker} & $\bm{967}$ & $408$ & $828$  \vspace{-0.75ex} \\
\texttt{stand} &$\bm{\scriptstyle{\pm3}}$& $\scriptstyle{\pm149}$ & $\scriptstyle{\pm129}$  \vspace{0.75ex} \\
\texttt{walker} & $\bm{895}$ & $476$ & $861$  \vspace{-0.75ex} \\
\texttt{walk} &$\bm{\scriptstyle{\pm24}}$& $\scriptstyle{\pm60}$ & $\scriptstyle{\pm10}$  \\
\midrule
\texttt{avg} & $\bm{863}$ & $565$ & $757$  \\
    \bottomrule
    \end{tabular}
    }
}
\quad
\subfloat[video\_hard]{
    \resizebox{0.3\textwidth}{!}{%
    \begin{tabular}{lcccccc}
    \toprule
 augmentation & overlay & conv & splice \\
\midrule
\texttt{ball\_in\_cup} & $\bm{758}$ & $84$ & $610$  \vspace{-0.75ex} \\
\texttt{catch} &$\bm{\scriptstyle{\pm135}}$& $\scriptstyle{\pm30}$ & $\scriptstyle{\pm194}$  \vspace{0.75ex} \\
\texttt{cartpole} & $827$ & $320$ & $\bm{879}$  \vspace{-0.75ex} \\
\texttt{balance} & $\scriptstyle{\pm25}$ & $\scriptstyle{\pm34}$ &$\bm{\scriptstyle{\pm90}}$ \vspace{0.75ex} \\
\texttt{cartpole} & $619$ & $161$ & $\bm{769}$  \vspace{-0.75ex} \\
\texttt{swingup} & $\scriptstyle{\pm24}$ & $\scriptstyle{\pm14}$ &$\bm{\scriptstyle{\pm27}}$ \vspace{0.75ex} \\
\texttt{finger} & $358$ & $61$ & $\bm{566}$  \vspace{-0.75ex} \\
\texttt{spin} & $\scriptstyle{\pm25}$ & $\scriptstyle{\pm18}$ &$\bm{\scriptstyle{\pm130}}$ \vspace{0.75ex} \\
\texttt{walker} & $\bm{920}$ & $210$ & $824$  \vspace{-0.75ex} \\
\texttt{stand} &$\bm{\scriptstyle{\pm14}}$& $\scriptstyle{\pm29}$ & $\scriptstyle{\pm128}$  \vspace{0.75ex} \\
\texttt{walker} & $504$ & $99$ & $\bm{858}$  \vspace{-0.75ex} \\
\texttt{walk} & $\scriptstyle{\pm33}$ & $\scriptstyle{\pm12}$ &$\bm{\scriptstyle{\pm17}}$ \\
\midrule
\texttt{avg} & $664$ & $156$ & $\bm{751}$  \\
    \bottomrule
    \end{tabular}
    }
}
\quad
\subfloat[distracting\_cs]{
    \resizebox{0.3\textwidth}{!}{%
    \begin{tabular}{lcccccc}
    \toprule
 augmentation & overlay & conv & splice \\
\midrule
\texttt{ball\_in\_cup} & $477$ & $165$ & $\bm{518}$  \vspace{-0.75ex} \\
\texttt{catch} & $\scriptstyle{\pm99}$ & $\scriptstyle{\pm61}$ &$\bm{\scriptstyle{\pm190}}$ \vspace{0.75ex} \\
\texttt{cartpole} & $377$ & $241$ & $\bm{409}$  \vspace{-0.75ex} \\
\texttt{balance} & $\scriptstyle{\pm12}$ & $\scriptstyle{\pm8}$ &$\bm{\scriptstyle{\pm24}}$ \vspace{0.75ex} \\
\texttt{cartpole} & $252$ & $176$ & $\bm{300}$  \vspace{-0.75ex} \\
\texttt{swingup} & $\scriptstyle{\pm30}$ & $\scriptstyle{\pm8}$ &$\bm{\scriptstyle{\pm23}}$ \vspace{0.75ex} \\
\texttt{finger} & $\bm{211}$ & $111$ & $72$  \vspace{-0.75ex} \\
\texttt{spin} &$\bm{\scriptstyle{\pm30}}$& $\scriptstyle{\pm42}$ & $\scriptstyle{\pm20}$  \vspace{0.75ex} \\
\texttt{walker} & $\bm{866}$ & $271$ & $346$  \vspace{-0.75ex} \\
\texttt{stand} &$\bm{\scriptstyle{\pm16}}$& $\scriptstyle{\pm56}$ & $\scriptstyle{\pm59}$  \vspace{0.75ex} \\
\texttt{walker} & $\bm{570}$ & $378$ & $230$  \vspace{-0.75ex} \\
\texttt{walk} &$\bm{\scriptstyle{\pm49}}$& $\scriptstyle{\pm31}$ & $\scriptstyle{\pm28}$  \\
\midrule
\texttt{avg} & $\bm{459}$ & $224$ & $313$  \\
    \bottomrule
    \end{tabular}
    }
}
\end{figure}

\subsection{Robotic experiments with \texttt{conv} augmentation}
\label{app:ur5_conv}

\begin{figure}[H]
    \centering
    \begin{subfigure}[b]{0.48\textwidth}
        \includegraphics[width=\textwidth]{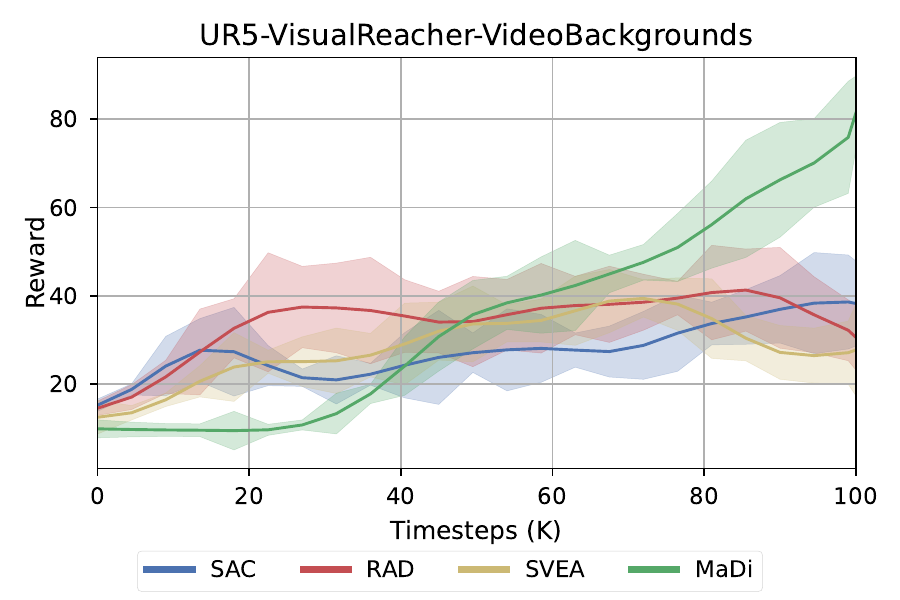}
        \caption{testing environment}
        \vspace{-0.5em}
        \label{fig:subfig1-ur5conv}
    \end{subfigure}
    \hfill
    \begin{subfigure}[b]{0.48\textwidth}
        \includegraphics[width=\textwidth]{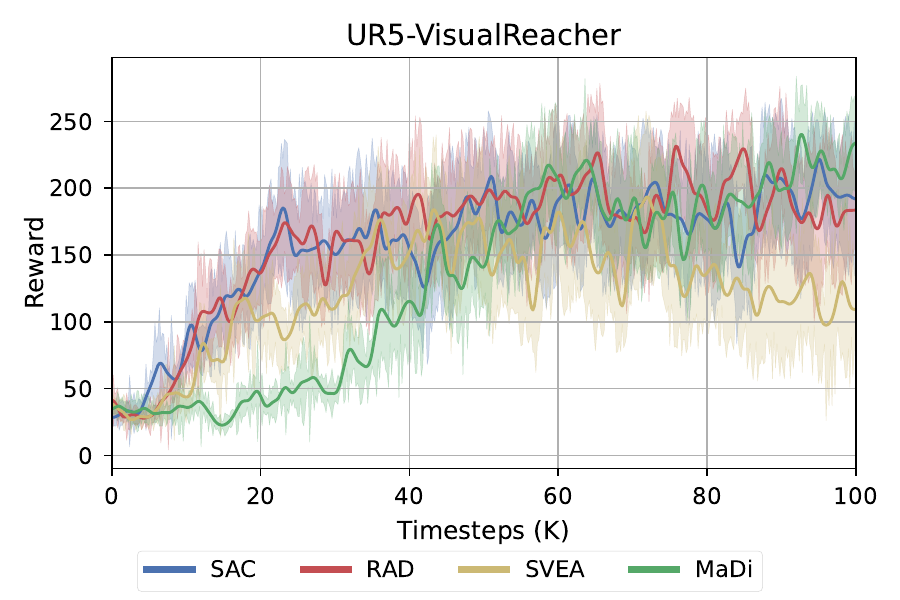}
        \caption{training environment}
        \vspace{-0.5em}
        \label{fig:subfig2-ur5conv}
    \end{subfigure}
    \captionsetup{width=0.97\textwidth}
    \caption{Performance of MaDi and three baselines on the UR5-VisualReacher environment, where both MaDi and SVEA use the \texttt{conv} augmentation now. Agents are trained on the clean setup (white background with the red dot) for $100K$ steps. Curves show the mean over five seeds with standard error shaded alongside.}
    \label{fig:ur5_conv_graphs}
    \Description{This figure showcases two graphs detailing the performance of MaDi and three other baselines on the UR5-VisualReacher-VideoBackgrounds (on the left) and UR5-VisualReacher (on the right) tasks. 
    1. For the testing environment on the left.
The vertical y-axis, marked 'Reward', ranges from 0 to a bit above 80, while the horizontal x-axis, labeled 'Timesteps (K)', spans from 0 to 100. There are four distinct learning curves, each represented by a unique colored line: SAC (blue), RAD (red), SVEA (yellow), and MaDi (green). The performance of each model is tracked over time, with areas of uncertainty surrounding each curve highlighted with a lighter shade of the respective color. Over the span of 100K timesteps, MaDi's performance generally trends upward, achieving a reward of around 80. Meanwhile, the other models a learn a bit quicker in the beginning, but plateau around a value of 40. MaDi surpasses the other three around halfway, at 50K steps.
    2. For the training environment on the right.
The vertical y-axis, marked 'Reward', ranges from 0 to a bit above 250, while the horizontal x-axis, labeled 'Timesteps (K)', spans from 0 to 100. The same four learning algorithms are shown. Over the span of 100K timesteps, MaDi's performance generally trends upward, achieving a reward of around 230. Meanwhile, the other models a learn a bit quicker in the beginning, but plateau around a value of 200. MaDi reaches the other three around halfway, at 40K steps. SVEA seems to underperform, as its curves maxes out around 180 at 60K steps, but ends around 120 after the full run.}
\end{figure}

We have also run experiments with the \texttt{conv} augmentation on our UR-VisualReacher task, to see whether it can help to improve generalization on the video backgrounds, see Figure~\ref{fig:ur5_conv_graphs}. The experiments presented in the main body use the \texttt{overlay} augmentation. (These results are shown in Figure~\ref{fig:ur5-eval} and the corresponding training environment curves in Figure~\ref{fig:ur5-train}.) 

\begin{figure}[H]
  \centering
  \resizebox{0.9\linewidth}{!}{
  \Large
  \begin{tabular}{m{0.1\textwidth} m{0.3\textwidth} m{0.3\textwidth} m{0.3\textwidth}}
    & \textbf{Mask} & \textbf{Masked Observation} & \textbf{Observation} \\
    \textbf{Step 0} & 
      \includegraphics[width=\linewidth]{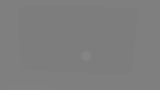} &
      \includegraphics[width=\linewidth]{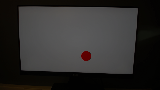} &
      \includegraphics[width=\linewidth]{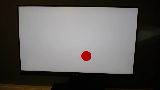} \\
    \textbf{Step 20K} &
      \includegraphics[width=\linewidth]{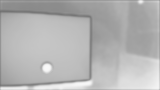} &
      \includegraphics[width=\linewidth]{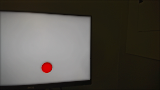} &
      \includegraphics[width=\linewidth]{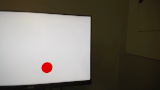} \\
    \textbf{Step 100K} &
      \includegraphics[width=\linewidth]{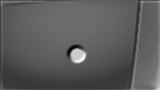} &
      \includegraphics[width=\linewidth]{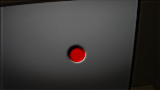} &
      \includegraphics[width=\linewidth]{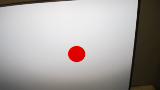} \\
  \end{tabular}
  }
  \captionsetup{width=0.9\textwidth}
  \vspace{-0.2cm}
  \caption{Masks produced by MaDi (with \texttt{conv} augmentation) on the UR-VisualReacher training environment. After just $20,000$ timesteps MaDi has already learned to output useful masks. The initial Masker outputs values close to $0.5$, shown by the gray image in the top-left corner (which vaguely reveals the shape of the target circle).}
  \label{fig:masks-ur5-conv-train}
  \vspace{8em}
  \Description{The figure showcases a series of images that detail the progression of masks generated by MaDi in the UR-VisualReacher training environment across three notable steps: Step 0, Step 20K, and Step 100K. The grid layout consists of three rows, each representing a time step, and three columns, displaying the Mask, Masked Observation, and raw Observation.
Step 0:
Mask: An entirely dark gray rectangle, with no distinct features visible except for a vague small lightgrey circle.
Masked Observation: Primarily dark with a faint suggestion of a monitor, where the red dot is somewhat perceptible.
Observation: A straightforward image of a white monitor screen with a clear red dot in the middle.
Step 20K:
Mask: The mask is already much better, showing a whitish dot in the lower center.
Masked Observation: The monitor screen and red dot are more discernible against a subdued background, revealing the red dot's position.
Observation: Similar to the prior observation, a white monitor screen is evident with a central red dot.
Step 100K:
Mask: Predominantly dark gray, white circle in the center, indicating where the red dot is situated.
Masked Observation: The red dot with dark grey around it. Some edges of the screen are also still visible.
Observation: A consistent white monitor screen displays the red dot, similar to previous steps.
The sequence effectively demonstrates MaDi's capability to produce increasingly precise masks over time, emphasizing the red dot's location amidst the consistent background of the monitor.}
\end{figure}

\begin{figure}[H]
  \centering
  \resizebox{0.9\linewidth}{!}{
  \Large
  \begin{tabular}{m{0.1\textwidth} m{0.3\textwidth} m{0.3\textwidth} m{0.3\textwidth}}
    & \textbf{Mask} & \textbf{Masked Observation} & \textbf{Observation} \\
    \textbf{Step 0} & 
      \includegraphics[width=\linewidth]{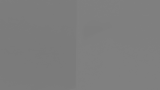} &
      \includegraphics[width=\linewidth]{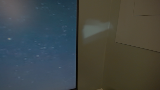} &
      \includegraphics[width=\linewidth]{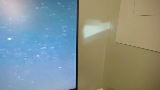} \\
    \textbf{Step 20K} &
      \includegraphics[width=\linewidth]{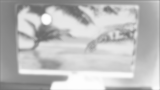} &
      \includegraphics[width=\linewidth]{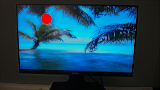} &
      \includegraphics[width=\linewidth]{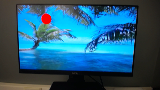} \\
    \textbf{Step 100K} &
      \includegraphics[width=\linewidth]{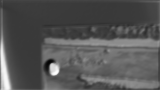} &
      \includegraphics[width=\linewidth]{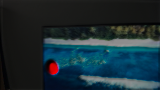} &
      \includegraphics[width=\linewidth]{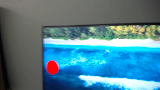} \\
  \end{tabular}
  }
  \captionsetup{width=0.9\textwidth}
  \vspace{-0.2cm}
  \caption{Masks produced by MaDi (with \texttt{conv} augmentation) on the UR-VisualReacher-VideoBackgrounds testing environment. After just $20,000$ timesteps MaDi can generate useful masks on the unseen testing environment. The initial Masker network outputs values close to $0.5$, shown by the gray image in the top-left corner.}
  \label{fig:masks-ur5-conv-eval}
  \vspace{2em}
  \Description{The figure presents a series of images that demonstrate the evolution of masks developed by MaDi in the UR-VisualReacher-VideoBackgrounds testing environment at three distinct steps: Step 0, Step 20K, and Step 100K. The arrangement consists of three rows, with each row corresponding to a specific time step, and three columns representing the Mask, Masked Observation, and the Observation.
Step 0:
Mask: A uniform dark gray rectangle.
Masked Observation: The image is mainly dimmed out, with only a faint hint of the blue starry background from the video and no red dot here.
Observation: The monitor displays a starry night scene on the screen, but the camera is pointing to a part of the screen that apparently does not have the red dot on it.
Step 20K:
Mask: Displays blurred shapes, seemingly capturing the silhouette of the beach scene including the water, sky, and the palm tree. A clear white dot is already showing.
Masked Observation: Here, the beach video is more visible, with the mask allowing the blue water and palm tree to show through more clearly, as well as the red dot.
Observation: Identical beach scene as described before, with the waters, palm tree, and the red dot.
Step 100K:
Mask: Dark gray, with a clear white circle around the lower center, highlighting the area around the red dot.
Masked Observation: The red dot is accentuated, while the surrounding beach scene is subdued.
Observation: A view of the same beach scene with the red dot.
The sequence of images emphasizes MaDi's ability to refine its mask generation over time, progressively emphasizing the red dot against the video backgrounds.}
\end{figure}

The masks generated by MaDi under the \texttt{conv} augmentation seem to be a little clearer than the masks generated under the \texttt{overlay} augmentation. We think that MaDi with \texttt{overlay} was still able to perform better as the encoder network (see Figure~\ref{fig:overview}) also benefits from seeing some strong \texttt{overlay} augmentation during training. Especially early on, when the masks are not fully formed yet, the encoder will receive quite a lot of augmented pixels, which can help its robustness.
In Figures~\ref{fig:masks-ur5-conv-train} and \ref{fig:masks-ur5-conv-eval} we show some examples of masks that MaDi creates under the \texttt{conv} augmentation. These are all from one particular seed, masks from other seeds are generally similar but can differ slightly in their clarity. 

\vphantom{X}

\vphantom{X}

\vphantom{X}

\section{Hardware Resources}

\paragraph{Simulation experiments}
The experiments of this study were all individually run on single GPU setups. We have operated on distinct Compute Clusters with as GPU types: NVIDIA RTX A4000, NVIDIA V100SXM2, and NVIDIA P100 Pascal, with access to $4-6$ CPU cores each. A full training run of $500$K timesteps can take up to $36$ hours with batch size $128$. Most of our exploratory runs were done with batch size $64$. This halves the runtime approximately, while begetting little degradation in performance.

\paragraph{Robotic experiments}
We use a workstation with an AMD Ryzen Threadripper 2950 processor, an NVIDIA 2080Ti GPU, and 128G memory for the UR5 Robotic Arm experiments. The software setup uses multi-processing and multi-threading extensively to communicate between the arm, the camera, and the workstation. A full training run of $100$K timesteps takes around $2.5$ hours. In the sparse reward setting we train for $200$K steps, requiring $5$ hours. The batch size has no impact here, as we update the neural networks asynchronously in this setup.

\vspace{2em}

\section{Task Descriptions}
\label{app:task-descriptions}




In this section we provide additional details on the particular environments used in this study, both in simulation and on a physical robot.

\subsection{Simulation experiments}

We experiment on tasks from the benchmarks DMControl-GB \citep{hansen2021soda} and DistractingCS \citep{stone2021distracting}, both of which build on environments from DMControl \citep{tassa2018deepmind}. We clarify and provide properties of all the tasks considered in our study below, based on the descriptions from \citep{tassa2018deepmind, hansen2021svea}.
These DMControl tasks are selected based on previous work on both sample efficiency \citep{drq, hafner2020dream} and generalization \citep{hansen2021soda, stone2021distracting}, and represent a diverse and challenging skill set in the context of image-based RL. 
All environments emit observation frames $\mathbf{o} \in \mathbb{R}^{84\times84\times3}$. Three frames are stacked together to make one state $\mathbf{s} \in \mathbb{R}^{84\times84\times9}$. See Figure~\ref{fig:mask-examples-all-domains} for examples of observations from each domain.

\begin{itemize}
    \item \texttt{ball\_in\_cup-catch} ($\mathbf{a} \in \mathbb{R}^{2}$). An actuated planar receptacle is to move around in the plane and catch a ball attached by a string to its bottom. Sparse rewards.
    \item \texttt{cartpole-balance} ($\mathbf{a} \in \mathbb{R}$). Balance an unactuated pole by moving a cart left or right at the base of the pole. The pole starts in upright position. The agent is rewarded for balancing the pole within a fixed angle. Dense rewards.
    \item \texttt{cartpole-swingup} ($\mathbf{a} \in \mathbb{R}$). Swing up and balance an unactuated pole by moving a cart left or right at the base of the pole. The pole starts in downwards orientation. The agent is rewarded for balancing the pole within a fixed angle. Dense rewards.
     \item \texttt{finger-spin} ($\mathbf{a} \in \mathbb{R}^{2}$). A manipulation problem with a planar 2 DoF finger. The task is to continually rotate a free body on an unactuated hinge. Sparse rewards.
    \item \texttt{walker-stand} ($\mathbf{a} \in \mathbb{R}^{6}$). A bipedal planar walker that is rewarded for standing with an upright torso at a constant minimum height. Dense rewards.
    \item \texttt{walker-walk} ($\mathbf{a} \in \mathbb{R}^{6}$). A bipedal planar walker that is rewarded for walking forward at a certain target velocity. Dense rewards.
\end{itemize}

\vphantom{X}

\subsection{Robotic experiments}

\begin{figure}[H]
    \centering
    \begin{subfigure}[b]{0.35\columnwidth}
        \includegraphics[width=\textwidth]{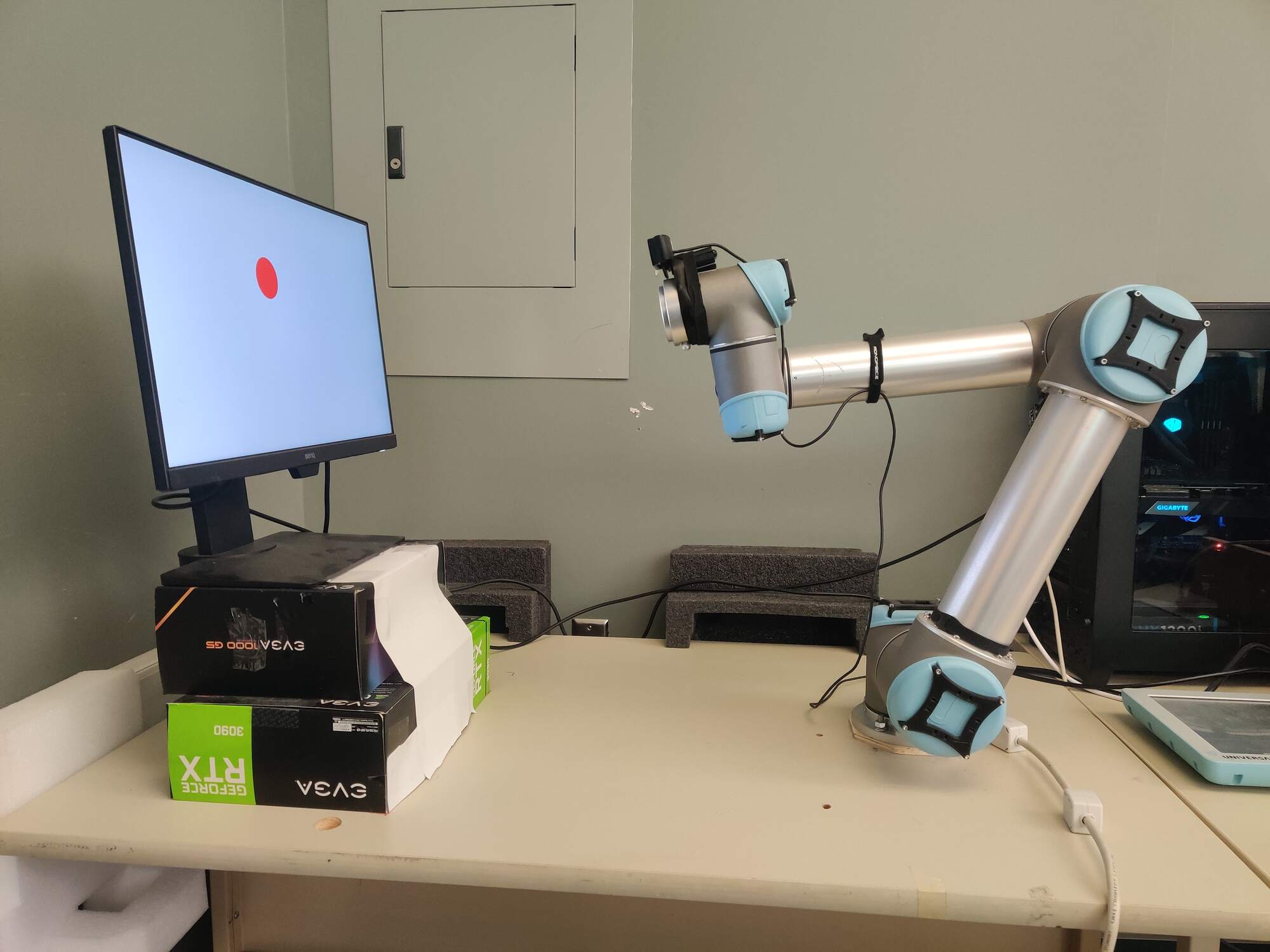}
        \caption{Initial state}
        \label{fig:subfig1-ur5-init}
    \end{subfigure}
    \hspace{0.03\columnwidth}
    \begin{subfigure}[b]{0.35\columnwidth}
        \includegraphics[width=\textwidth]{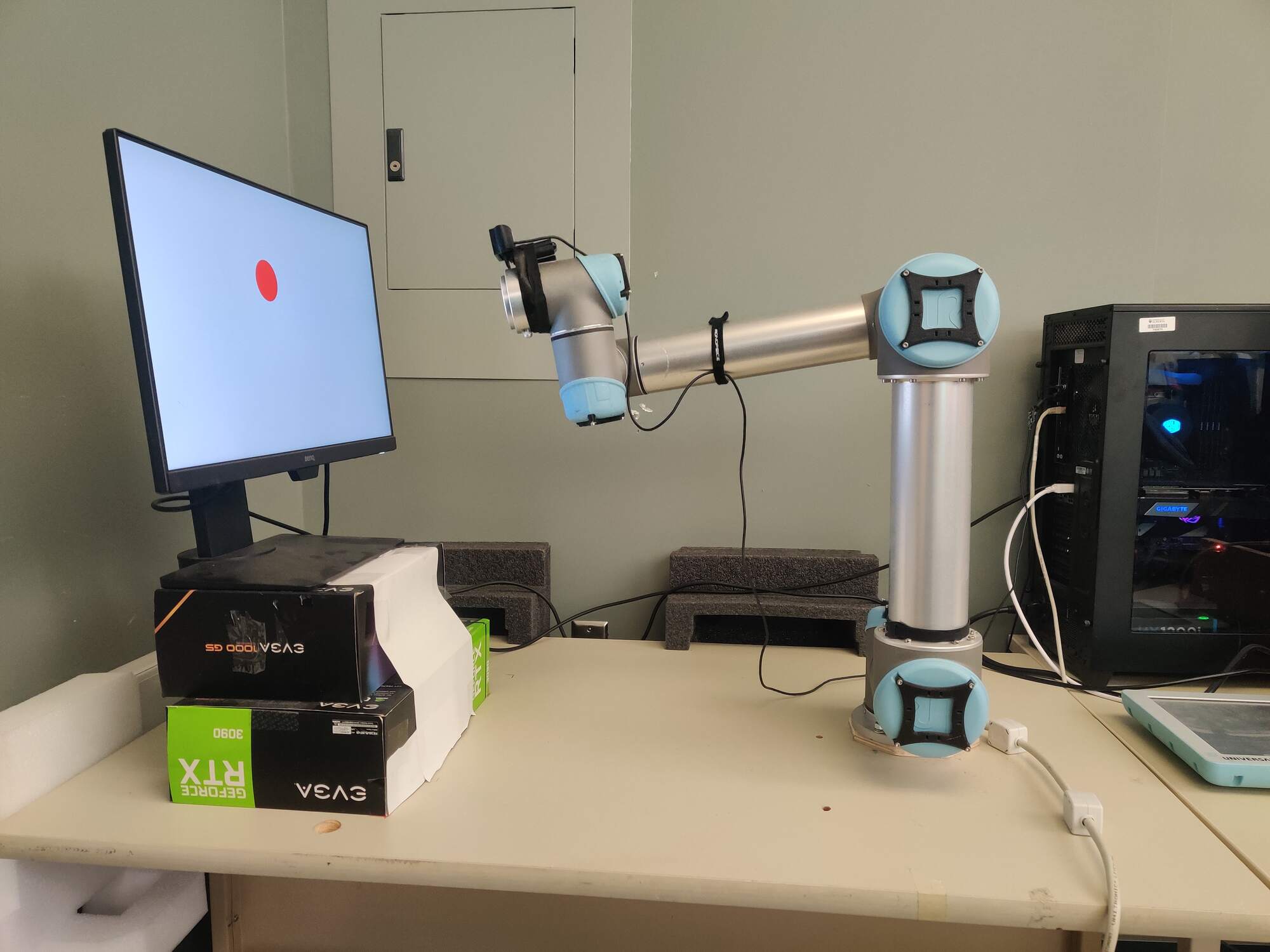}
        \caption{Reaching towards the goal}
        \label{fig:subfig2-ur5-reach}
    \end{subfigure}
    \captionsetup{width=0.8\textwidth}
    \caption{The UR5 Robotic Arm setup used in our experiments. The agent is rewarded for getting its camera as close as possible to the red dot randomly located on the screen.}
    \label{fig:ur5-og}
    \Description{This figure presents two images displaying the UR5 Robotic Arm setup from a side view.
In the first image, labeled "(a) Initial state," there's a computer monitor on the left showing a white screen with a red dot about 3 cm wide (in real life). To the right of the monitor, the UR5 robotic arm with 6 joints (only 4 are clearly visible) is in a resting position. It has a webcam camera attached to the tip of the arm.
The second image, "(b) Reaching towards the goal," depicts the same desk with monitor and robotic arm setup. However, in this image, the UR5 robotic arm has extended towards the monitor, aiming to approach the red dot.
The robotic arm is sophisticated, with joints and sections allowing for flexibility in movement. Its arm section are made of silver-colored pipes. The joints between sections are colored light blue.}
\end{figure}

\begin{figure}[H]
    \centering
        \begin{subfigure}[b]{0.19\textwidth}
        \includegraphics[width=\textwidth]{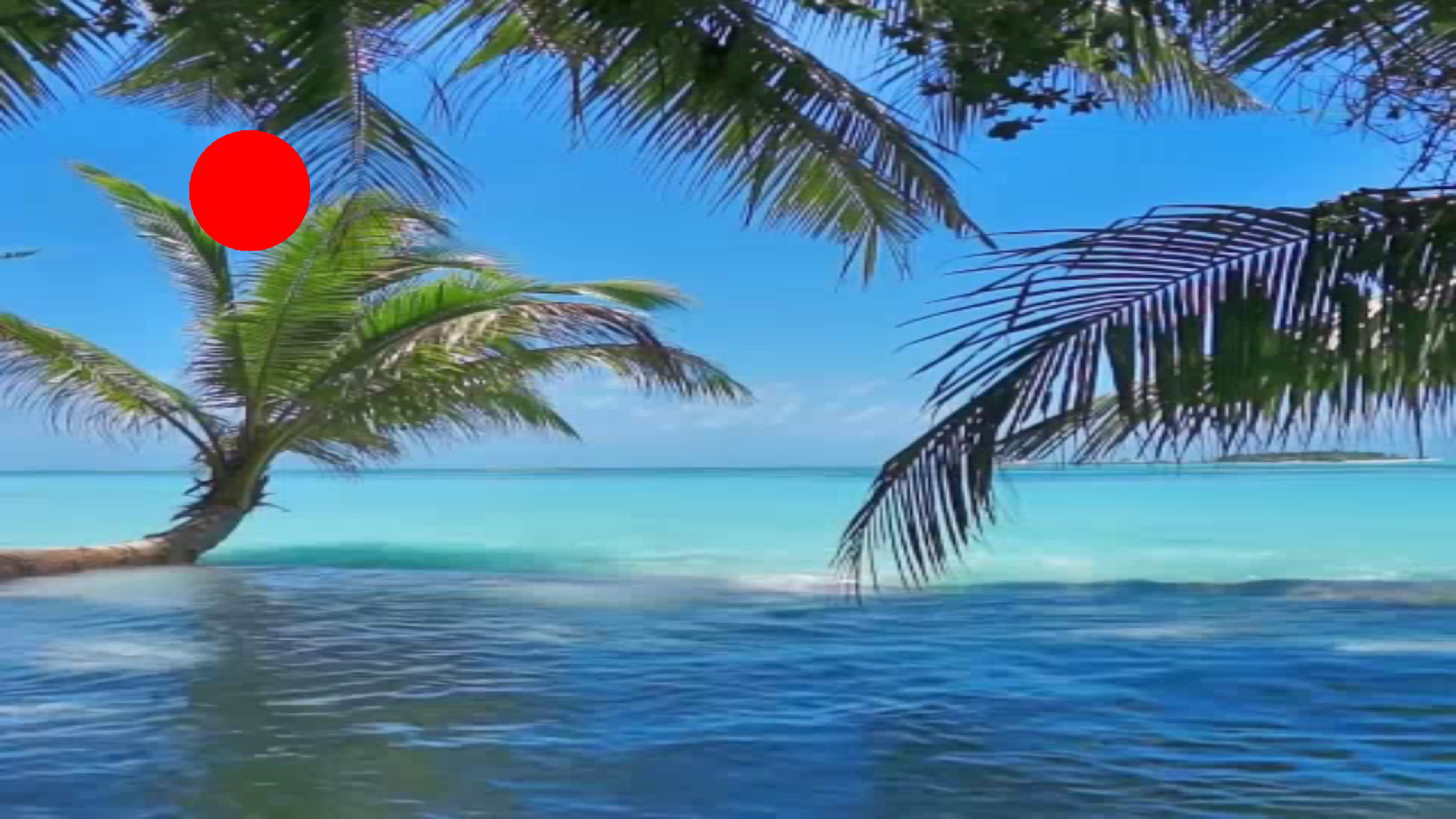}
        \vspace{-1.6em}
        \caption{}
        \vspace{0.8em}
        \label{fig:subfig1-vid}
    \end{subfigure}
    \hfill
    \begin{subfigure}[b]{0.19\textwidth}
        \includegraphics[width=\textwidth]{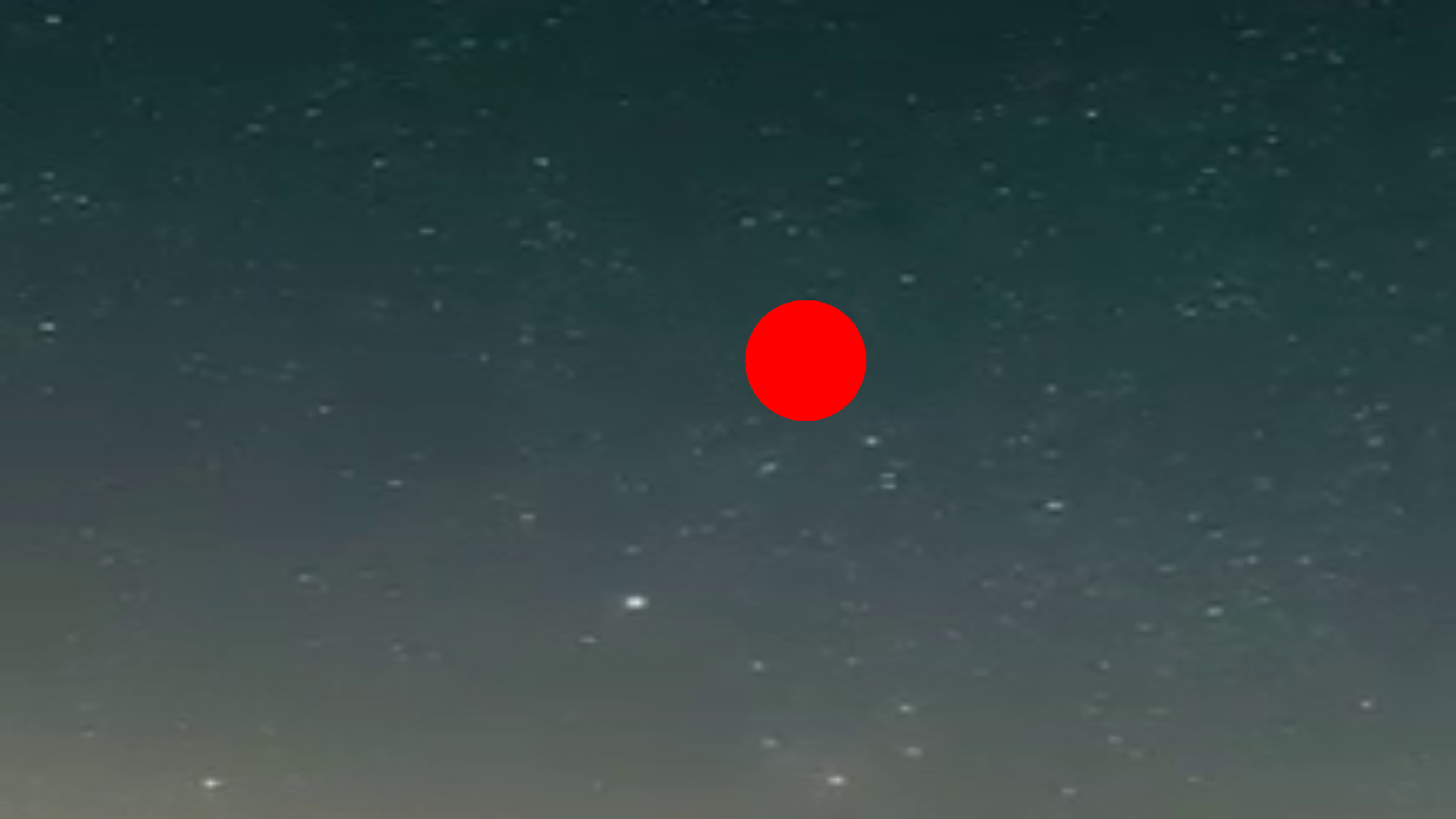}
        \vspace{-1.6em}
        \caption{}
        \vspace{0.8em}
        \label{fig:subfig2-vid}
    \end{subfigure}
    \hfill
    \begin{subfigure}[b]{0.19\textwidth}
        \includegraphics[width=\textwidth]{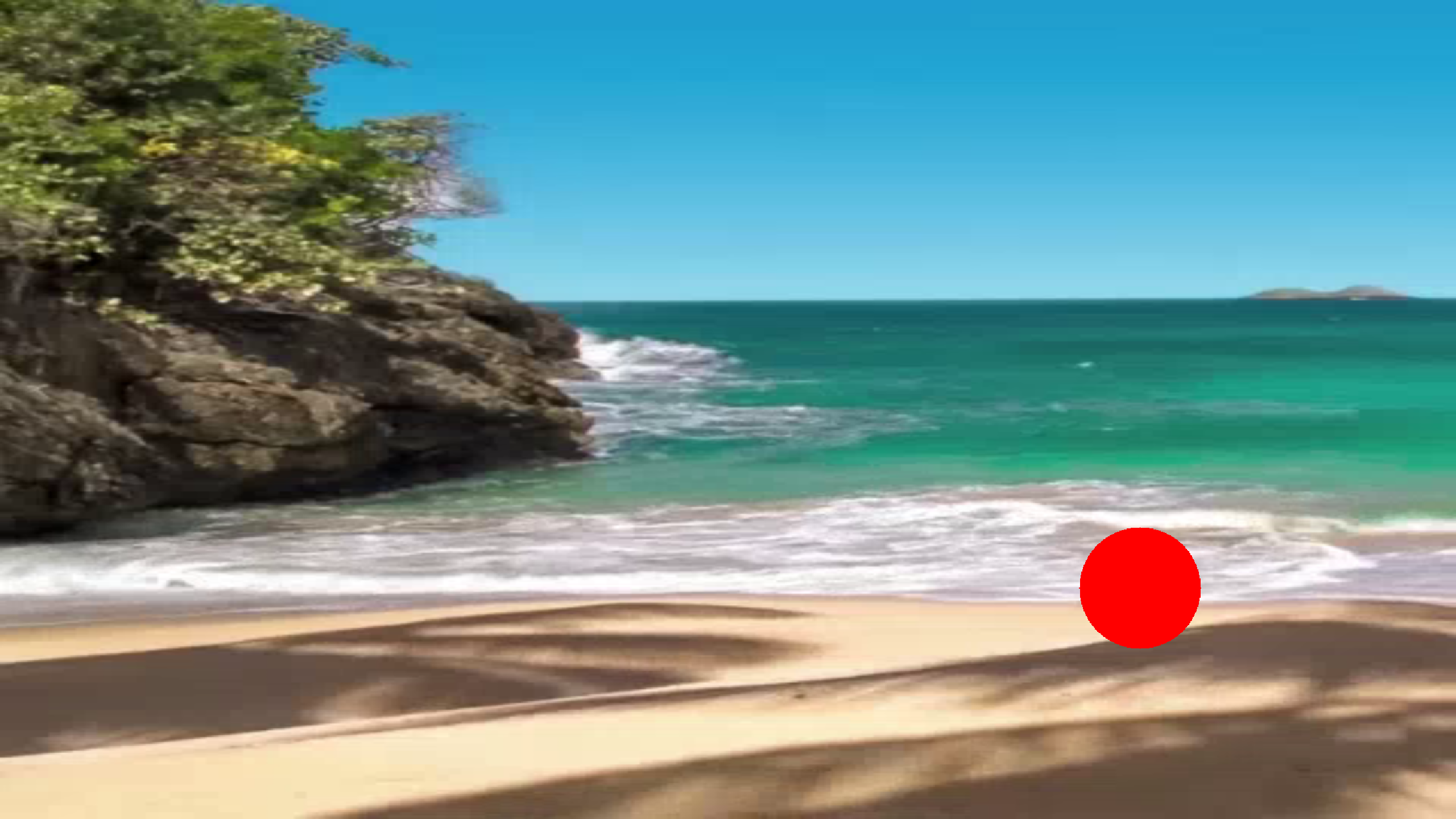}
        \vspace{-1.6em}
        \caption{}
        \vspace{0.8em}
        \label{fig:subfig3-vid}
    \end{subfigure}
        \hfill
    \begin{subfigure}[b]{0.19\textwidth}
        \includegraphics[width=\textwidth]{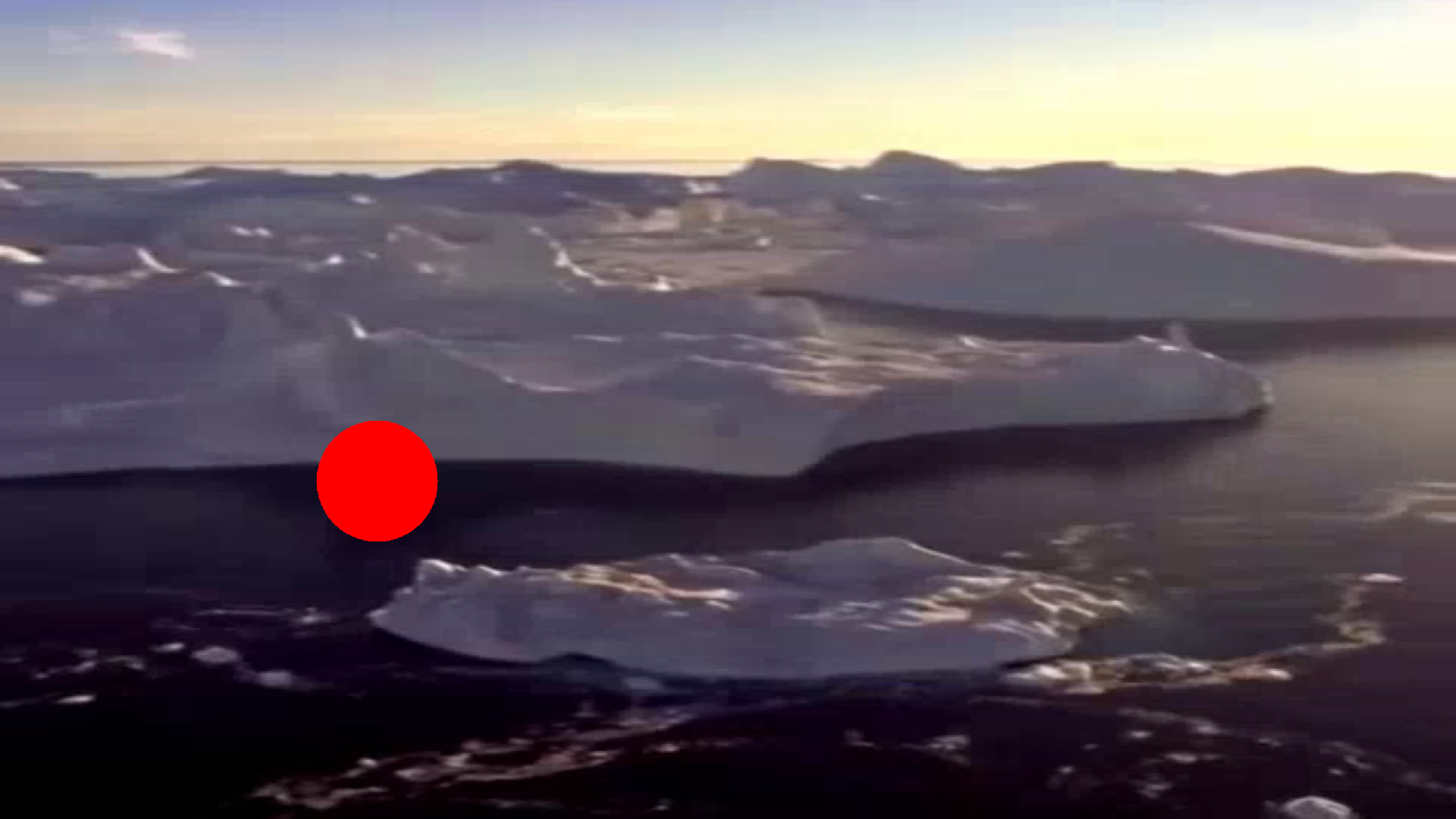}
        \vspace{-1.6em}
        \caption{}
        \vspace{0.8em}
        \label{fig:subfig4-vid}
    \end{subfigure}
        \hfill
    \begin{subfigure}[b]{0.19\textwidth}
        \includegraphics[width=\textwidth]{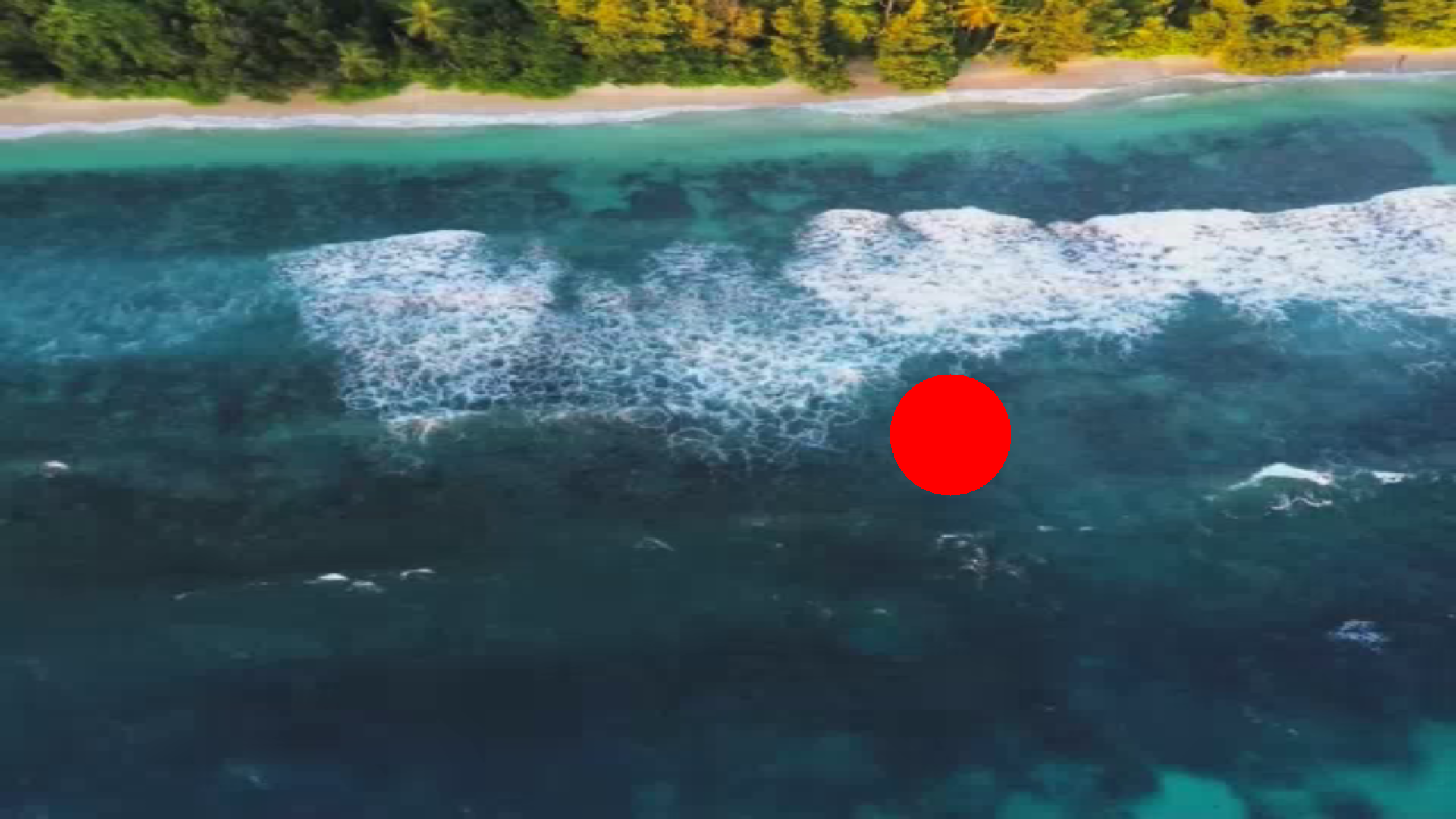}
        \vspace{-1.6em}
        \caption{}
        \vspace{0.8em}
        \label{fig:subfig5-vid}
    \end{subfigure}
    \vspace{-1.6em}
    \caption{Examples of frames from each video shown on the screen in the UR5-VisualReacher task to test generalization.}
    \label{fig:vid-frames}
    \Description{This figure shows five distinct rectangular frames, each depicting a different visual environment with a prominent red dot located in different random positions in the frame. These frames were displayed on the screen during the UR5-VisualReacher-VideoBackgrounds task to test generalization:
(a) A serene tropical setting, featuring a beach with many swaying palm trees, calm azure waters, and a clear blue sky.
(b) A more minimalistic and uniform blue gradient background, showing a starry night.
(c) An image of a tranquil forested lakeside, where calm waters reflect the dense trees and foliage of the surrounding forest.
(d) A majestic scene of icy mountains towering in the background, bathed in the light of a setting sun. The foreground showcases a calm lake with some ice floating around.
(e) A beach environment seen from the top where waves flow towards the beach. Most of the image is dark blue-green from the ocean. Beach is at the top.}
\end{figure}

We train MaDi and three other baselines on the UR-VisualReacher task \citep{yuan2022asynchronous, wang2023real} shown in Figure~\ref{fig:ur5-og}, and test the generalization performance on a novel experiment that we designed, dubbed UR-VisualReacher-VideoBackgrounds. The task of the robotic arm is the same in both environments: it gets rewarded for positioning its camera as close as possible to the red dot on the screen. In the testing environment, one of the videos shown in Figure~\ref{fig:vid-frames} is playing in the background. We take videos from the DeepMind Control Generalization Benchmark \citep{hansen2021soda}.

For the reward function: we adjust the RGB thresholds used to determine whether pixels are part of the red dot or not, such that it is robust against the video backgrounds. We selected videos that do not have any strong red colors in them, so that the rewards are well calibrated. It would be an interesting direction for future work to verify whether MaDi can detect the red dot even if other task-irrelevant pixels are red too. This would require a different setup of the reward function.


\end{document}